\documentclass[axioms,article,submit,pdftex,moreauthors]{Definitions/mdpi} 

\usepackage{graphicx}%
\usepackage{multirow}%
\usepackage{amsmath,amssymb,amsfonts}%
\usepackage{amsthm}%
\usepackage{mathrsfs}%
\usepackage[title]{appendix}%
\usepackage{xcolor}%
\usepackage{textcomp}%
\usepackage{manyfoot}%
\usepackage{booktabs}%
\usepackage{algorithm}%
\usepackage{algorithmicx}%
\usepackage{algpseudocode}%
\usepackage{listings}%
\usepackage{tikz-cd}

\let\linenumbers\relax
\usepackage{amssymb}
\usepackage{amsmath}

\usepackage{proof} 

\firstpage{1} 
\pubvolume{1}
\issuenum{1}
\articlenumber{0}
\pubyear{2024}
\copyrightyear{2024}
\datereceived{ } 
\daterevised{ } 
\dateaccepted{ } 
\datepublished{ } 
\hreflink{https://doi.org/} 

\usepackage{url}

\usepackage{xcolor}
\usepackage{amsmath,amsfonts}
\usepackage{thmtools} 
\usepackage{graphicx}
\usepackage{booktabs}
\usepackage{subcaption} 
\usepackage{mathtools}
\usepackage{float}
\usepackage{caption}

\usepackage{stmaryrd} 
\usepackage{todonotes}
\usepackage{scalefnt}

\newcommand{\doi}[1]{\textsc{doi}: \href{http://dx.doi.org/#1}{\nolinkurl{#1}}}
\DeclareMathOperator{\Ca}{\mathsf{C}}

\newcommand{\Z}{\ensuremath{\mathbb Z}}

\newcommand{\diset}[2]{\left({{#1} \atop {#2}}\right)} 

\newcommand{\NamedCat}[1]{\mathsf{#1}}

\newcommand{\PolyZ}{\NamedCat{POLY}_{\Z_2}}

\newcommand{\Para}{\textbf{Para}}

\newcommand{\Lens}{\textbf{Lens}}

\newcommand{\<}{\langle}
\renewcommand{\>}{\rangle}

\newcommand{\id}{\text{id}}

\newcommand{\putt}{\ensuremath{\mathsf{put}}}

\newcommand{\cp}[0]{\ensuremath{\fatsemi}} 

\usepackage{listings}
\makeatletter
\def\lst@makecaption{%
  \def\@captype{table}%
  \@makecaption
}
\makeatother

\usepackage{amsthm}
\usepackage{amsmath}
\usepackage{amssymb}
\usepackage{bm}
\usepackage{bbm}
\usepackage{pdfpages}
\usepackage{float}
\usepackage{dsfont}
\usepackage{relsize}
\usepackage{titlesec}
\usepackage{todonotes}
\def\references#1{\vspace*{5mm}\noindent{References:}\list
{[\arabic{enumi}]}{\settowidth\labelwidth{[#1]}\leftmargin\labelwidth
\advance\leftmargin\labelsep
\usecounter{enumi}}
\def\newblock{\hskip .11em plus .33em minus .07em}
\sloppy\clubpenalty4000\widowpenalty4000
\sfcode`\.=1000\relax}

\usepackage{stmaryrd}

\Title{Category-Theoretical and Topos-Theoretical Frameworks in Machine Learning: A Survey}

\TitleCitation{Category-Theoretical and Topos-Theoretical Frameworks in Machine Learning: A Survey}

\Author{Yiyang Jia $^{1,*}$, Guohong Peng $^{2}$, Zheng Yang $^{3}$, and Tianhao Chen $^{4}$}

\AuthorNames{Yiyang Jia, Guohong Peng, Zheng Yang, Tianhao Chen}

\AuthorCitation{Jia, Y.; Peng, G.; Yang, Z.; Chen, T.}

\address{%
$^{1}$ \quad Department of Mathematical, Physical, and Information Science, Japan Women's University, Tokio, Japan; jiay@fc.jwu.ac.jp\\
$^{2}$ \quad Department of Mechanical and Electrical Engineering, XiChang University, Sichuan, China; ghpeng@xcc.edu.cn\\
$^{3}$ \quad Pittsburgh Institute, Sichuan University, Chengdu, Sichuan, China; zhengyang2018@scu.edu.cn\\
$^{4}$ \quad Faculty of Applied Sciences, Macao Polytechnic University, Macao, China; tianhao.chen@mpu.edu.mo}

\corres{Correspondence: jiay@fc.jwu.ac.jp (Y.J.)}

\abstract{In this survey, we provide an overview of category theory-derived machine learning from four mainstream perspectives: gradient-based learning, probability-based learning, invariance and equivalence-based learning, and topos-based learning. For the first three topics, we primarily review research in the past five years, updating and expanding on the previous survey by Shiebler et al.. The fourth topic, which delves into higher category theory, particularly topos theory, is surveyed for the first time in this paper. 
In certain machine learning methods, the compositionality of functors plays a vital role, prompting the development of specific categorical frameworks. However, when considering how the global properties of a network reflect in local structures and how geometric properties and semantics are expressed with logic, the topos structure becomes particularly significant and profound.}


\begin{document}

\section{Introduction and Background}
In recent years, there has been an increasing amount of research involving category theory in machine learning. This survey primarily reviews recent research on the integration of category theory in various machine learning paradigms. Roughly, we divide this research into two main directions:

\begin{enumerate}
    \item Studies on specific categorical frameworks corresponding to specific machine learning methods. For instance, research has explored the backpropagation algorithm within Cartesian differential categories, probabilistic machine learning methods within Markov categories, and clustering algorithms within the category of metric spaces, etc.
    
    \item Methodological approaches that explore the potential applications of category theory to various aspects of machine learning from a broad mathematical perspective. For example, considerations of the compositional nature and properties of deep learning in the context of topological spaces, 2-categories, toposes, stacks, and other categorical spaces.
\end{enumerate}

\begin{figure}[ht]
\centering
\includegraphics[width=1\textwidth]{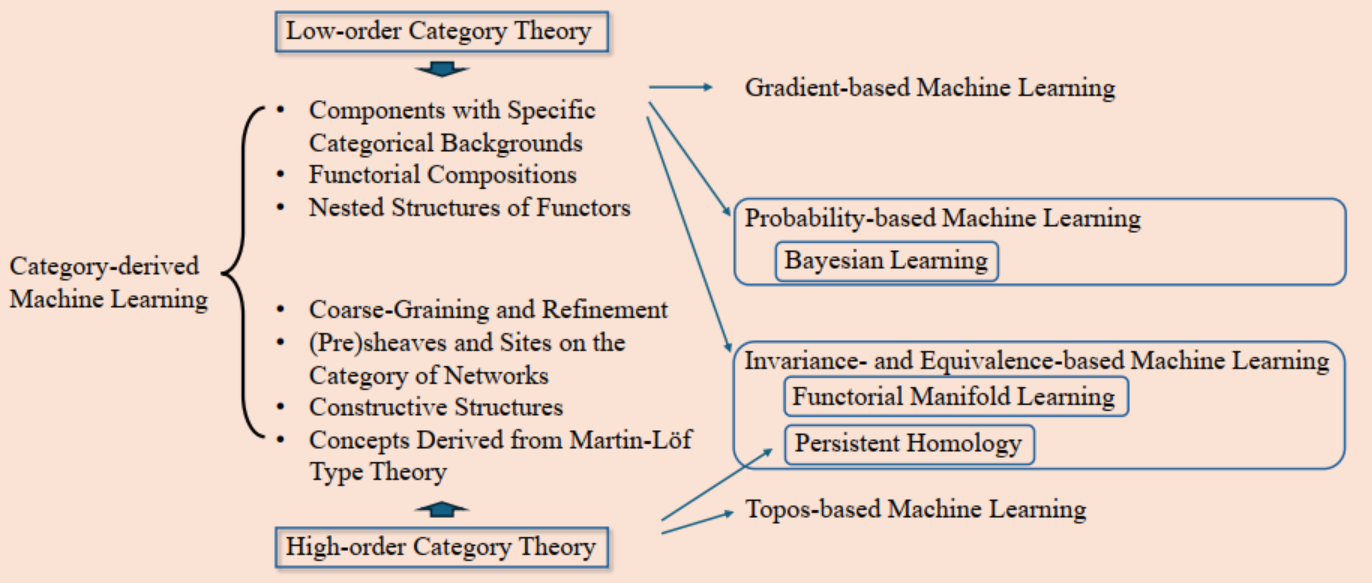}
\caption{Category-derived Machine Learning Framework}
\label{framework}
\end{figure}

Our framework, shown in Figure \ref{framework}, outlines the structure of category-derived learning, which we divide into two parts based on the order of category theory. Low-order category theory primarily provides functorial frameworks, where a specific category serves as the foundation. Components are defined independently, with their own inputs, outputs, and parameters, and are composed using functors. Functors, much like modules in programming languages, can be composed or nested. This perspective has inspired work in fields such as gradient-based machine learning, probability-based machine learning (notably Bayesian learning), and invariance- and equivalence-based machine learning (particularly functorial manifold learning). High-order category theory primarily offers a comprehensive perspective on global properties. For instance, coarse-graining and refinement can be encoded within high-order category-derived structures, such as sites on networks. Additionally, constructive structures like stacks may be considered in certain contexts, while the semantics of the learning process are analyzed using concepts derived from Martin-Löf Type Theory.

For the first direction, the survey by Shiebler et al. \cite{shiebler2021categorya} provided a detailed overview of the main results up to 2021. They focused on three main topics: gradient-based learning, learning involving probability and statistics (such as Bayesian learning), and invariant and equivariant learning. Composability is the most emphasized property in these investigations, as it leverages the functoriality to explain the decomposition of entire framework of learning processes and to combine basic components to meet specific requirements in various practical learning tasks. In this survey, we complement their survey by including more recent results from 2021 to the present. Additionally, we introduce some other branches beyond these mainstream results.

The functionality derived from the composition of components with algebraic features or constrained metric-based structures ensures the uniformity of the system's overall design. It also provides a framework to trace how the components interact and how the functions of each part contribute to the overall effects of the system. However, certain aspects, such as causality, cannot be fully captured or encoded within the low order category-based design. Causality, in fact, plays a crucial role in understanding and enhancing machine learning systems, especially in tasks that demand reasoning about underlying mechanisms rather than relying only on statistical correlations. 

The second direction, high-order category theory-derived learnings, especially emphasized in this survey, explores causality from a deeper perspective by leveraging the rich structures of sheaves and presheaves. 
These constructs effectively capture local-global relationships in datasets, such as the impact of local interventions on global outcomes. Furthermore, the internal logic of a topos provides a robust foundation for reasoning about counterfactuals and interventions, both of which are central to causal inference. For example, a systematic framework based on functor composition provides a robust design-oriented approach and effectively encodes ``parallelism'' by describing independent processes. However, it struggles to fully encode ``concurrency'', where events can be both 'fired' and 'enabled' at the same time, and to model the dynamic state transitions of these events. On the other hand, a topos, with its inherent (co)sheaf structures, naturally accommodates graph-like networks. When applied in this context, it implicitly supports a Petri net structure, which excels at capturing the dynamic activation and interactions of events, including concurrency and state changes\cite{lu2022causalnetworkcondensation}.

Existing research in the second direction mainly applies topos theory to machine learning, particularly in the context of deep learning networks. These studies primarily serve two main purposes: providing new theoretical insights to address issues of 'uninterpretability' and improving model performance. Since the discovery of connections between topos structures and machine learning, new solutions have been proposed for specific challenges.
For example, in the dimension-reduction of high-dimensional data, traditional methods often result in information loss, risking the omission of important features. However, by introducing topos theory and utilizing its algebraic properties, it is possible to achieve dimensionality reduction while preserving the essential structure of high-dimensional data, minimizing information loss, and extracting important features.
Another example is the interpretation and explanation of complex machine learning models, such as deep learning models, which often have internal structures that are difficult to interpret and explain, commonly referred to as ``black boxes''. By using the logical reasoning capabilities of topos theory, it becomes possible to logically interpret and explain the internal structures and outputs of these models. The introduction of semantics allows for the interpretation of ``emergent'' phenomena in large-scale models.
Furthermore, the analysis of dynamic data is challenging with traditional static data analysis methods due to the changing nature of the data over time. Topos theory allows for the natural capture of temporal changes and evolution in the data in a local vs. global context. In this survey, we summarize the recent results involving topos-based structures and geometric theories applied in machine learning.
Currently, much of the work introduced in the second direction is spearheaded by Laurent Lafforgue's team. While their contributions are innovative and highly creative, their applicability in practical machine learning scenarios remains a topic of debate. We include their work in this survey because it is not only groundbreaking but also highly instructive for advancing this field. Furthermore, several of their arguments present valuable points for discussion and critique, serving as a foundation for further exploration in this direction. We anticipate that over time, many other studies based on topos theory will emerge, enriching the landscape of research and broadening its applications in machine learning.

In the following, we generally use a sans serif font to represent categories and a bold font to represent functors, though we may occasionally emphasize sheaves/toposes with different fonts.

\section{Developments in Gradient-based Learning}
\label{sec_2}
The primary objective of integrating concepts from category theory into concrete, modularizable learning processes or methods is to leverage compositionality, enabling existing processes to be represented through graphical (diagrammatic) representations and calculus, where the modular design facilitates replacements of individual components. For existing methods, in \cite{ruder2016overview}, the author listed different gradient descent optimization algorithms and compared their behavior. 
The categorical approach, on the other hand, highlights their similarities and integrates different algorithms/optimizers into a common framework within the learning process. In \cite{cruttwell2022categorical}, the authors developed categorical frameworks focused on the most classical and straightforward gradient-based learning methods, demonstrating the variations achievable through the composition of the categorical components they defined. Specifically, the graphical representation, semantic properties, and diagrammatic reasoning—key aspects of category theory—are regarded as a rigorous semantic foundation for learning. They also facilitate the construction of scalable deep learning systems through compositionality.

In this section, we aim to explain how (low-order)category semantics are applied to understand the fundamental structures of gradient-based learning, which is a core component of the deep learning paradigm. The mainstream approach is to decompose learning into independent components: the parameter component, the bidirectional dataflow component, and the differentiation component. These components, especially the bidirectional components—lenses and optics—are widely applied in other categorical frameworks of machine learning. After introducing the basic framework, we will present related research in Section~\ref{other_re}.

Shiebler et al. \cite{shiebler2021categorya} provided an overview of the fundamental structures in their survey. However, their definitions and explanations may be quite challenging for general readers without a mathematical background. Therefore, we briefly outline the most commonly used concepts in a more accessible manner.To technically introduce the critical parametric lens structure in this work, we begin by outlining the three key characteristics of gradient-based learning identified in \cite{cruttwell2022categorical, cruttwell2024deep}. 
Each characteristic motivates a specific categorical construction, respectively.
\begin{enumerate}
    \item The \textbf{parametric} nature of computation corresponds to the $\Para$ functor;
    \item \textbf{Bidirectional} information flow justifies the use of the $\Lens$ functor, in which a pair of forward-propagation and back-propagation are considered at the same time;
    \item  The \textbf{differentiation} involved in the update of parameters (or inputs in some cases) via gradient descent is captured by the background category: a Cartesian Reverse Differential Category (CRDC). 
\end{enumerate}

These concepts are summarized in \autoref{table:insights}.

\begin{table}[htp]
\vspace{-15pt}
\caption{Summary of the key characteristics in gradient-based learning with corresponding construction} 
\label{table:insights}
\vskip 0.02in
\centering
\begin{small}
\begin{tabular}{l|l|p{6cm}} 
\toprule
\textbf{Characteristic} & \textbf{Construction} & \textbf{Motivation} \\
\toprule
Parametricity & $\Para$ & A neural network is a mapping with parameter, i.e., a function \( f : P \times X \to Y \), and a supervised \emph{learning} is to find a "good" parameter \( p : P \) for \( f(p, -) \). Parameters also arise elsewhere like the loss function. \\
\midrule
Bidirectionality & $\Lens$ & Information flows bidirectionally as inputs (forward) are sent to outputs and loss through sequential layers, and backpropagation then reverses this flow to update parameters (backward). \\
\midrule
Differentiation & CRDC & Differentiate the loss function that maps a parameter to its associated loss to reduce that loss, which the CRDC can capture. \\
\bottomrule
\end{tabular}
\end{small}
\end{table}

Note that these basic settings can be extended or modified with more complex structures or analogous frameworks to accommodate various learning tasks, methods, or datasets. For instance, bicategory-based and actegory-based approaches have been developed to adapt to learning scenarios involving different objects (such as polynomial circuits) or methodologies (such as the Bayesian learning framework introduced in Section 3. In this case, the general lenses derived from the \(\Lens\) functor are replaced by Grothendieck lenses). 
The ``Cartesian'' property of the background category enables duality, particularly in the form of products and coproducts. Some discussions in \cite{capucci2021towards, gavranovic2024fundamental} enriched the above framework with dual components. For instance, the concepts of ``parametrized'' and ``coparametrized'' morphisms, states and costates, and algebra and coalgebra are introduced as paired concepts within this context, integrating various networks such as GCNNs, GANs, and different learning problems as open learners and open games into a unified framework. From a higher-order categorical perspective, the ``Cartesian'' property in informatics often signifies the potential to apply Lawvere theory for modeling algebraic structures within the base category.

Now we give the definitions of these functors and categories in order. 

\subsection{Fundamental Components: The Base Categories and Functors}
\begin{Definition}[Functor \(\mathbf{Para}\) \cite{cruttwell2022categorical}] \label{def:paracat}
Let \((\Ca, \otimes, I)\) be a \textbf{strict symmetric monoidal category}. The mapping by \(\mathbf{Para}\) results in a category \(\mathbf{Para}(\Ca)\) with the following structure:

\begin{itemize}
  \item Objects: the objects of \(\Ca\).
  \item Morphisms: pairs of the form \((P, f)\), representing a map from the input \(A\) to the output \(B\), with $P$ being an object of $\Ca$ and $f: P \otimes A \to B$. 
  \item Compositions of morphisms: the composition of morphisms \((P, f): A \to B\) and \((Q, g): B \to C\) is given by the pair \((Q \otimes P, (1_{Q} \otimes f) \cp g): A \to C\). 
  \item Identity: The identity endomorphism on \(A\) is the pair \((I, 1_A)\). (\(I \otimes A = A\) due to the strict monoidal property).
\end{itemize}
\end{Definition}

Every parametric morphism has a horizontal, but also a vertical component, emphasized by its string diagram representation 

$$
\begin{tikzpicture}[auto]

\node (P) [above] {$P$};
\node (f) [below of=P, draw, rectangle] {$f$};
\node (A) [left of=f] {$A$};

\node (B) [right of=f] {$B$};

\draw[->] (P) -- (f);
\draw[->] (A) -- (f);
\draw[->] (f) -- (B);

\end{tikzpicture}
$$

In supervised parameter learning, the update of parameters is formulated as 2-morphisms in $\Para(\Ca)$, called \emph{reparameterisations}.
A reparameterisation from $(P, f) \Rightarrow (Q, f')$ is a morphism $\alpha : Q \to P$ in $\Ca$ such that the diagram 

$$
\begin{tikzcd}
P \otimes A \arrow[rd, "f"]                                       &   \\
Q \otimes A \arrow[u, "\alpha \otimes 1_A"] \arrow[r, "f^\prime"] & B
\end{tikzcd}
$$

is commutes in $\Ca$, yielding a new map $(Q, (\alpha \otimes 1_A) \cp f): A \to B$. 
Following the authors write $f^\alpha$ for the reparameterisation of $f$ with $\alpha$, as show in this string diagram representation

\[
\begin{tikzpicture}[auto]

\node (Q1) at (0, 2) {$Q$};
\node (alpha) at (0, 1) [draw, rectangle] {$\alpha$};
\node (P) at (0, 0) {$P$};
\node (f) at (0, -1) [draw, rectangle] {$f$};
\node (A) at (-2, -1) {$A$};
\node (B) at (2, -1) {$B$};

\draw[->] (Q1) -- (alpha);
\draw[->] (alpha) -- (P);
\draw[->] (P) -- (f);
\draw[->] (A) -- (f);
\draw[->] (f) -- (B);

\node (equals) at (3, -0.5) {$=$};

\node (Q2) at (6, 1) {$Q$};
\node (fa) at (6, -1) [draw, rectangle] {$f^\alpha$};
\node (A2) at (4, -1) {$A$};
\node (B2) at (8, -1) {$B$};

\draw[->] (Q2) -- (fa);
\draw[->] (A2) -- (fa);
\draw[->] (fa) -- (B2);

\end{tikzpicture}
\]

where $f^{\alpha}=f'$. 
In this context, 
based on the vertical morphism setting, \(\mathbf{Para}(\Ca)\) can be viewed as a bicategory with models as its 1-morphisms associating inputs and outputs, and with reparameterizations as its 2-morphisms associating models.

The following \(\Lens\) construction facilitates bidirectional data transmission.
\begin{Definition}[ Functor \(\Lens\) \cite{cruttwell2022categorical}]

For any Cartesian category $\Ca$, the mapping of functor \(\Lens\) results in the category with the following data: 

\begin{itemize}
  \item Objects are pairs $(A,A')$ of objects in $\Ca$.
  \item A morphism from $(A,A')$ to $(B, B')$ consists of a pair of morphisms in $\Ca$, denoted as $(f, f_r)$, as illustrated in 

$$
\begin{tikzpicture}[auto]

\node (A) at (1.5, 0.2) {$A$};
\node (A') at (1.5, -0.2) {$A'$};
\node (B) at (4.5, 0.2) {$B$};
\node (B') at (4.5, -0.2) {$B'$};
\node (f) at (3, 0) [draw, rectangle] 
    {\(
    \begin{pmatrix}
    f \\
    f_r
    \end{pmatrix}
    \)};

\draw[->] (A.east) -- ([yshift=0.2cm]f.west);
\draw[->] (A'.east) -- ([yshift=-0.2cm]f.west);
\draw[->] ([yshift=0.2cm]f.east) -- (B.west);
\draw[->] ([yshift=-0.2cm]f.east) -- (B'.west);

\end{tikzpicture}
$$
  where $f: A \to B$ is called the \textbf{get} or \textbf{forward} part of the lens and $f_r: A \times B' \to A'$ is called the \textbf{put} or \textbf{backwards} part of the lens. The inside construction is illustrated as in 
$$
\scalebox{0.5}{\includegraphics{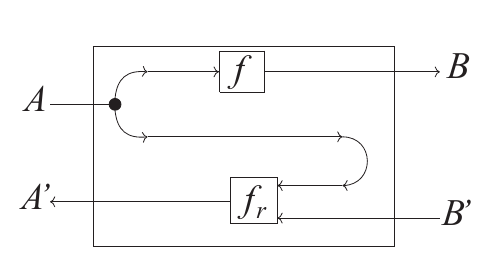}}
$$

  \item The composition of $(f, f_r): (A,A') \to (B, B')$ and $(g, g_r): (B, B') \to (C, C')$ is given by get $f \cp g$ and put $\<\pi_0,\<\pi_0 \cp f, \pi_1\> \cp g_r\> \cp f_r$. The graphical notation is 
$$
\scalebox{0.5}{\includegraphics{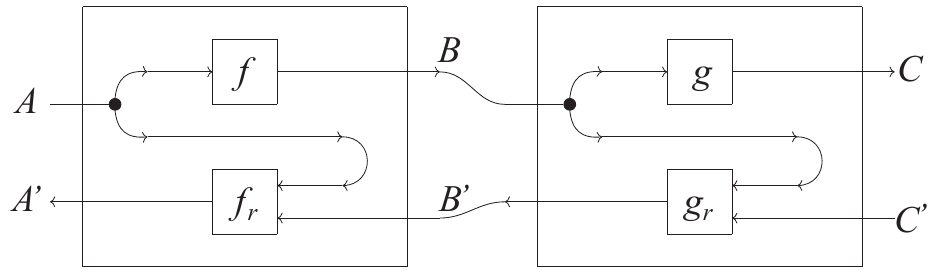}}
$$
\item The identity on $(A,A')$ is the pair $(1_A, \pi_1)$.
\end{itemize}
\end{Definition}

\begin{Definition}[Cartesian left additive category \cite{blute2009Cartesian}]
A \textbf{Cartesian left additive category} \(\Ca\) is both a Cartesian category and a left additive category. A Cartesian category is specified by four components: binary products \(\times\), projection maps \(\pi_i\), pairing operation \(\langle -, -\rangle\), and a terminal object. A left additive category is defined by commutative monoid hom-sets, an addition operation \(+\), zero maps 0, with compositions on the left that are compatible with the addition operation. The compatibility in \(\Ca\) is reflected in the fact that the projection maps of \(\Ca\) as a Cartesian category are additive.
\end{Definition}

\begin{Definition}[Cartesian differential category~\cite{blute2009Cartesian}]
A \textbf{Cartesian differential category} is a Cartesian left additive category with a \textbf{differential combinator} \(D\), with which the inference rule is given by:
\[
\frac{A_1 \xrightarrow{f} A_2}{A_1 \times A_1 \xrightarrow[{D[f]}]{} A_2}
\]
\(D[f]\), which satisfies the following axioms, is called the derivative of \(f\).

\noindent {\bf [Preservation of addition]} 
$D[f+g] = D[f] + D[g]$ and $D[0]=0$;\\
\noindent{\bf [Additivity in the linear variable]}
$\<a,b+c\>D[f] = \<a,b\> D[f]+\<a,c\>D[f]$ and $\<a,0\>D[f] = 0$;\\
\noindent {\bf [Derivative of identities and projections]} 
$D[1]=\pi_1$, $D[\pi_0]=\pi_1\pi_0$, and $D[\pi_1] =\pi_1\pi_1$;\\
\noindent {\bf [Derivative of pairings]} 
$D[\<f,g\>] = \<D[f],D[g]\>$;\\
\noindent{\bf [Chain Rule of Derivative]} 
$D[fg] = \<\pi_0f,D[f]\>D[g]$;\\
\noindent {\bf [Linearity of Derivative]} 
$\<\<a,b\>,\<0,c\>\>D[D[f]] = \<a,c\>D[f]$;\\
\noindent {\bf [Symmetry of Mixed Partial]}
$\<\<a,b\>,\<c,d\>\>D[D[f]] = \<\<a,c\>,\<b,d\>\>D[D[f]]$.
\end{Definition}

A straightforward relationship between the forward and reverse derivatives in standard computations, in terms of function approximations, is given as follows:

\[
\begin{aligned}
&f(x) + y' \approx f(x + R[f](x, y')) \\
&f(x + x') \approx f(x) + D[f](x, x')
\end{aligned}
\]
where the following correspondence becomes clear

\[
\begin{aligned}
&D[f](x, x') \leftrightarrow y' \\
&x' \leftrightarrow R[f](x, y')
\end{aligned}
\]

\begin{Definition}[Cartesian reverse differential category(CRDC), first introduced by \cite{blute2009Cartesian} and first applied to the context of machine learning and automatic differentiation by \cite{cockett2019reverse}]
A \textbf{Cartesian reverse differential category} is a Cartesian left additive category \(\mathcal{X}\) with a \textbf{reverse differential combinator} $R$, with which the inference rule is given by:

\[
\frac{A_1 \xrightarrow{f} A_2}{A_1 \times A_2 \xrightarrow[\mathit{R[f]}]{} A_1}
\]

where \(R[f]\) satisfying the following axioms is called the reverse derivative of \(f\):

\noindent {\bf[Derivative preserves addition]} 
$R[f+g] = R[f] + R[g]$ and $R[0]=0$; \\
\noindent {\bf[Additivity in the linear variable]} 
$\<a,b+c\>R[f] = \<a,b\>R[f] + \<a,c\>R[f]$ and  $\<a,0\>R[f] = 0$ \\
\noindent {\bf[Derivatives of identities and projections]}  
$R[1] = \pi_1$, while for the projections, the following diagrams commute:\\
$$
\begin{tikzpicture}[auto]

\node (A1A2) at (1, 0) {$A_1 \times A_2$};
\node (A1) at (3, 0) {$A_1$};
\node (line1) at (-1.5, -0.5) {}; 
\node (line2) at (4.5, -0.5) {};

\draw[->] (A1A2) -- (A1) node[midway, above] {$\pi_0$};

\draw (line1) -- (line2);

\node (A1A2A1) at (0, -1.3) {$(A_1 \times A_2) \times A_1$};
\node (A1A2b) at (3, -1.3) {$A_1 \times A_2$};

\draw[->] (A1A2A1) -- (A1A2b) node[midway, above] {$R[\pi_0]$};

\end{tikzpicture}
\hspace{1cm}
\begin{tikzpicture}[auto]

\node (A1A2A1b) at (-1, -0.5) {$(A_1 \times A_2) \times A_1$};
\node (A1A2c) at (2, -0.5) {$A_1 \times A_2$};
\node (A1c) at (0.5, -2) {$A_1$};

\draw[->] (A1A2A1b) -- (A1A2c) node[midway, above] {$R[\pi_0]$};
\draw[<-] (A1c) -- (A1A2A1b) node[midway, left] {$\pi_1$};
\draw[->] (A1c) -- (A1A2c) node[midway, right] {$\iota_0$};
\end{tikzpicture}
$$

$$
\begin{tikzpicture}[auto]

\node (A1A2) at (1, 0) {$A_1 \times A_2$};
\node (A2) at (3, 0) {$A_2$};
\node (line1) at (-1.5, -0.5) {}; 
\node (line2) at (4.5, -0.5) {};

\draw[->] (A1A2) -- (A2) node[midway, above] {$\pi_1$};

\draw (line1) -- (line2);

\node (A1A2A2) at (0, -1.3) {$(A_1 \times A_2) \times A_2$};
\node (A1A2b) at (3, -1.3) {$A_1 \times A_2$};

\draw[->] (A1A2A2) -- (A1A2b) node[midway, above] {$R[\pi_1]$};

\end{tikzpicture}
\hspace{1cm}
\begin{tikzpicture}[auto]

\node (A1A2A2b) at (-1, -0.5) {$(A_1 \times A_2) \times A_2$};
\node (A1A2c) at (2, -0.5) {$A_1 \times A_2$};
\node (A2c) at (0.5, -2) {$A_2$};

\draw[->] (A1A2A2b) -- (A1A2c) node[midway, above] {$R[\pi_1]$};
\draw[<-] (A2c) -- (A1A2A1b) node[midway, left] {$\pi_1$};
\draw[->] (A2c) -- (A1A2c) node[midway, right] {$\iota_1$};
\end{tikzpicture}
$$
%

\noindent {\bf[Derivative of Pairings]}
For a tupling of maps $f$ and $g$, the following equality holds:      
$$ 
\infer{A_1 \times A_2 \xrightarrow[\mathit{R[f]}]{} A_1}{A_1 \xrightarrow{f} A_2} 
\qquad 
\infer{A_1 \times A_3 \xrightarrow[\mathit{R[g]}]{} A_1}{A_1 \xrightarrow{g} A_3} 
\qquad 
\infer{A_1 \times (A_2\times A_3)  \xrightarrow[\mathit{{R[\<f,g\>]}}]{} A_1}{A_1 \xrightarrow{\<f,g\>} A_2\times A_3} $$

$$
R[\<f,g\>] = (1\times \pi_0)R[f]+ (1\times \pi_1)R[g]
$$
While for the unique map to the terminal object: \( !_{A} : A \to 1 \), the following equality holds:
\[
R[!_{A}] = 0
\]

\noindent {\bf [Reverse Chain Rule]} For composable maps \(f\) and \(g\), the following diagram commutes:
$$ 
\frac{A_1 \xrightarrow{f} A_2}{A_1 \times A_2 \xrightarrow[\mathit{R[f]}]{} A_1} 
\qquad 
\frac{A_2 \xrightarrow{g} A_3}{A_2 \times A_3 \xrightarrow[\mathit{R[g]}]{} A_2} 
\qquad  
\frac{A_1 \xrightarrow{\<f,g\>} A_3}{A_1 \times A_3 \xrightarrow[\mathit{R[\<f,g\>]}]{} A_1}
$$

$$
\begin{tikzcd}
A_1 \times A_3 \arrow[r, "{R[f \circ g]}"] \arrow[d, "{\langle \pi_0, \langle \pi_0 f, \pi_1 \rangle \rangle}"'] & A_1                                 \\
A_1 \times (A_2 \times A_3) \arrow[r, "{1_{A_1} \times R[g]}"']                                                  & A_1 \times A_2 \arrow[u, "{R[f]}"']
\end{tikzcd}
$$

\noindent {\bf[Linearity on the Reverse Derivative]}
$ \<1\times \pi_0,0\times \pi_1\>(\iota_0 \times 1)R[R[R[f]]]\pi_1 = (1\times \pi_1)R[f] $  

\noindent {\bf [Symmetry of mixed partials]} 
$(\iota_0 \times 1)R[R[(\iota_0 \times 1)R[R[f]]\pi_1]]\pi_1  = \mathsf{ex} (\iota_0 \times 1)R[R[(\iota_0 \times 1)R[R[f]]\pi_1]]\pi_1$, 
where $\iota_0 := \<1,0\>:A \to A\times B$, and $\mathsf{ex}$ is the \textbf{exchange} natural isomorphism defined as $\mathsf{ex} := \<\pi_0 \times \pi_0, \pi_1\times \pi_1\>$. 
\end{Definition}

The \(\Lens\) structure integrates seamlessly with the reverse differential combinator \({R}\) of CRDC. Specifically, the pair \((f, R[f])\) forms a morphism in \(\Lens(\Ca)\) when \(\Ca\) is a CRDC. In the context of learning, \({R}[f]\) functions as a backward map, enabling the "learning" of \(f\). The type assignment \(A \times B \to A\) for \({R}[f]\) conceals essential distinctions: \({R}[f]\) takes a tangent vector (corresponding to the gradient descent algorithm) at \(B\) and outputs one at \(A\). Since on the \({R}[f]\) side, both outputs and inputs are different from those on the \(f\) side, the diagram representation is revised as follows.

$$
\scalebox{0.6}{\includegraphics{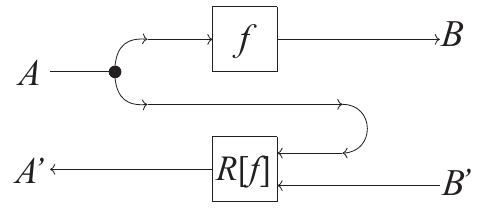}}
$$

The fundamental category where gradient-based learning takes place is the composite $\Para(\Lens(\Ca))$ of the $\Para$ and $\Lens$ constructions given the base CRDC $\Ca$. 

\subsection{Composition of Components}

This section highlights principal results from \emph{Deep Learning with Parametric Lenses}\cite{cruttwell2024deep}, which describes parametric lenses as homogeneous components functioning in the gradient-based learning process.  

\begin{table}[htb]
  \vspace{-15pt}
  \caption{Summary of key components at work in the learning process: parametric lenses\cite{cruttwell2024deep}.}\label{tab:component}
  \vskip 0.02in
  \centering
  \begin{small}
  \renewcommand{\multirowsetup}{\centering}
  \setlength{\tabcolsep}{6pt}
  \begin{tabular}{p{2.5cm}|c|c}
   \toprule
   \textbf{Component} & \textbf{Pictorially definition} & \textbf{Categorical Construction} \\
    \toprule 
    \multirow{2}{*}[-5ex]{Model} & \multirow{2}{*}{} & \multirow{2}{*}[-1.5ex]{$(P, f): \Para(\Lens(\Ca))(X, Y)$} \\
    & \scalebox{0.8}{\includegraphics{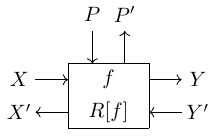}} & \scalebox{1.2}{{$\diset{f}{R[f]}: \diset{P}{P'} \times \diset{X}{X'} \to \diset{Y}{Y'}$}} \\
    \midrule
    \multirow{2}{*}[-5ex]{Loss map} & \multirow{2}{*}{} & \multirow{2}{*}[-1.5ex]{$(Y,\mathsf{loss}): \Para(\Lens(\Ca))(Y, L)$} \\
    & \scalebox{0.8}{\includegraphics{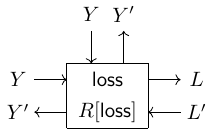}} & \scalebox{1.2}{{$\diset{\mathsf{loss}}{R[\mathsf{loss}]}: \diset{Y}{Y'} \times \diset{Y}{Y'} \to \diset{L}{L'}$}} \\
    \midrule
   \multirow{4}{*}[-12ex]{Optimiser} & {Gradient descent} & {\multirow{2}{*}[-1.5ex]{$G: \Lens(\Ca)(P, P)$}} \\
    & \scalebox{0.8}{\includegraphics{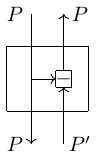}} & \raisebox{5ex}{\scalebox{1.2}{$\diset{\id_P}{-_P} : \diset{P}{P} \to \diset{P}{P'}$}} \\
    \cmidrule{2-3} 
    & {Stateful Optimiser} & {\multirow{2}{*}[-1.5ex]{$U: \Lens(\Ca)(S \times P, S \times P)$}} \\
    & \scalebox{0.8}{\includegraphics{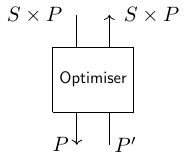}} &  \raisebox{5ex}{\scalebox{1.2}{$\diset{U}{U^*}: \diset{S \times P}{S \times P} \to \diset{P}{P'}$}} \\
    \midrule
   \multirow{2}{*}[-2.5ex]{Learning rate} & \multirow{2}{*}{} & \multirow{2}{*}[-0.5ex]{$\alpha: \Lens(\Ca)((L, L'), (1,1))$} \\
   & \raisebox{3ex}{\scalebox{0.8}{\includegraphics{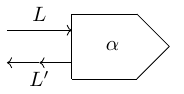}}} & \raisebox{1ex}{\scalebox{1.2}{$\alpha: \diset{L}{L'} \to \diset{1}{1}$}} \\
    \midrule
    \multirow{2}{*}[-5ex]{Corner} & \multirow{2}{*}{} & \multirow{2}{*}[-0.5ex]{$(X,\eta):\Lens(\Ca)(1, X)$}  \\
    & \raisebox{0ex}{\scalebox{0.8}{\includegraphics{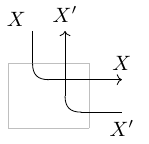}}} & \raisebox{1ex}{\scalebox{1.2}{$\diset{\id_{X}}{\pi_{1}}: \diset{X}{X'} \times \diset{1}{1} \to \diset{X}{X'}$}} \\
    \bottomrule
    \end{tabular}
    \end{small}
\end{table}

For the most basic situation, they discussed a typical supervised learning scenario and its categorization. It involves finding a parametrized \textbf{model} \(f: P \times X \to Y\) with parameters \(p \in P\), which are updated step by step until certain requirements are met. The gradient-based updating algorithm, referred to as the \textbf{optimizer}, updates these parameters iteratively based on a \textbf{loss map} and is controlled by a \textbf{learning rate} \(\alpha\). The authors of \cite{cruttwell2024deep} emphasize that each component, including the model, loss map, optimizer, and learning rate, can vary independently but are uniformly formalized as parametric lenses. Their pictorial definitions and types are summarized in \autoref{tab:component}. In this table, \(\PolyZ\) is introduced as the base category for the learning of digital circuits (as inputs) involved in the same framework \cite{wilson2023category}.

In \cite{cruttwell2024deep}, a variety of models (such as linear-bias-activation neural networks, Boolean circuits \cite{wilson2021reverse}), and polynomial circuits), loss functions (including mean squared error and Softmax cross-entropy), gradient update algorithms (including Nesterov momentum, Adagrad, and ADAM), and even multiple learning variations (including parameter learning, deep dreaming, and unsupervised learning) are covered by the comprehensive framework of parametric lenses. 
The pictorial definitions of the parametric lenses imply that all these examples are homogeneous components featuring three kinds of interfaces: inputs, outputs, and parameters. These features form the basis of the component-based approach, wherein components are composed along their interfaces to construct a gradient-based learning process. 

\begin{figure}[htp]
\centering
\begin{subfigure}{.25\textwidth}
   \centering
  {\scalefont{0.5}
    \includegraphics[width=\textwidth]{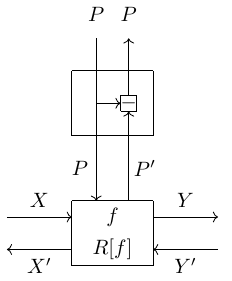}
  }
\end{subfigure}%
\begin{subfigure}{.55\textwidth}
  \centering
  {\scalefont{0.8}
    \includegraphics[width=\textwidth]{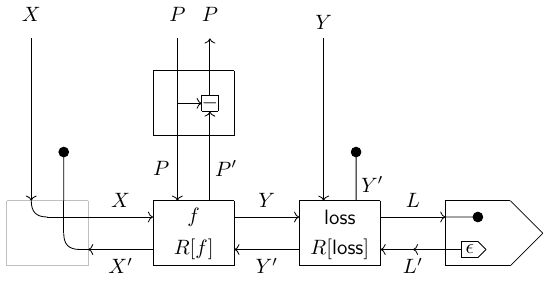}
  }
\end{subfigure}
\caption{Model reparameterised by basic gradient descent (left, adapted from Figure 4 in \cite{cruttwell2024deep}) and a
 full picture of an end-to-end supervised learning process (right, adapted from Equation 5.1 in \cite{cruttwell2024deep}).}
\label{fig:model_and_optimiser}
\end{figure}

\begin{figure}[htp]
    \centering
    {\scalefont{0.75}
    \includegraphics[width=0.65\textwidth]{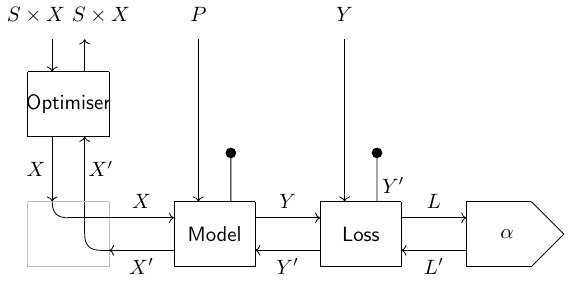}
    }
    \caption{Deep dreaming as a parametric lens, adapted from Equation 4.3 in \cite{cruttwell2024deep}.}
    \label{fig:deep_dreaming}
\end{figure}

Here, we pick up two learning processes from their work (depicted in \autoref{fig:model_and_optimiser} on the right side and \autoref{fig:deep_dreaming}) to briefly introduce how to compose the components. The composition of components can be seen as a "plugging" operation for graphical descriptions of lenses. Specifically, a model \((f, R[f]): (P, P') \times (X, X') \to (Y, Y')\) represents a typical parametric lens. Additionally, the gradient descent optimizer, categorically represented by \(G: (P, P) \to (P, P')\), is a lens that facilitates the required change in parameters by subtracting a value from the current parameters. By ``plugging'' the lens \(G\) above the model \((f, R[f])\) through their ``interface'' \((P, P')\), the parametric morphism \((f, R[f])\) is explicitly given with reparameterization by the parametric lens \(G\).
As a result, the parameters are reparameterized by the specified optimizers, while the backward wires \(X'\) on the input side and \(Y'\) on the output side are discarded.
As aforementioned, the ``plugging process'' is a lens composition, resulting in another parametric lens with type $(P,P') \times (X, X') \to (Y, Y')$, as illustrated in \autoref{fig:model_and_optimiser} (left). 

In this manner, the authors successfully integrate the components listed in \autoref{tab:component} to assemble a gradient-based learning system for parameter learning in a supervised setting (see \autoref{fig:model_and_optimiser} on the right), as well as the categorical framework of a DeepDream learning system (see \autoref{fig:deep_dreaming}). For the former, when fixing the optimizer \((U, U^*): (S \times P, S \times P) \to (P, P')\), the system learns a morphism in \(\Para(\Lens(\Ca))\) from \((1,1)\) to \((1,1)\), with its parameter space being \((A \times S \times P \times B, S \times P)\). In other words, this is a lens of type \((A \times S \times P \times B, S \times P) \to (1,1)\), where its ``get'' is the identity morphism. Its ``put'' is \(A \times S \times P \times B \to S \times P\). 
The following formulas can summarize the learning process.
\[ put(a, s, p, b_t) = U^*(s, p, p') \]
where
\begin{itemize}
    \setlength{\itemsep}{0pt}
    \setlength{\parskip}{0pt}

    \item[] \(\hat{p} = U(s, p)\)
    \begin{itemize}
        \setlength{\itemsep}{0pt}
        \setlength{\parskip}{0pt}
        \item Updates of parameters for the network provided by the state and step size
    \end{itemize}
    
    \item[] \(b_p = f(\hat{p}, a)\)
    \begin{itemize}
        \setlength{\itemsep}{0pt}
        \setlength{\parskip}{0pt}
        \item Predicted output of the network
    \end{itemize}
    
    \item[] \((b'_t, b'_p) = R[\mathsf{loss}](b_t, b_p, \alpha(\mathsf{loss}(b_t, b_p)))\)
    \begin{itemize}
        \setlength{\itemsep}{0pt}
        \setlength{\parskip}{0pt}
        \item Difference between prediction and true value
    \end{itemize}
    
    \item[] \((p', a') = R[f](\hat{p}, a, b'_p)\)
    \begin{itemize}
        \setlength{\itemsep}{0pt}
        \setlength{\parskip}{0pt}
        \item Updates of parameters and input
    \end{itemize}
\end{itemize}

The latter, DeepDream, uses the parameters \( p \) of a well-trained classifier network to generate the dream or amplify certain features and shapes of a type \( b \) on the input \( a \). This allows the network to dream specific features and shapes on a selected image. The objective of the corresponding categorical framework is to describe how to connect the gradient descent lens to the input end, providing a method to enhance the input image to elicit a specific interpretation. The system learns a morphism in $\mathbf{Para}(\Lens(\Ca))$ from \((1,1)\) to \((1,1)\), with its parameter space being \((S \times A \times P \times B, S \times A)\). In other words, this is a lens of type \((S \times A \times P \times B, S \times A) \to (1,1)\), where its ``get'' is trivial. Its ``put'' is \(S \times A \times P \times B \to S \times A\). The learning process can be summarized by the following formulas.
\[ put(s, a, p, b_t) = U^*(s, a, a') \]
where
\begin{itemize}
    \setlength{\itemsep}{0pt}
    \setlength{\parskip}{0pt}

    \item[] \(\hat{a} = U(s, a)\)
    \begin{itemize}
        \setlength{\itemsep}{0pt}
        \setlength{\parskip}{0pt}
        \item The updated input provided by the state and step size
    \end{itemize}

    \item[] \(b_p = f(p, \hat{a})\)
    \begin{itemize}
        \setlength{\itemsep}{0pt}
        \setlength{\parskip}{0pt}
        \item Prediction of the network
    \end{itemize}

    \item[] \((b'_t, b'_p) = R[\mathsf{loss}](b_t, b_p, \alpha(\mathsf{loss}(b_t, b_p)))\)
    \begin{itemize}
        \setlength{\itemsep}{0pt}
        \setlength{\parskip}{0pt}
        \item Changes in prediction and true value
    \end{itemize}

    \item[] \((p', a') = R[f](p, \hat{a}, b'_p)\)
    \begin{itemize}
        \setlength{\itemsep}{0pt}
        \setlength{\parskip}{0pt}
        \item Changes in parameters and input
    \end{itemize}
\end{itemize}

Notably, the authors also illustrated the component-based approach that covers the iterative process, which is a parametric morphism. 
The key insight is that the morphism $\putt{} : X \times P \times Y \to P$ is fundamentally an endomorphism $P \to P$ with additional inputs $X \times Y$. Thus, the ``put" functions as a parametric morphism $(X \times Y, put) : \mathbf{Para}(\Ca)(P, P)$. As an endomorphism, it can compose itself, producing a composite endomorphism that handles two ``parameters'': the input-output pairs at time steps $0$ and $1$. This iterative process is streamlined by $\mathbf{Para}$ composition, which manages the algebra of iteration, like (Figure in \cite{cruttwell2022categorical}, page 20)

$$
\begin{tikzpicture}[node distance=2.5cm, auto]

\node (P1) at (0, 0) {$P$};
\node (Put1) [draw, rectangle, right of=P1] {put};
\node (XY1) [above of=Put1, yshift=-1cm] {$X \times Y$};

\node (Put2) [draw, rectangle, right of=Put1] {put};
\node (XY2) [above of=Put2, yshift=-1cm] {$X \times Y$};
\node (dots) [right of=Put2] {$\dots$};

\node (Put3) [draw, rectangle, right of=dots] {put};
\node (XY3) [above of=Put3, yshift=-1cm] {$X \times Y$};
\node (P3) [right of=Put3] {$P$};

\draw[->] (P1) -- (Put1);
\draw[->] (Put1) -- node[above] {$P$}(Put2);
\draw[->] (XY1) -- (Put1);

\draw[->] (Put2) -- node[above] {$P$}(dots);
\draw[->] (XY2) -- (Put2);

\draw[->] (dots) -- node[above] {$P$}(Put3);
\draw[->] (Put3) -- (P3);
\draw[->] (XY3) -- (Put3);

\end{tikzpicture}
$$
\subsection{Other Related Research}
\label{other_re}
The aforementioned categorical framework for gradient-based learning was first introduced by Cruttwell et al. in 2021 \cite{cruttwell2022categorical} and subsequently refined through a series of works \cite{cruttwell2022monoidal, cruttwell2023reverse, cruttwell2024deep}. Their work was inspired by earlier pioneering efforts in the same direction, such as those by Fong et al. \cite{fong2019backprop, fong2019lenses}. The work of Fong et al. is particularly noteworthy because it provided the initial developmental directions, beginning with the establishment of a simple model of supervised learning and then considering probabilistic learning behaviors, as many input data sets are obtained under conditions of uncertainty and randomness \cite{fong2013causal}. The category of stochastic causal models was initially developed independently of gradient-based learning but was later incorporated into the categorical framework of learning \cite{gavranovic2024fundamental}. Especially, they also introduced methodologies based on graphical calculus to this field.

Following \cite{cruttwell2022categorical}, Cruttwell et al. focused on applying the framework, particularly the composition of basic components in specific learning processes, and investigating the semantics of learning. 
Their work demonstrated that this framework offers benefits such as abstraction and uniformity, enhanced by its compositional nature. Abstraction is achieved by encompassing a wide range of optimizers, models, loss functions, and more as instances of the framework. Uniformity is realized by reducing all components involved in the learning process to a single concept: \textbf{parametric lenses}. Because their compositions are functorial, parametric lenses are expressed using graphical notation with ports and interfaces, which gives rise to the ``open (externally connectable and integrable)'' properties of these composition-derived systems, which can also be analogized to the ``open petri-net'' system. 

In \cite{cruttwell2022categorical, cruttwell2024deep}, the authors additionally provide a proof-of-concept implementation through a Python library \cite{Anonymous2022Optics}, which focuses on using lenses and reverse derivatives to construct and train neural networks. This gradient-based learning is implemented via parametric lens composition, fundamentally differing from the approaches taken by Keras and other existing libraries. The authors implemented end-to-end learning models for the MNIST and Iris image classification tasks, achieving performance comparable to models built with Keras, thereby empirically demonstrating the correctness of their library. Building upon \cite{cruttwell2022categorical, cruttwell2024deep}, data-parallel algorithms have been explored in string diagrams which efficiently compute symbolic representations of gradient-based learners based on reverse derivatives \cite{wilson2023data}. Based on these efficient algorithms, several Python implementations \cite{Wilson2023Yarrow, Wilson2023polycirc, Wilson2024catgrad} have been released. These implementations are characterized by their speed, data-parallel capabilities, and GPU compatibility; they require minimal primitives/dependencies and are straightforward to implement correctly. 

An extension of lenses is ``optics''. They correspond to the case where learning processes do not involve the ``differentiation'', thus not requiring the Cartesian property of the background category. While the base category still needs to permit parallel composition of processes, it then degenerates to a monoidal structure\cite{fong2019lenses, gavranovic2024fundamental}. 

In \cite{spivak2021functorial}, the authors discuss aggregating data in a database using optics structures. They also consider the category Poly, which is closely related to machine learning in (analog) circuit designs\cite{wilson2023category, ghica2022fully}. Another work, \cite{videla2022lenses}, demonstrates that servers with composability can also be abstracted using the lenses structure.

In \cite{gavranovic2022space}, the higher categorical (2-categorical) properties of optics are specifically mentioned for programming their internal setups. Specifically, the internal configuration of optics is described by the 2-cells. From another viewpoint, the 2-cells encapsulate homotopical properties, which are often considered as the internal semantics of a categorical system. This aligns with the topos-based research of \cite{belfiore2021topos}. Some contents were also mentioned in \cite{spivak2022learn}.

In \cite{capucci2022diegetic}, game theory factors are also integrated into this categorical framework, focusing more on the feedback mechanism rather than the learning algorithm. We consider combining their ideas with reinforcement learning \cite{hedges2024reinforcement} (see the next paragraph) as an interesting direction, with existing instances provided in \cite{lanctot2019openspiel}.

A critical example is that, instead of lenses, optics are used for composing the Bayesian learning process \cite{kamiya2021categorya}. We will introduce this work in detail in the next section. Because decisions are made based on policies and Bayesian inference, they introduced the concept of ``actegory'' to express actions on categories, which is also related to the monoidal structure(Cartesian categories are a specific type of monoidal category with additional structures, such as symmetric tensor products and projection maps). Moreover, a similar structure is employed in \cite{hedges2024reinforcement}, where the component-compositional framework is expanded to reinforcement learning. They demonstrated how to fit several major algorithms of reinforcement learning into the framework of categorical parametrized bidirectional processes, where the available actions of a reinforcement learning agent depend on the current state of the Markov chain.
In \cite{gavranovicposition}, the authors considered how to specify constraints in a top-down manner and implementations in a bottom-up manner together. This approach introduces algebraic structures to play the role of ``action'', or more precisely, the invariance or equivariance under action. In this context, they chose to introduce the monad structure and employ monad algebra homomorphisms to describe equivariance. An important instance of their framework is geometric deep learning, where the main objective is to find neural network layers that are monad algebra homomorphisms of monads associated with group actions. It is worth noting that this structure is further related to the ``sheaf action'' in \cite{belfiore2021topos}.

When this framework is further expanded to learning on manifolds, where a point has additional data (in its tangent space), reverse differential categories considering only the data itself are not sufficient. \cite{cruttwell2023reverse} extended the base category to reverse tangent categories, i.e., categories with an involution operation for their differential bundles.

Other research applying category theory to gradient-based learning includes \cite{vakar2022chad}. This work primarily serves programming languages with expressive features. The authors built their results on the original automatic differentiation algorithm, considering it as a homomorphic (structure-preserving) functor from the syntax of its source programming language to the syntax of its target language. This approach is extensible to higher-order primitives, such as map-operations over constructive data structures. Consequently, they can use automatic differentiation to build instances such as differential and algebraic equation solvers.

Some studies, such as \cite{wilson2023data, abbott2024neural, abbott2024functor}, have also provided comprehensive ideas on the compositional, graphical, and symbolic properties of categorical learning in neural circuit diagrams and deep learning. Their work also discussed related semantics.
 
\section{Developments in Probability-based Learning}
Probability, which quantifies the likelihood of events with values between 0 and 1, is fundamental for understanding machine learning and is utilized in most machine learning methodologies, especially in areas like probabilistic modeling, Bayesian inference, and generative models. For example, given a fixed dataset, many machine learning problems can be formulated as optimization problems, where the objective is to minimize a loss function. This formulation requires careful reasoning about the origins and limitations of the dataset, such as biases, generalization capacity, and data quality, to effectively solve the problem.
In probabilistic approaches, stochastic uncertainty is modeled using probability theory and Bayesian inference. This reframes supervised learning from a deterministic function approximation problem into a distribution approximation problem, where the goal is to estimate distributions over functions or outputs.
From a categorical perspective, stochastic behavior is often analyzed within a specific category, such as a \textbf{Markov category} or \textbf{a category of stochastic maps}(such as $\mathsf{Stoch}$ and $\mathsf{FinStoch}$). In these frameworks, morphisms represent probabilistic transitions or stochastic kernels, capturing how uncertainty evolves or diminishes over time, such as in time series analysis or Markov processes.

Here is an overview of key probability types relevant to ML:
\begin{itemize}
    \item Empirical Probability: Calculated by dividing the number of times an event occurs by the total number of observations. In category theory, empirical probability is typically addressed within three structures. The first structure views the distribution as a functor, where empirical probability is represented as an empirical measure functor \( \delta \), mapping a set of observations \( A \) to the set of probability distributions \( \Delta(A) \) over \( A \). The second structure extends the first by employing the \textbf{Giry monad}, which captures empirical probability as a finite measure. The monad structure encapsulates the functorial process of mapping sets to measures. The third structure situates empirical probability within the context of \textbf{measure-theoretic probability}, involving categories analogous to \( \mathsf{Meas} \), the category of measurable spaces and measurable functions.
    \item Theoretical Probability: Found by dividing the ways an event can happen by all possible outcomes. In category theory, theoretical probability can be addressed using the \textbf{Giry monad} or categories analogous to \( \mathsf{Meas} \), the category of measurable spaces and measurable functions. It can also be analyzed through \textbf{monoidal categories} and \textbf{functors}, as monoidal categories equipped with tensor products provide a framework to describe how independent probability distributions combine and how random variables can be translated into morphisms between probabilistic spaces.
    \item Joint Probability: Measures the likelihood of two events occurring together, expressed as \(P(A \cap B) = P(A) \times P(B)\), where \(P(A \cap B)\) is the probability of both events A and B occurring. In category theory, joint probability can be modeled using the copying structure in Markov categories, which captures probabilistic independence and facilitates reasoning about joint distributions. Alternatively, it can also be described within the tensor product space of general monoidal categories, which models the combination of independent spaces.
    \item Conditional Probability: The probability of event A given event B has already occurred, noted as \(P(A|B) = \frac{P(A \cap B)}{P(B)}\). In contrast, \(P(B|A) = \frac{P(A \cap B)}{P(A)}\). This concept can extend to express the joint probability as \(P(A \cap B) = P(A) \times P(B|A)\), linking the occurrence of both events. In Markov categories, conditional probability is interpreted through stochastic maps, with restrictions to subsets of outcomes enabled by the category's copying structure. In categories of measurable spaces, such as $\mathsf{Meas}$, conditional probability can be represented using pullbacks, which restrict or select parts of the joint distribution morphism. Alternatively, it can be derived through disintegration, which decomposes a joint distribution into its marginal and conditional components.
    \item Bayes' Theorem: It is fundamental concept in probability theory that describes how to update the probability of a hypothesis based on new evidence, expressed as: \( P(A|B) = \frac{P(B|A) \times P(A)}{P(B)}\), where \( P(A|B) \) is the probability of hypothesis \( A \) given the evidence \( B \); \( P(B|A) \) is the probability of observing the evidence \( B \) given that hypothesis \( A \) is true; \( P(A) \) is the prior probability of hypothesis \( A \), which is the initial degree of belief in \( A \); and \( P(B) \) is the probability of the evidence \( B \) under all possible hypotheses.
\end{itemize}

On the other hand, statistics forms a foundational element of machine learning, providing essential tools for extracting insights from data. This field of applied mathematics focuses on the collection, analysis, interpretation, and presentation of empirical data, thereby facilitating informed decisions in business and investment.

The main statistical elements employed in machine learning are divided into two primary branches.
\begin{itemize}
    \item Descriptive Statistics: Summarizes data features, often through visualizations, providing a snapshot of data points' distribution and trends.
    \item Inferential Statistics: Uses sample data to make inferences or predictions about a larger population, helping in hypothesis testing and model predictions.
\end{itemize}
In machine learning, statistical methods are crucial for understanding training data and evaluating machine learning models' performance. This understanding helps to improve decision making processes in business and technology applications. Especially, the applications in ML include: Poisson processes, martingales, probability metrics, empirical processes, VC theory, large deviations, the exponential family, and simulation techniques like Markov Chain Monte Carlo. We specifically highlight that the field of algebraic geometry is also closely related to statistical methods in machine learning \cite{HAUENSTEIN202393}.

\subsection{Categorical Background of Probability and Statistics Learning}
In machine learning, where preset datasets are often employed to formulate optimization problems, effectively solving these problems requires a comprehensive understanding of the data's origin and limitations to solve optimization problems effectively. Thus, a thorough understanding of the data distribution is essential.Random uncertainties in machine learning models can be accurately captured using probability theory and Bayesian inference, enabling a transition from function approximation to distribution approximation in supervised learning. This transition is crucial, as it incorporates the concept of inherent variability—known as aleatoric uncertainty—that cannot be mitigated simply by increasing the amount of data. 
From this perspective, category theory serves as a powerful tool for analyzing, interpreting, and reasoning randomness within certain categories or models by defining a compositional framework for the models and simultaneously bridging probability, statistics, and information theory, thereby laying the foundation for more resilient and broadly applicable learning frameworks.

A key milestone in this area is the formal categorization of the Bayesian learning framework. 
Bayesian learning revolves around combining prior beliefs with observational data to refine the understanding of model parameters. The prior distribution encodes knowledge about the parameters before any data is observed, while the posterior distribution integrates this prior information with new observations, offering an updated and more informed perspective. Bayes' rule facilitates this process by formally deriving the posterior distribution from the prior and likelihood, ensuring that parameter estimates align systematically with observed data. Bayesian methods are particularly well-suited for handling uncertainty, as they not only provide optimal parameter estimates but also quantify the uncertainty through the posterior distribution, encapsulating both the variability in the data and the inherent randomness of the system. This makes Bayesian learning a powerful framework for developing robust and adaptive models. 
In fact, the categorization of the Bayesian learning framework builds upon earlier categorizations of probability and statistical methodologies, which predate its application in machine learning. Category-theoretic approaches to probability theory provide an abstract and unifying framework, which can be broadly summarized as follows:

\begin{itemize}
    \item \textbf{Categorization of existing structures in traditional probability theory}: Core constructs such as probability spaces and integration can be categorized using structures like the \textbf{Giry monad}, which maps measurable spaces to the space of probability measures on them, preserving the measurable structure. These frameworks formalize the relationships between spaces and their probability measures, enabling a compositional view of probabilistic models. 
    \item \textbf{Categorization of synthetic probability and statistical concepts}: Certain axioms and structures are regarded as ``fundamental'' or ``essential'' within probabilistic logic frameworks, which derive inference processes. Measure-theoretic models, in this context, serve as concrete instances or semantics of these abstract frameworks. Markov categories are frequently employed in this direction, offering a formal representation of stochastic maps, conditional probabilities, and compositional reasoning within probabilistic systems. 
\end{itemize}

In the categorization of existing structures, early categorical frameworks were developed to formalize probability measures. Pioneering work by Lawvere \cite{lawvere1962category} and Giry \cite{giry2006categorical} first introduced a categorical perspective on probability measures. In \cite{leinster2012codensity}, the focus was placed on the resulting \textbf{ultrafilter monad}, a structure derived when the probability monad is induced from a functor without an adjoint, known as the \textbf{codensity monad} of the functor. They demonstrated that an ultrafilter monad on a set can be interpreted as a functional mapping subsets of the set to the two-element set, satisfying the properties of a finitely additive probability measure. Subsequently, \cite{sturtz2014categorical} generalized this framework using the ultrafilter concept, showing that such a probability measure can be described as a weakly averaging, affine measurable functional mapping into \([0,1]\). Moreover, the probability measures on a space were shown to constitute the elements of a submonad of the Giry monad.
However, when working with real-world datasets, scenarios where discreteness plays a critical role take on deeper significance. As demonstrated in \cite{Belle2021ProbabilityMA}, the Giry monad can be restricted to countable measurable spaces equipped with discrete $\sigma$-algebras, resulting in a restricted Giry functor derived from the codensity monad. This observation suggests that the natural numbers \(\mathbb{N}\) are, in some sense, `sufficient' for these applications. In this context, standard measurable spaces are replaced by \textbf{Polish spaces}, a class of spaces that has yet to be thoroughly explored in the domain of categorical probability-related machine learning.

The Giry monad structure has been used in many applications, such as in \cite{burroni2009distributive}, where stochastic automata are built as algebras on the monoid and Giry monads. In this survey, we intend to emphasize the categorical approach to Bayesian Networks \cite{fong2013causal}, Bayesian reasoning, and inference \cite{Culbertson2013BayesianML,culbertson2014categorical} using the Giry monad. The details will be provided in the next subsection.

For the categorization of synthetic probability and statistics concepts, Markov categories, which depict ``randomness'' using random functions, provide a unique synthetic approach to understanding probability. The earliest research on Markov categories dates back to Lawvere \cite{lawvere1962category} and Chentsov \cite{chentsov1965categories}. Related studies include early research on synthetics \cite{Giry1982ACA, golubtsov1999axiomatic, golubtsov2002monoidal} and the specialization of Markov categories with restrictions by Kallenberg \cite{kallenberg2017random}, Fritz \cite{Fritz2018BimonoidalSO,Fritz2019ASA,fritz2020infinite,fritz2020synthetic,fritz2023representable,Fritz2023WeaklyMC}, Sabok \cite{sabok2021probabilistic}, and others. One of the most convenient advantages of Markov categories is that they facilitate graphical calculus, making ideas expressed through diagram operations easy to translate into programming languages \cite{kamiya2021categorya, moggi1988computational, sabok2021probabilistic}.

The application of category theory in probabilistic machine learning aims to elucidate properties and facilitate data updating, particularly in the context of probability distributions and uncertainty inference. This involves foundational steps such as formalizing random variables as fundamental entities, modeling learning processes within a categorical framework, and establishing general principles for probabilistic reasoning. For example, random variables can be treated as morphisms in a suitable category, and learning processes can be modeled as functors between categories of data and probabilistic models.
Bayesian machine learning leverages categorical methods to construct models for both parametric and nonparametric Bayesian reasoning on function spaces. This includes representing priors, likelihoods, and posteriors using categorical structures. Such frameworks enable compositional reasoning, where complex probabilistic updates are modeled compositionally. This abstraction is particularly effective in supervised learning, where categorical Bayesian probability provides a formal structure for incorporating prior knowledge and updating beliefs based on observed data.
The categorical approach to probabilistic modeling offers practical advantages, such as simplifying proofs, complementing measure-theoretic methods, and intuitively representing complex probabilistic relationships. For instance, Markov categories provide a framework for reasoning about stochastic processes, conditional probability, and independence in a compositional manner. 
Research in this area includes contributions by \cite{fong2013causal, kamiya2021categorya, shiebler2021categorya, Culbertson2013BayesianML, culbertson2014categorical}, among others. Experimental studies, such as \cite{Sennesh2023ComputingWC}, demonstrate the implementation of categorical theories in automatic inference systems \cite{ambrogioni2021automatic, Meulen2020AutomaticBF, Braithwaite2022DependentBL}, showcasing the practical utility of these frameworks in machine learning applications.

\subsection{Preliminaries and Notions}
In this section, we provide the main preliminaries and notions relevant to the related research.

\begin{Definition}[Measurable space~\cite{bogachev2007measure}]
\label{def:Measurable_Space}
A \textbf{\(\sigma\)-algebra} \(\mathcal{A}\) on a set \(X\) is a collection of subsets of $X$ that satisfies the following properties:
\begin{enumerate}
    \item The empty set \(\emptyset\) is in \(\mathcal{A}\).
    \item If $A \in \mathcal{A}$, then $A^c$ (the complement of $A$ in $X$) is also in \(\mathcal{A}\).
    \item If \(\{A_i\}_{i \in I}\) is a countable collection of sets in \(\mathcal{A}\), then $\bigcup_{i \in I} A_i \in \mathcal{A}$.
\end{enumerate}
A \textbf{measurable space} \((X, \mathcal{A})\) consists of a set \(X\) and a \(\sigma\)-algebra \(\mathcal{A}\) of subsets of \(X\).
A function \(f: (X, \mathcal{A}) \to (Y, \mathcal{B})\) between two measurable spaces is called measurable if for every set \(B \in \mathcal{B}\), the preimage \(f^{-1}(B) \in \mathcal{A}\).
Measureable spaces form the category of measurable spaces $\mathsf{Meas}$ with measurable functions as morphisms.
\end{Definition}

\begin{Definition}[Measure space~\cite{bogachev2007measure}] A \textbf{measure space} \((X, \mathcal{A}, \mu)\) consists of:
\begin{itemize}
    \item A set \(X\),
    \item A \(\sigma\)-algebra \(\mathcal{A}\) of subsets of \(X\),
    \item A measure \(\mu\), which is a function \(\mu: \mathcal{A} \to [0, \infty]\) satisfying the following properties:
    \begin{enumerate}
        \item \(\mu(\emptyset) = 0\),
        \item \textbf{Countable Additivity}: For any countable collection of pairwise disjoint sets \(\{A_i\}_{i \in I} \subseteq \mathcal{A}\),
        \[
        \mu\left(\bigcup_{i \in I} A_i\right) = \sum_{i \in I} \mu(A_i).
        \]
    \end{enumerate}
\end{itemize}
A \textbf{probability space} is a measure space where the measure of the entire set is normalized to one, i.e., \(\mu(X) = 1\).
\end{Definition}

When dealing with probabilistic learning algorithms, the distributions involved are not arbitrary but typically originate from a fixed global sample space.
In other words, the focus is primarily on random variables defined over a predefined sample space, rather than on arbitrary probability measures.
In probabilistic programming, reasoning about the space of probabilistic maps within a given probabilistic category often requires the categorical property of being \textit{Cartesian closed}. However, $\mathsf{Meas}$ is not Cartesian closed because the space of measurable functions between measurable spaces is not always measurable, i.e., the $\sigma$-algebra on the space of functions does not naturally align with the $\sigma$-algebras of the domain and codomain. 
To address this limitation, researchers have proposed two main approaches. The first is to $\mathsf{Meas}$ with various monoidal products to support probabilistic reasoning and compositionality within the existing framework \cite{Culbertson2013BayesianML}. The second approach is to generalize measurable spaces to categories that are inherently Cartesian closed, such as \textbf{quasi-Borel spaces (QBS)} \cite{heunen2017convenient}.

\begin{Definition}[Quasi-Borel space~\cite{heunen2017convenient}]
A \textbf{Quasi-Borel space} is a pair $(X, \mathcal{M})$ where $X$ is a set and $\mathcal{M}$ is a collection of ``measurable'' maps from some standard Borel space $B$ to $X$, satisfying the following conditions:
\begin{enumerate}
    \item For every $b \in B$, the map $\delta_b : B \to X$ defined by $$\delta_b(b') = \begin{cases}
        x & \text{if } b = b' \\
        \text{undefined} & \text{otherwise}
    \end{cases}$$
    is in $\mathcal{M}$.
    \item If $f, g : B \to X$ are in $\mathcal{M}$, then any measurable function $h : B \to B$ such that $f = g \circ h$ is also in $\mathcal{M}$.
    \item $\mathcal{M}$ contains all constant functions.
\end{enumerate}
\end{Definition}

\begin{Definition}[Stochastic process]
A \textbf{stochastic process} is defined as a collection of random variables defined on a common probability space \((\Omega, \mathcal{F}, \mathbb{P})\), where \(\Omega\) is a sample space, \(\mathcal{F}\) is a \(\sigma\)-algebra, and \(\mathbb{P}\) is a probability measure; and the random variables, indexed by some set \(T\), all take values in the same mathematical space \(S\), which must be measurable with respect to some \(\sigma\)-algebra \( \mathcal{S}\).  
\end{Definition}

\begin{Definition}[Stochastic Process~\cite{kallenberg2002foundations}]
A \textbf{stochastic process} is a collection of random variables \(\{X_t\}_{t \in T}\) defined on a common probability space \((\Omega, \mathcal{F}, \mathbb{P})\), where:
\begin{itemize}
    \item \(\Omega\) is the sample space,
    \item \(\mathcal{F}\) is a \(\sigma\)-algebra of subsets of \(\Omega\),
    \item \(\mathbb{P}\) is a probability measure on \((\Omega, \mathcal{F})\).
\end{itemize}
Each random variable \(X_t\) is indexed by \(t \in T\) and takes values in a measurable space \((S, \mathcal{S})\), where \(S\) is the state space and \(\mathcal{S}\) is a \(\sigma\)-algebra on \(S\).
\end{Definition}

\begin{Definition}[Stochastic map~\cite{kallenberg2002foundations}]
Let \(X\) and \(Y\) be sets. A \textbf{stochastic map (stochastic matrix)} from \(X\) to \(Y\) is a function \(f: X \times Y \to [0,1]\), whose entries (denoted as \(f(y \mid x)\), meaning for a conditional probability of moving to \(y\) giving the condition \(x\)) satisfying:
\begin{enumerate}
    \item For all \(x \in X\) and \(y \in Y\), the number \(f(y \mid x)\) is non-negative.
    \item For all \(x \in X\), the numbers \(f(y \mid x)\) are nonzero for only finitely many \(y \in Y\).
    \item For all \(x \in X\), \(\sum_{y} f(y \mid x) = 1\).
\end{enumerate}
\end{Definition}

A stochastic map provides the transition probabilities between states, whereas a stochastic process uses these transition probabilities to describe the random evolution of a system over time. The stochastic map defines the local probabilistic structure, and the stochastic process describes the global behavior of the system as it evolves.

\begin{Definition}[Stochastic Kernel~\cite{kallenberg2002foundations}]
A \textbf{stochastic kernel} (or \textbf{probability kernel}) from a measurable space \((X, \mathcal{A})\) to another measurable space \((Y, \mathcal{B})\) is a function \(k: X \times \mathcal{B} \to [0, 1]\) satisfying the following conditions:
\begin{enumerate}
    \item \textbf{Probability Measure Condition}: For each fixed \(x \in X\), the function \(k(x, -): \mathcal{B} \to [0, 1]\) is a probability measure on \((Y, \mathcal{B})\).
    \item \textbf{Measurability Condition}: For each fixed \(B \in \mathcal{B}\), the function \(k(-, B): X \to [0, 1]\) is \(\mathcal{A}\)-measurable.
\end{enumerate}
\end{Definition}

Using stochastic kernels and Markov kernels, we can define the following categories commonly used in categorical probabilistic learning research, where these kernels serve as morphisms.

\begin{Definition}[Stochastic Category~\cite{fritz2020markov}]
A \textbf{stochastic category} \(\mathsf{S}\) is a category enriched over the category of measurable spaces \((\mathsf{Meas}, \otimes, \textbf{1})\), where:
\begin{itemize}
    \item \textbf{Objects:} The objects of \(\mathsf{S}\) are measurable spaces.
    \item \textbf{Morphisms:} The morphisms of \(\mathsf{S}\) are stochastic kernels. Specifically, for every pair of objects \(X\) and \(Y\), the hom-set \(\mathsf{S}(X, Y)\) consists of measurable functions \(k: X \times \mathcal{B}_Y \to [0, 1]\) such that:
    \begin{enumerate}
        \item For every fixed \(x \in X\), \(k(x, -)\) is a probability measure on \(Y\),
        \item For every \(B \in \mathcal{B}_Y\), \(k(-, B)\) is a measurable function with respect to \(\mathcal{A}_X\).
    \end{enumerate}
\end{itemize}
The composition of morphisms in \(\mathsf{S}\) corresponds to the integration of kernels.
\end{Definition}

\begin{Definition}[Category \(\mathsf{FinStoch}\)~\cite{fritz2020markov}]
The category \(\mathsf{FinStoch}\) (Finite Stochastic Category) is defined as follows:
\begin{itemize}
    \item \textbf{Objects:} The objects of \(\mathsf{FinStoch}\) are finite sets.
    \item \textbf{Morphisms:} The morphisms between objects are stochastic maps.
    \item \textbf{Composition:} Morphisms compose via matrix multiplication of the corresponding probability matrices.
\end{itemize}
\end{Definition}

\begin{Definition}[Category $\mathsf{ProbStoch}$~\cite{giry2006categorical}]
The category \(\mathsf{ProbStoch}\) (Probabilistic Stochastic Category) is defined as follows:
\begin{itemize}
    \item \textbf{Objects:} The objects of \(\mathsf{ProbStoch}\) are measurable spaces \((X, \mathcal{A})\).
    \item \textbf{Morphisms:} The morphisms between objects \((X, \mathcal{A})\) and \((Y, \mathcal{B})\) are stochastic kernels \(f: X \times \mathcal{B} \to [0,1]\) such that:
    \begin{enumerate}
        \item For each fixed \(x \in X\), the function \(f(x, \cdot): \mathcal{B} \to [0, 1]\) is a probability measure on \((Y, \mathcal{B})\).
        \item For each fixed \(B \in \mathcal{B}\), the function \(f(\cdot, B): X \to [0, 1]\) is \(\mathcal{A}\)-measurable.
    \end{enumerate}
\end{itemize}

When given a base category $\mathsf{C}$, we can further define the category $\mathbf{ProbStoch}(\mathsf{C})$ as a $\mathsf{ProbStoch}$ enriched over $\mathsf{C}$. $\mathbf{ProbStoch}(\mathsf{C})$ satisfies that

\begin{itemize}
    \item The objects are measurable spaces enriched over $\mathsf{C}$.
    \item The morphisms are stochastic kernels that respect the enrichment, i.e., every hom-set $\mathbf{ProbStoch}(\mathsf{C})(X, Y)$ consists of enriched stochastic kernels.
\end{itemize}
\end{Definition}

\begin{Definition}[Category \(\mathsf{BorelStoch}\)~\cite{fritz2020markov}]
Given a topological space \(X\), the \textbf{Borel \(\sigma\)-algebra} \(\mathcal{B}(X)\) is the smallest \(\sigma\)-algebra containing all open sets of \(X\). It is generated by the open sets and includes all sets formed through countable unions, countable intersections, and relative complements of open sets.
A \textbf{Borel space} is a measurable space \((X, \mathcal{B}(X))\), where \(\mathcal{B}(X)\) is the Borel \(\sigma\)-algebra of a topological space \(X\).

\textbf{Category \(\mathsf{BorelStoch}\)}: The category \(\mathsf{BorelStoch}\) (\textbf{Borel Stochastic Category}) is defined as follows:
\begin{itemize}
    \item \textbf{Objects}: The objects of \(\mathsf{BorelStoch}\) are \textit{standard Borel spaces}, which are measurable spaces isomorphic to the Borel \(\sigma\)-algebra of a Polish space (a separable, completely metrizable topological space).
    \item \textbf{Morphisms}: The morphisms of \(\mathsf{BorelStoch}\) are Borel-measurable Markov kernels, i.e., stochastic kernels \(k: X \times \mathcal{B}(Y) \to [0, 1]\) satisfying:
    \begin{enumerate}
        \item For each fixed \(x \in X\), \(k(x, -): \mathcal{B}(Y) \to [0, 1]\) is a probability measure on \(Y\),
        \item For each \(B \in \mathcal{B}(Y)\), \(k(-, B): X \to [0, 1]\) is \(\mathcal{B}(X)\)-measurable.
    \end{enumerate}
\end{itemize}
The composition of morphisms is given by integrating the kernel values, ensuring the compositional structure is preserved.
\end{Definition}

The categorization of probability-based learning processes involves using monads to handle probabilistic measures.

\begin{Definition}[Distribution monad~\cite{Giry1982ACA}]
A \textbf{distribution monad} \((\mathbf{D}, \eta, \mu)\) on a category \(\mathsf{C}\) of measurable spaces consists of the following:
\begin{itemize}
    \item \textbf{Functor:} A covariant functor \(\mathbf{D}: \mathsf{C} \to \mathsf{C}\) that assigns to each object \(X\) in \(\mathsf{C}\) the space \(\mathbf{D}(X)\), the set of all probability measures on \(X\), equipped with the smallest \(\sigma\)-algebra making the evaluation maps \(\mathbf{D}(X) \times X \to [0, 1]\), \((\mu, A) \mapsto \mu(A)\), measurable.
    \item \textbf{Unit Natural Transformation:} A unit natural transformation \(\eta: \text{Id}_{\mathsf{C}} \Rightarrow \mathbf{D}\) such that for each object \(X\) in \(\mathsf{C}\), \(\eta_X: X \to \mathbf{D}(X)\) assigns to each \(x \in X\) the Dirac measure \(\delta_x \in \mathbf{D}(X)\), where \(\delta_x(A) = 1\) if \(x \in A\) and \(\delta_x(A) = 0\) otherwise.
    \item \textbf{Multiplication Natural Transformation:} A multiplication natural transformation \(\mu: \mathbf{D} \circ \mathbf{D} \Rightarrow \mathbf{D}\) such that for each object \(X\), \(\mu_X: \mathbf{D}(\mathbf{D}(X)) \to \mathbf{D}(X)\) satisfies:
    \[
    \mu_X(\nu)(A) = \int_{\mathbf{D}(X)} \mu(A) \, d\nu(\mu),
    \]
    where \(\nu \in \mathbf{D}(\mathbf{D}(X))\) is a measure on the space of probability measures \(\mathbf{D}(X)\), and \(A \subseteq X\) is measurable.
\end{itemize}
The functor \(\mathbf{D}\), together with \(\eta\) and \(\mu\), satisfies the monad laws:
\begin{enumerate}
    \item Associativity: \(\mu \circ \mathbf{D}(\mu) = \mu \circ \mu_{\mathbf{D}}\),
    \item Unit: \(\mu \circ \eta_{\mathbf{D}} = \text{Id}_{\mathbf{D}} = \mu \circ \mathbf{D}(\eta)\).
\end{enumerate}
\end{Definition}

The probability monad can be constructed upon the distribution monad.

\begin{Definition}[Probability Monad~\cite{Giry1982ACA}]
A \textbf{probability monad} \((\mathbf{P}, \eta', \mu')\) on a category \(\mathsf{C}\) of measurable spaces is a monad that assigns probability measures to measurable spaces. It consists of the following components:
\begin{itemize}
    \item \textbf{Functor:} A functor \(\mathbf{P}: \mathsf{C} \to \mathsf{C}\) that assigns to each measurable space \((X, \mathcal{A})\) the space \(\mathbf{P}(X)\), the set of all probability measures on \(X\) satisfying \(\mu(X) = 1\), equipped with the smallest \(\sigma\)-algebra making the evaluation maps measurable.
    \item \textbf{Unit Natural Transformation:} A unit natural transformation \(\eta': \text{Id}_{\mathsf{C}} \Rightarrow \mathbf{P}\), such that for each \(X \in \text{Ob}(\mathsf{C})\), the map \(\eta'_X: X \to \mathbf{P}(X)\) assigns to each \(x \in X\) the Dirac measure \(\delta_x \in \mathbf{P}(X)\), where:
    \[
    \delta_x(A) = 
    \begin{cases} 
      1 & \text{if } x \in A, \\
      0 & \text{if } x \notin A.
    \end{cases}
    \]
    \item \textbf{Multiplication Natural Transformation:} A multiplication natural transformation \(\mu': \mathbf{P} \circ \mathbf{P} \Rightarrow \mathbf{P}\) that combines measures in \(\mathbf{P}(\mathbf{P}(X))\) to produce a measure in \(\mathbf{P}(X)\). For a measure \(\nu \in \mathbf{P}(\mathbf{P}(X))\), \(\mu'\) satisfies:
    \[
    \mu'_X(\nu)(A) = \int_{\mathbf{P}(X)} \mu(A) \, d\nu(\mu),
    \]
    for \(A \in \mathcal{A}\).
\end{itemize}

The monad \((\mathbf{P}, \eta', \mu')\) must satisfy the monad laws:
\begin{enumerate}
    \item \textbf{Left and Right Unit Laws:}
    \[
    \mu'_X \circ \mathbf{P}(\eta'_X) = \text{Id}_{\mathbf{P}(X)} \quad \text{and} \quad \mu'_X \circ \eta'_{\mathbf{P}(X)} = \text{Id}_{\mathbf{P}(X)}.
    \]
    \item \textbf{Associative Law:}
    \[
    \mu'_X \circ \mathbf{P}(\mu'_X) = \mu'_X \circ \mu'_{\mathbf{P}(X)}.
    \]
\end{enumerate}
\end{Definition}

As mentioned earlier, Markov categories are categories where morphisms encode ``randomness''. In categorical probability-based learning, a Markov category background is necessary to describe Bayesian inversion, stochastic dependencies, and the interplay between randomness and determinism.

\begin{Definition}[Markov Category~\cite{fritz2020markov}]
A \textbf{Markov category} is a symmetric monoidal category \((\mathsf{C}, \otimes, I)\) in which every object \(X \in \text{Ob}(\mathsf{C})\) is equipped with a commutative comonoid structure, consisting of:
\begin{itemize}
    \item A \textbf{comultiplication map} \(\text{cp}_X: X \to X \otimes X\),
    \item A \textbf{counit map} \(\text{del}_X: X \to I\),
\end{itemize}
These maps must satisfy coherence laws with the monoidal structure. Additionally, the counit map \(\text{del}\) must be natural with respect to every morphism \(f\).
\end{Definition}

One of the most common ways to construct a Markov category is by constructing it as a \textbf{Kleisli category}, which involves adding a commutative monad structure to a base category. In the following, we introduce a typical example: the Giry monad on \(\mathsf{Meas}\). The Kleisli morphisms of the Giry monad on \(\mathsf{Meas}\) (and related subcategories) are Markov kernels. Therefore, its Kleisli category is the category \(\mathsf{Stoch}\), which is an essential example of a Markov category.

\begin{Definition}[Kleisli Category~\cite{maclane1998categories}]
Given a monad \((\mathbf{T}, \eta, \mu)\) on a category \(\mathsf{C}\), the \textbf{Kleisli category} \(K(\mathbf{T})\) associated with \(\mathbf{T}\) is defined as follows:
\begin{itemize}
    \item \textbf{Objects:} The objects of \(K(\mathbf{T})\) are the same as the objects of \(\mathsf{C}\).
    \item \textbf{Morphisms:} For objects \(A, B \in K(\mathbf{T})\), the morphisms in \(K(\mathbf{T})\) are given by:
    \[
    K(\mathbf{T})(A, B) = \mathsf{C}(A, \mathbf{T}(B)),
    \]
    where \(\mathsf{C}(A, \mathbf{T}(B))\) denotes the set of morphisms in \(\mathsf{C}\) from \(A\) to \(\mathbf{T}(B)\).
    \item \textbf{Composition:} For \(f: A \to \mathbf{T}(B)\) and \(g: B \to \mathbf{T}(C)\), their composition \(g \circ f: A \to \mathbf{T}(C)\) in \(K(\mathbf{T})\) is defined as:
    \[
    g \circ f = \mu_C \circ \mathbf{T}(g) \circ f,
    \]
    where \(\mathbf{T}(g): \mathbf{T}(B) \to \mathbf{T}(\mathbf{T}(C))\) is the functorial action of \(\mathbf{T}\), and \(\mu_C: \mathbf{T}(\mathbf{T}(C)) \to \mathbf{T}(C)\) is the monad multiplication.
    \item \textbf{Identity:} For each object \(A \in K(\mathbf{T})\), the identity morphism \(\text{Id}_A: A \to \mathbf{T}(A)\) is given by the unit \(\eta_A: A \to \mathbf{T}(A)\) of the monad.
\end{itemize}
\end{Definition}

\begin{Definition}[Giry Monad\cite{Giry1982ACA}]
The \textbf{Giry monad} is a monad defined on the category of measurable spaces \((\mathsf{Meas})\). It assigns to each measurable space \(X = (X, \mathcal{A})\) the space \(\mathbf{G}(X)\), the collection of all probability measures on \(X\), equipped with the \(\sigma\)-algebra generated by the evaluation maps:
\[
ev_U : \mathbf{G}(X) \to [0, 1], \quad P \mapsto P(U),
\]
where \(U \in \mathcal{A}\).

The Giry monad consists of:
\begin{itemize}
    \item \textbf{Functor:} The endofunctor \(\mathbf{G}: \mathsf{Meas} \to \mathsf{Meas}\) maps a measurable space \(X\) to \(\mathbf{G}(X)\) and a measurable function \(f: X \to Y\) to the pushforward map \(\mathbf{G}(f): \mathbf{G}(X) \to \mathbf{G}(Y)\), defined as:
    \[
    \mathbf{G}(f)(P)(V) = P(f^{-1}(V)) \quad \text{for } P \in \mathbf{G}(X) \text{ and } V \in \mathcal{B}(Y).
    \]
    \item \textbf{Unit:} The unit natural transformation \(\eta: \text{Id}_{\mathsf{Meas}} \Rightarrow \mathbf{G}\) maps each point \(x \in X\) to the Dirac measure \(\delta_x \in \mathbf{G}(X)\), defined by:
    \[
    \delta_x(U) = 
    \begin{cases} 
      1 & \text{if } x \in U, \\
      0 & \text{if } x \notin U,
    \end{cases}
    \]
    for \(U \in \mathcal{A}\).
    \item \textbf{Multiplication:} The multiplication natural transformation \(\mu: \mathbf{G} \circ \mathbf{G} \Rightarrow \mathbf{G}\) is defined for \(Q \in \mathbf{G}(\mathbf{G}(X))\) and \(U \in \mathcal{A}\) as:
    \[
    \mu_X(Q)(U) = \int_{\mathbf{G}(X)} P(U) \, \mathrm{d}Q(P).
    \]
\end{itemize}

The Giry monad satisfies the monad laws, ensuring associativity of \(\mu\) and compatibility with \(\eta\), making it a valid monad in the category of measurable spaces.
\end{Definition}

To generalize, given a distribution monad \(\mathbf{D}\) where:
\[
\mathbf{D}(X) = \left\{ f: X \to [0,1] : \left\lvert \text{supp}(f) \right\rvert < +\infty, \sum_{x \in X} f(x) = 1 \right\}
\]
The corresponding Kleisli category (a monad structure in addition to a base category, with morphisms like \(\mathsf{C}(X, \mathbf{D}(Y))\)) can describe the probability distribution of one object associated with another, forming a Markov category. Hence, once there is a probability distribution, there is a natural corresponding Markov category.

Next, we introduce another commonly used distribution monad in probability-based learning.

\begin{Definition}[Kantorovich Monad~\cite{villani2009optimal}]
A \textbf{Kantorovich monad} \((\mathbf{T}, \eta, \mu)\) on a category \(\mathsf{C}\), enriched over the category of measurable spaces, is a monad satisfying the following:
\begin{enumerate}
    \item \textbf{Functor:} \(\mathbf{T}\) is an endofunctor on \(\mathsf{C}\) that maps each object \(X \in \mathsf{C}\) to the space \(\mathbf{T}(X)\), the set of probability measures on \(X\), equipped with additional structure, such as a Wasserstein metric or a related topology.
    \item \textbf{Unit:} The unit natural transformation \(\eta: 1_{\mathsf{C}} \Rightarrow \mathbf{T}\) maps each point \(x \in X\) to the Dirac measure \(\delta_x \in \mathbf{T}(X)\), preserving the measurable structure.
    \item \textbf{Multiplication:} The multiplication natural transformation \(\mu: \mathbf{T} \circ \mathbf{T} \Rightarrow \mathbf{T}\) is defined for \(Q \in \mathbf{T}(\mathbf{T}(X))\) by:
    \[
    \mu_X(Q)(A) = \int_{\mathbf{T}(X)} P(A) \, \mathrm{d}Q(P),
    \]
    where \(P \in \mathbf{T}(X)\), \(Q\) is a probability measure on \(\mathbf{T}(X)\), and \(A\) is a measurable subset of \(X\).
    \item \textbf{Monad Laws:} The following laws must hold:
    \begin{align*}
        &\mu_X \circ \mathbf{T}(\mu_X) = \mu_X \circ \mu_{\mathbf{T}(X)} \quad \text{(associativity)}, \\
        &\mu_X \circ \eta_{\mathbf{T}} = 1_{\mathbf{T}(X)} \quad \text{(unit law)}.
    \end{align*}
\end{enumerate}
\end{Definition}

\begin{Definition}[Bayesian Inverse (Posterior Distribution)~\cite{kallenberg2002foundations}]
Let \(K: X \to Y\) be a Markov kernel from a measurable space \((X, \mathcal{A})\) to a measurable space \((Y, \mathcal{B})\), and let \(\pi\) be a prior probability measure on \(X\). The \textbf{Bayesian inverse} or \textbf{posterior distribution} \(K^\ast: Y \to X\) is a Markov kernel defined for \(y \in Y\) and measurable sets \(A \in \mathcal{A}\) by:
\[
K^\ast(y)(A) = \frac{\int_A K(x, y) \, \mathrm{d}\pi(x)}{\int_X K(x, y) \, \mathrm{d}\pi(x)},
\]
provided that \(\int_X K(x, y) \, \mathrm{d}\pi(x) > 0\). 

Here:
\begin{itemize}
    \item \(K(x, y)\) represents the probability density or likelihood of \(y\) given \(x\),
    \item \(\int_X K(x, y) \, \mathrm{d}\pi(x)\) is the marginal probability of \(y\) under the prior \(\pi\),
    \item \(K^\ast(y)\) is a probability measure on \((X, \mathcal{A})\), representing the conditional distribution of \(x\) given \(y\) (the posterior distribution).
\end{itemize}
\end{Definition}

Finally, the concept of entropy is introduced to measure the information (uncertainty) in probability distributions.

\begin{Definition}[Entropy~\cite{cover2012elements}]
The \textbf{entropy} of a random variable \(X\) quantifies the uncertainty or information content associated with its probability distribution. It is defined as follows:

\begin{itemize}
    \item \textbf{Discrete Case:} If \(X\) is a discrete random variable with a probability mass function \(P(X)\) defined on a finite set \(\mathcal{X}\), the entropy \(H(X)\) is given by:
    \[
    H(X) = - \sum_{x \in \mathcal{X}} P(x) \log P(x),
    \]
    where the base of the logarithm determines the unit of entropy.

    \item \textbf{Continuous Case:} If \(X\) is a continuous random variable with a probability density function \(f(x)\) defined on a support set \(\mathcal{X}\), the entropy \(H(X)\), also referred to as \textbf{differential entropy}, is given by:
    \[
    H(X) = - \int_{\mathcal{X}} f(x) \log f(x) \, \mathrm{d}x.
    \]
\end{itemize}
The discrete entropy is always non-negative, whereas the continuous entropy can take negative values due to the use of densities. In both cases, \(H(X)\) measures the average uncertainty or surprise in the random variable \(X\).
\end{Definition}

\subsection{Framework of Categorical Bayesian Learning}
In the work of Kamiya et al. \cite{kamiya2021categorya}, the authors introduced a categorical framework for Bayesian inference and learning. Their methodology is mainly based on the ideas from \cite{cruttwell2022categorical} and \cite{fong2019backprop}, with relevant concepts from Markov categories introduced to formalize the entire framework. The key ideas can be summarized as two points: 
\begin{itemize}
    \item The combination of Bayesian inference and backpropagation induces the Bayesian inverse.
    \item The gradient-based learning process is further formalized as a functor \(\mathbf{GL}\).
\end{itemize}

As a result, they found Bayesian learning to be the simplest case of the learning paradigm described in \cite{cruttwell2022categorical}. The authors also constructed a categorical formulation for batch Bayesian updating and sequential Bayesian updating and verified the consistency of these two in a special case.

\subsubsection{Probability Model}
The basic idea of their work is to model the relationship between two random variables using conditional probabilities \( p(y \mid x) \). Unlike gradient-based learning methods, Bayesian machine learning updates the prior distribution \( q(\theta) \) on \( \theta \) using Bayes' theorem. The posterior distribution is defined by the following formula, up to a normalization constant: 
\[ p(\theta \mid y, x) \propto p(y \mid x, \theta) q(\theta \mid x). \]
This approach is fundamental to Bayesian machine learning, as it leverages Bayes' theorem to update parameter distributions rather than settling on fixed values. By focusing on these distributions instead of fixed estimates, the Bayesian framework enhances its ability to manage data uncertainty, thereby improving model generalization.

The model from \cite{cruttwell2022categorical} as a function $f$ with inputs, outputs, and parameters, is adjusted to the following model:
For the input data and parameters, the data parallelism and data fusion are much more complicated when they are distributions. Therefore, the following conceptions(such as joint distribution, disintegration, conditional distribution) become necessary.
\begin{itemize}
    \item A morphism \(\pi_X: 1 \rightarrow X\) in a Markov category \(\mathsf{C}\) can be viewed as a probability distribution on \(X\). A morphism \(f: X \rightarrow Y\) is called a \textbf{channel}.
    \item Given a pair \( f: X \to Y \) and \( \psi: 1 \to X \), a state on \( X \times Y \) can be defined as the following diagram, referred to as the \textbf{jointification of \( f \) and \( \psi \)}.

$$
\begin{tikzpicture}[auto]
\node (psi) [draw, rectangle] at (0, 2.5) {$\psi$};
\node (dot) [circle, fill, inner sep=1.5pt] at (0, 1) {};
\node (f) [draw, rectangle] at (1, -0.6) {$f$};

\draw[-] (psi) -- node[right] {$X$} (dot);
\draw[-] (dot) to[out=180, in=90] node[left] {$X$}(-1, -0.5) to (-1, -1.5);
\draw[-] (dot) to[out=0, in=90] node[right] {$X$} (1, 0);
\draw[-] (f) -- node[right] {$Y$} (1, -1.5);
\draw[-] (1, 0) -- (f);

\end{tikzpicture}
    $$
    
    \item Let \(\pi_{X \otimes Y} : 1 \to X \otimes Y\) be a joint state. A \textbf{disintegration} of \(\pi_{X \otimes Y}\) consists of a channel \(f: X \to Y\) and a state \(\pi_X : 1 \to X\) such that the following diagram commutes. If every joint state in a category allows for decomposition, then the category is said to allow for \textbf{conditional distribution}.

$$
\begin{tikzpicture}[node distance=2cm, auto]

\node (block) [draw, rectangle] at (-4, 2) {$\pi_{ X \otimes Y}$};

\draw (-4.3, 0)  -- node[left] {$X$}(-4.3, 1.72);
\draw (-3.7, 0)-- node[right] {$Y$}(-3.7, 1.72);

\node (equals) at (-2.3, 0.7) {$=$};

\node (psi) [draw, rectangle] at (0, 2.5) {$\pi_{X}$};
\node (dot) [circle, fill, inner sep=1.5pt] at (0, 1) {};
\node (f) [draw, rectangle] at (1, -0.6) {$f$};

\draw[-] (psi) -- node[right] {$X$} (dot);
\draw[-] (dot) to[out=180, in=90] node[left] {$X$}(-1, -0.5) to (-1, -1.5);
\draw[-] (dot) to[out=0, in=90] node[right] {$X$} (1, 0);
\draw[-] (f) -- node[right] {$Y$} (1, -1.5);
\draw[-] (1, 0) -- (f);

\end{tikzpicture}
$$

    \item Let \(\mathsf{C}\) be a Markov category. If for every morphism \( s: A \to X \otimes Y \), there exists a morphism \( t: X \otimes A \to Y \) such that the following commutative diagram holds, then \(\mathsf{C}\) is said to have \textbf{conditional distribution} (i.e., \( X \) can be factored out as a premise for \( Y \)).
$$
\scalebox{2}{\includegraphics{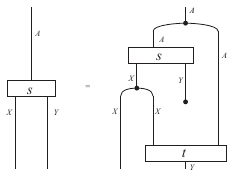}}
$$

    \item \textbf{Equivalence of Channels}:
    Let \(\pi_X: 1 \to X\) be a state on the object \(X \in \text{Ob}(\mathsf{C})\). Let \(f, g: X \to Y\) be morphisms in \(\mathsf{C}\). \(f\) is said to be almost everywhere equal to \(g\) if the following commutative diagram holds. If \(\pi_{X \times Y}: 1 \to X \times Y\) is a state and \(\pi_X: 1 \to X\) is the corresponding marginal distribution, and \(f, g: X \to Y\) are channels such that \((f, \pi_X)\) and \((g, \pi_X)\) both form decompositions with respect to \(\pi_{X \times Y}\), then \(f\) is almost everywhere equal to \(g\) with respect to \(\pi_X\).

    \item \textbf{Bayesian Inverse}: Let \(\pi_X: 1 \to X\) be a state on \(X \in \text{Ob}(\mathsf{C})\). Let \(f: X \to Y\) be a channel. The Bayesian inverse of \(f\) with respect to \(\pi_X\) is a channel \(f_{\pi_X}^{\dagger}: Y \to X\). If a Bayesian inverse \(f_{\pi_X}^{\dagger}\) exists for every state \(\pi_X\) and channel \(f\), then \(\mathsf{C}\) is said to support Bayesian inverses. This definition can be rephrased using the concept of decomposition. The Bayesian inverse \(f_{\pi_X}^{\dagger}\) can be obtained by decomposing the joint distribution \(\pi_{X \times Y}: 1 \to X \times Y\) which results from integrating over \((f, \pi_X)\), where \(f\) is a channel. The Bayesian inverse is not necessarily unique. However, if \(c_1\) and \(c_2\) are Bayesian inverses of a channel \(f: X \to Y\) with respect to the state \(\pi_X: 1 \to X\), then \(c_1\) is almost everywhere equal to \(c_2\).    
\end{itemize}

As an example, in \(\mathsf{FinStoch}\), \(\pi_{X \times Y}: 1 \to X \times Y\) corresponds to a probability distribution on \(X \times Y\). The decomposition of \(\pi_{X \times Y}\) is given by the related conditional distribution and the state \(\pi_X: 1 \to X\) obtained by marginalizing \(y\) in \(\pi_{X \times Y}\). Define the conditional distribution \(c: X \to Y\) such that, if \(\pi_X(x)\) is non-zero, then \[c(x)(y) := \frac{\pi_{X \times Y}(x, y)}{\pi_X(x)}. \] If \(\pi_X(x) = 0\), then define \(c(x)(-)\) as an arbitrary distribution on \(Y\). This can also be discussed in the subcategory \(\mathsf{BorelStoch}\), where objects are Borel spaces.

However, in order to ensure that the composition of Bayesian inverses is strict (to define a \(\mathbf{BayesLearn}\) functor similar to gradient learning functors), equivalent channels are no longer used, and instead, the category \(\mathsf{ProbStoch}\) (abbreviated to \(\mathsf{PS}\) in some references) is used (based on the following propositions). 

\begin{itemize}
    \item If a category admits conditional probabilities, it is causal. However, the converse does not hold.
    \item The categories \(\mathsf{Stoch}\) and \(\mathsf{FinStoch}\) are both causal. 
    \item If \(\mathsf{C}\) is causal, then \(\mathbf{PS}(\mathsf{C})\)(or written as \(\mathbf{ProbStoch}(\mathsf{C})\)) is symmetric monoidal.
\end{itemize}

\subsubsection{Introduction of Functor $\mathbf{Para}$}
The concept of parameterized functions has a natural expression in the categorification of machine learning models: the $\mathbf{Para}$ functor. By modifying the $\mathbf{Para}$ functor in gradient learning algorithms, the objective here is to understand conditional distributions between random variables or morphisms within a Markov category that satisfies certain constraints. In this context, a parameterized function \( f(x; \theta) \) is used to model the conditional distribution \( p(y|x) \), and the learning algorithm updates \(\theta\) using a given training set. In a Markov category, the type of the parameter \(\theta\) need not be the same as the type of the variable. By leveraging the concept of \textbf{actegories} (categories with an action), parameters are considered to act on the model.

\begin{Definition}[Actegory~\cite{etingof2015tensor}]
Let \((\mathsf{M}, \star, J)\) be a monoidal category, and let \(\mathsf{C}\) be a category.
\begin{itemize}
    \item \(\mathsf{C}\) is called an \textbf{\(\mathsf{M}\)-actegory} if there exists a strong monoidal functor \(\mathbf{\phi}: \mathsf{M} \to \mathbf{End}(\mathsf{C})\), where \(\mathbf{End}(\mathsf{C})\) is the category of endofunctors on \(\mathsf{C}\), with composition as the monoidal operation. For \(M \in \text{Ob}(\mathsf{M})\) and \(X \in \text{Ob}(\mathsf{C})\), the action is denoted by \(M \odot X = \mathbf{\phi}(M)(X)\).
    
    \item \(\mathsf{C}\) is called a \textbf{right \(\mathsf{M}\)-actegory} if it is an \(\mathsf{M}\)-actegory and is equipped with a natural isomorphism:
    \[
    \kappa_{M,X,Y}: M \odot (X \otimes Y) \cong (M \odot X) \otimes (M \odot Y),
    \]
    where \((\mathsf{C}, \otimes, I)\) is a monoidal category, and \(\kappa\) satisfies the coherence conditions.
    
    \item If \((\mathsf{C}, \otimes, I)\) is a right \(\mathsf{M}\)-actegory, the following natural isomorphisms must exist:
    \[
    \alpha_{M,X,Y}: M \odot (X \otimes Y) \cong (M \odot X) \otimes Y,
    \]
    called the \textbf{mixed associator}, and
    \[
    \mu_{M,N,X,Y}: (M \star N) \odot (X \otimes Y) \cong (M \odot X) \otimes (N \odot Y),
    \]
    called the \textbf{mixed interchanger}.
\end{itemize}
\end{Definition}

The structure of $\mathbf{Para}$ is then adjusted as follows. Let \(\mathsf{M}\) be a symmetric monoidal category and \(\mathsf{C}\) be an \(\mathsf{M}\)-actegory. \(\mathbf{Para}_{\mathsf{M}}(\mathsf{C})\) becomes a 2-category, which is defined as follows:
\begin{itemize}
    \item  Objects: The objects of \(\mathsf{C}\).
    \item 1-morphism: \(f: X \to Y\) in \(\mathbf{Para}_{\mathsf{M}}(\mathsf{C})\) consists of a pair \((P, \phi)\), where \(P \in \mathsf{M}\) and \(\phi: P \odot X \to Y\) is a morphism in \(\mathsf{C}\).
    \item Composition of 1-morphisms: Let \((P, \phi) \in \mathbf{Para}_{\mathsf{M}}(\mathsf{C})(X, Y)\) and \((Q, \psi) \in \mathbf{Para}_{\mathsf{M}}(\mathsf{C})(Y, Z)\). The composition \((P, \phi) \cp (Q, \psi)\) is the morphism in \(\mathsf{C}\) given by:
    \[
    Q \odot (P \odot X) \xrightarrow{\text{1} \odot \phi} Q \odot Y \xrightarrow{\psi} Z.
    \]
    \item 2-morphism: Let \((P, \phi), (Q, \psi) \in \mathbf{Para}_{\mathsf{M}}(\mathsf{C})(X, Y)\). A 2-morphism \(\alpha: (P, \phi) \to (Q, \psi)\) is given by a morphism \(\alpha': Q \to P\) such that the following diagram commutes:
    $$
\scalebox{0.2}{\includegraphics{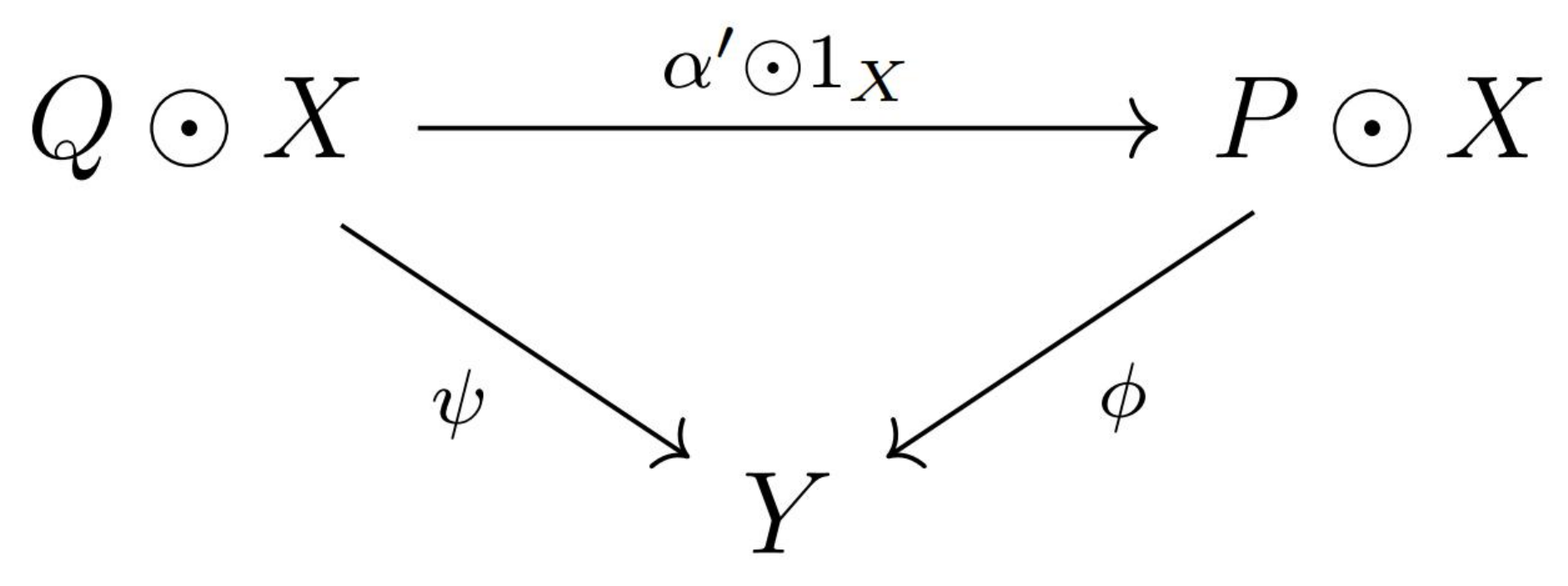}}
$$
    \item Identity morphisms and composition: These in the category \(\mathbf{Para}_{\mathsf{M}}(\mathsf{C})\) inherit from the identity morphisms and composition in \(\mathsf{M}\).    
\end{itemize}

\(\mathbf{Para}_{\mathsf{M}}(-)\) defines a pseudomonad on the category of \(\mathsf{M}\)-actegories \(\mathsf{M-Mod}\) (consider the monoidal structure of the endofunctor category, which can be composed with itself multiple times—a multi-parameter setting). In particular, if there is a functor \(\mathbf{F}: \mathsf{C} \to \mathsf{D}\), one can obtain a related functor:
\[
\mathbf{Para}_{\mathsf{M}}(\mathbf{F}): \mathbf{Para}_{\mathsf{M}}(\mathsf{C}) \to \mathbf{Para}_{\mathsf{M}}(\mathsf{D}).
\]

\subsubsection{The Final Combination: $\mathbf{BayesLearn}$ Functor}
The ultimate goal is to synthesize the entire Bayesian process into a functor \(\mathbf{BayesLearn}\), so that its construction can capture the characteristics of Bayesian learning. Beyond the aforementioned $\mathbf{Para}$ functor, a reverse feedback mechanism (lenses) needs to be incorporated. In this context, to maintain the characteristics of an actegory, \textbf{Grothendieck Lenses} are used in the construction.

Here is the text extracted from the image:

\begin{Definition}[Grothendieck Lenses\cite{maclane1998categories}]
Let \(\mathbf{F}: \mathsf{C}^{\text{op}} \to \mathsf{V-Cat}\) be a (pseudo)functor, and let \(\mathsf{Gr}(\mathbf{F})\) denote the total category arising from the Grothendieck construction of \(\mathbf{F}\). The category \(\mathsf{GrLens}_\mathbf{F}\) of \textbf{Grothendieck lenses} (or \textbf{category of \(\mathbf{F}\)-lenses}) is defined as follows:
\begin{itemize}
    \item \textbf{Objects:} Pairs \((C, X)\), where \(C \in \text{Ob}(\mathsf{C})\) and \(X \in \text{Ob}(\mathbf{F}(C))\).
    
    \item \textbf{Morphisms:} A morphism \((C, X) \to (C', X')\) consists of a pair \((f, f^\dagger)\), where:
    \begin{enumerate}
        \item \(f: C \to C'\) is a morphism in \(\mathsf{C}\),
        \item \(f^\dagger: \mathbf{F}(C')(X') \to \mathbf{F}(f)(X)\) is a morphism in \(\mathbf{F}(C')\).
    \end{enumerate}
    
    \item \textbf{Hom-Sets:} The Hom-set \(\mathsf{GrLens}_\mathbf{F}((C, X), (C', X'))\) is given by the dependent sum:
    \[
    \mathsf{GrLens}_\mathbf{F}((C, X), (C', X')) = \sum_{f \in \mathsf{C}(C, C')} \mathbf{F}(C')(X', \mathbf{F}(f)(X)).
    \]
\end{itemize}
\end{Definition}

Let \(\mathsf{C}\) be a Markov category with conditional probabilities (i.e., indicating that there are Bayesian inverses in \(\mathsf{C}\)). Bayesian inverses are typically defined up to an equivalence relation. Observing the category \(\mathbf{PS}(\mathsf{C})\), Bayesian inverses define a symmetric monoidal dagger functor. 

The gradient learning functor \(\mathbf{GL}\) is defined based on a functor \(\mathbf{R}: \mathsf{C} \to \mathbf{Lens}(\mathsf{C})\), where \(\mathsf{C}\) is a Cartesian reverse differential category. In the context of Bayesian learning, this structure is reflected through the construction of Bayesian inverses and generalized lenses. Therefore, the goal is to define a functor \(\mathbf{R}: \mathbf{PS}(\mathsf{C}) \to \mathbf{Lens}_\mathbf{F}\), where \(\mathbf{Lens}_\mathbf{F}\) is the \(\mathbf{F}\)-Lens associated with the functor \(\mathbf{PS}(\mathsf{C})^{op} \to \mathsf{Cat}\), where $\mathsf{Cat}$ is the category of small categories.

The entire construction of \(\mathbf{BayesLearn}\) is as follows.
\begin{enumerate}
    \item Define the functor \(\mathbf{Stat}: \mathbf{PS}(\mathsf{C})^{op} \to \mathsf{Cat}\): given \(X \in \text{Ob}(\mathbf{PS}(\mathsf{C}))\), let \(\mathbf{Stat}(X) := \mathbf{PS}(\mathsf{C})\).
    
    \item Define \(\mathbf{Lens}_{\mathbf{Stat}}\) as the lens corresponding to the functor \(\mathbf{Stat}\) with objects and morphisms as follows.
    \begin{itemize}
        \item Objects: For \(((X, \pi_X), (A, \pi_A))\), where \((X, \pi_X) \in \mathbf{PS}(\mathsf{C})\) and \((A, \pi_A) \in \mathbf{Stat}(X)\).
        \item Morphisms: A morphism \(((X, \pi_X), (A, \pi_A)) \to ((Y, \pi_Y), (B, \pi_B))\) is given by a morphism in \(\mathbf{PS}(\mathsf{C})\) \((X, \pi_X) \to (Y, \pi_Y)\) and a morphism in \(\mathbf{Stat}(X) = \mathbf{PS}(\mathsf{C})\) \((B, \pi_B) \to (A, \pi_A)\).
    \end{itemize}

    \item Combining with reverse derivatives, define the functor \(\mathbf{R}: \mathbf{PS}(\mathsf{C}) \to \mathbf{Lens}_{\mathbf{Stat}}\):
    Given \((X, \pi_X) \in \mathbf{PS}(\mathsf{C})\), let \(\mathbf{R}((X, \pi_X)) := ((X, \pi_X), (X, \pi_X))\).
    If \(f: (X, \pi_X) \to (Y, \pi_Y)\) is a morphism in \(\mathbf{PS}(\mathsf{C})\), then 
    $$\mathbf{R}(f): ((X, \pi_X), (X, \pi_X)) \to ((Y, \pi_Y), (Y, \pi_Y))$$
    is defined as the pair \((f, f_{\pi_X}^\dagger)\), where \(f_{\pi_X}^\dagger\) is the Bayesian inverse of \(f\) with respect to the state \(\pi_X\) on \(X\).
    
    \item Let \(\mathsf{M}\) and \(\mathsf{C}\) be Markov categories, with \(\mathsf{M}\) being causal. Assume \(\mathsf{C}\) is a symmetric monoidal \(\mathsf{M}\)-actegory consistent with \(\mathsf{M}\). Then \(\mathbf{PS}(\mathsf{M})\) is a symmetric monoidal category. The categories \(\mathbf{PS}(\mathsf{C})\) and \(\mathbf{Lens}_{\mathbf{Stat}}\) are \(\mathbf{PS}(\mathsf{M})\)-actegories. \(\mathbf{Para}_{\mathbf{PS}(\mathsf{M})}(-)\) is a functor that, when applied to \(\mathbf{R}\), yields:

    \[
    \mathbf{Para}_{\mathbf{PS}(\mathsf{M})}(\mathbf{R}): \mathbf{Para}_{\mathbf{PS}(\mathsf{M})}(\mathbf{PS}(\mathsf{C})) \to \mathbf{Para}_{\mathbf{PS}(\mathsf{M})}(\mathbf{Lens}_{\mathbf{Stat}})
    \]
    If \((\mathsf{P}, \star, J)\) is a symmetric monoidal category, then a \(\mathsf{P}\)-actegory \(\mathsf{A}\) allows a canonical functor \(\mathbf{j}_{\mathsf{P}, \mathsf{A}}: \mathsf{A} \to \mathbf{Para}_{\mathsf{P}}(\mathsf{A})\), where \(A \mapsto J \odot A\). The functor \(\mathbf{j}_{\mathsf{P}, \mathsf{A}}\) is the unit of the pseudomonad defined by \(\mathbf{Para}_{\mathsf{P}}(-)\). Thus, we obtain the following diagram:
 
    $$\scalebox{0.3}{\includegraphics{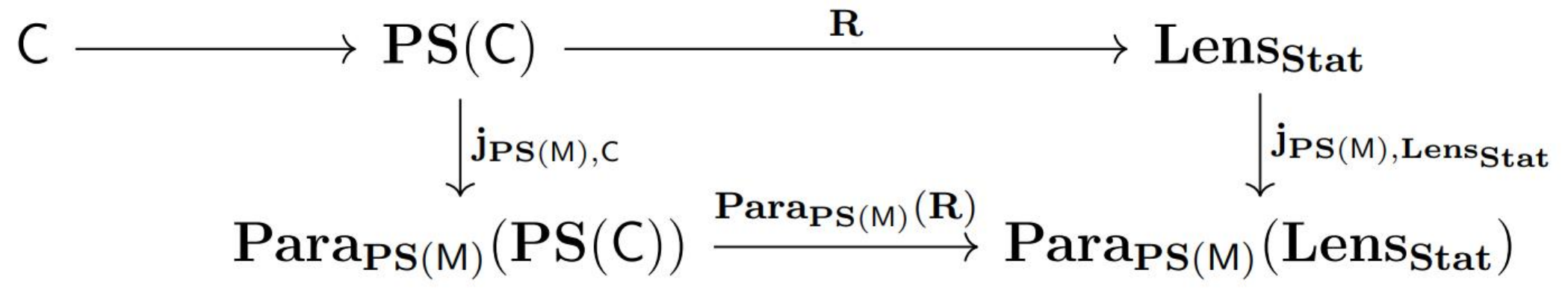}}$$ 

    \item Define the functor \(\mathbf{BayesLearn}:= \mathbf{Para}_{\mathbf{PS}(\mathsf{M})}(\mathbf{R})\).
\end{enumerate}

The $\mathbf{BayesLearn}$ functor does not include updates or shifts like the gradient learning functor. This is because the nature of Bayesian learning is relatively simplified; here, parameter updates correspond to obtaining the posterior distribution using the prior and likelihood, rather than as a result of optimization relative to a loss function.

The method of using the posterior distribution for prediction can be formalized within the category, i.e., by considering the following composition:
\[
\begin{aligned}
&(\mathsf{Y}_T, \pi_{\mathsf{Y}_T}) \otimes (\mathsf{X}^*, \pi_{\mathsf{X}^*}) 
\xrightarrow{\mathbf{f}^\dagger \otimes \text{1}} (\mathsf{M}, \pi_{\mathsf{M}}) \odot (\mathsf{X}_T, \pi_{\mathsf{X}_T}) \otimes (\mathsf{X}^*, \pi_{\mathsf{X}^*}) \\
&\cong (\mathsf{X}_T, \pi_{\mathsf{X}_T}) \otimes \big((\mathsf{M}, \pi_{\mathsf{M}}) \odot (\mathsf{X}^*, \pi_{\mathsf{X}^*})\big)
\end{aligned}
\]

\subsection{Other Related Research}
Fritz et al. \cite{fritz2023representable} systematically developed a framework for categorical probability constructs, which has become foundational for applications in probability-based machine learning. Leveraging concepts such as the Giry monad  and Markov categories, this framework aims to establish a unifying connection between classical probability theory and categorical structures. Similar frameworks have been extended and refined in works such as \cite{Meulen2020AutomaticBF, Culbertson2013BayesianML, kamiya2021categorya}, emphasizing coherence and adaptability.
One of the key applications of this framework is Bayesian inference, where log-linear models are used to represent relationships and conditional independence among variables in multivariate categorical data. By employing categorical-from-binary models, this approach facilitates efficient Bayesian analysis of generalized linear models, particularly when dealing with a large number of categories. For instance, categorical representations streamline computations by reducing the complexity of encoding and manipulating the models. Bayesian computational methods, such as conjugate priors, asymmetric hyperparameters, closed-form integration, Monte Carlo methods, and Markov Chain Monte Carlo (MCMC) techniques, are seamlessly integrated into this framework. These methods enable efficient inference while maintaining compatibility with the categorical structure. For example, MCMC procedures can be interpreted as morphisms within a Markov category, capturing the stochastic nature of transitions in a probabilistic system.
Within this framework, probability measures are interpreted as weakly averaging affine measurable functionals that preserve limits. These form a submonad of the double dualization monad on the category of measurable spaces \(\mathsf{Meas}\). This submonad has been shown to be isomorphic to the widely used Giry monad \cite{Fritz2023WeaklyMC}, reinforcing its applicability to classical probability theory. The Giry monad also serves as a bridge between categorical and probabilistic reasoning, providing a formal structure for describing and manipulating probability measures.
In addition to classical probability, this framework accommodates generalized models, including fuzzy probability theories. By describing probability measures in terms of enriched categories and submonads, it offers a flexible and extensible approach for addressing uncertainty in broader contexts. This adaptability makes it well-suited for integrating future expansions, such as models that account for imprecision or ambiguity in data.

As the earliest discussions on categorical probability learning, Baez et al. \cite{Baez2006} mentioned that from a categorical perspective, statistical field theories, which include quantum mechanics and quantum field theory in their Euclidean formulation, are technically similar to a nonparametric Bayesian approach.
They also explored the construction of probability theory in terms of a monad. They found that the Bayesian inclination to consider distributions over distributions aligns well with this construction. For instance, Dirichlet processes, a popular topic, are distributions over distributions. In \cite{corfield2009falsificationism}, the authors noted that in statistical learning theory, the richness of the hypothesis space is not controlled by the degree of a curve or the number of parameters of a hypothesis but by a construction known as the \textbf{VC-dimension}. This dimension can be viewed as measuring a degree of falsifiability. Therefore, the introduction of logical classification, such as topos structure, is natural in this context.

Beyond the previously mentioned research, some scholars emphasize the critical role of syntax in machine learning processes. Syntax provides a structured way to analyze relationships and dependencies within data, making it particularly important in fields like natural language processing. During machine learning, the process of encoding information from datasets must preserve structural and relational properties. In this regard, categorical methods offer robust performance in terms of both integrity—ensuring structural consistency during transformations—and interpretability—clarifying the propagation of data relationships. Through a functorial framework, the transformations and propagation of datasets can be systematically captured \cite{Bradley2022, Fabregat2023}.

The authors of \cite{Bradley2022} highlighted that learning algorithms do not create new properties, and treating data merely as sets often leads to information loss. To address this, they proposed modeling text data using enriched categories, which encode not just elements but also the relationships between them, such as syntactic trees or dependency structures in text. However, incorporating syntax into the learning process introduces additional complexity, as it requires metrics to evaluate similarities and differences between syntactic structures.
To tackle this challenge, \cite{Fabregat2023} introduced a \textbf{classification pseudodistance}, derived from the softmax function, to transform datasets into generalized metric spaces. This pseudodistance quantifies structural differences between entities, enabling syntax-based comparisons within a categorical framework. By framing datasets as objects in generalized metric spaces, their approach facilitates efficient analysis while preserving relational information. Additional research on metric-based approaches will be discussed in the next section.

At the application level, Bayesian synthesis significantly advances probabilistic programming by providing a structured and efficient framework for managing uncertainty in complex statistical models. Central to this approach are Bayesian networks, where the dependency structure of random variables is represented using directed acyclic graphs. They both simplify the representation of joint distributions and facilitate efficient inference by enabling the extraction of conditional independence relationships. Such capabilities are crucial for reducing computational complexity in large-scale probabilistic systems. Furthermore, by incorporating Markovian stochastic processes into probabilistic graphical models, Bayesian synthesis supports automated and semi-automated inference, allowing for the scalable management of extensive probabilistic programs \cite{Meulen2020AutomaticBF}.
Another key component of this framework is the use of probabilistic couplings, which enhance compositional reasoning and expand the expressivity of probabilistic programs. Probabilistic couplings, grounded in Strassen's theorem \cite{Aguirre2018RelationalRF}, allow for the construction of couplings between distributions without requiring a bijection between sampling spaces. This flexibility broadens the applicability of probabilistic reasoning frameworks, enabling models to accommodate diverse data structures and sampling schemes.
Additionally, \textbf{Bayesian lenses} provide a powerful mechanism for automatic Bayesian inversion, analogous to automatic differentiation in numerical computation. These lenses simplify the process of Bayesian updating—computing posterior distributions from priors and likelihoods—while allowing for the customization of numerical methods tailored to specific models. This flexibility enhances both the robustness and adaptability of probabilistic programming languages \cite{Braithwaite2022DependentBL}.

The synthetic approach to probability, which emphasizes relationships between probabilistic objects rather than their concrete definitions, provides a robust foundation for probability and statistics within the framework of Markov categories. This perspective abstracts away from specific representations, focusing instead on the compositional and structural properties of probabilistic systems. As a result, it facilitates the development of fundamental results and theorems, such as zero-one laws, which describe extreme probabilistic behaviors under certain conditions \cite{fritz2020infinite}. 
This foundation also enhances probabilistic programming, exemplified by languages like Stan and WebPPL, through the integration of Bayesian synthesis. Bayesian synthesis supports both soft constraints—allowing for flexible regularization or partial observations—and exact conditioning, which ensures that models strictly conform to observed data. This dual capability enables precise and adaptable statistical modeling. For instance, semantic analyses of exact conditioning in Gaussian probabilistic languages demonstrate how Bayesian synthesis operates within the structured framework of Markov categories, providing a rigorous basis for reasoning about probabilistic programs.

Shiebler et al. \cite{shiebler2021categorya} provide a categorical perspective on causality, covering causal independence, conditionals, and intervention effects—key components for reasoning about causality. This approach leverages the abstraction and compositionality of category theory to model causal relationships and infer causal effects. Within the Bayesian framework for causal inference, the potential outcomes approach systematically addresses causal estimands, identification assumptions, Bayesian estimation of causal effects, and sensitivity analysis. It highlights critical tools such as propensity scores, techniques for resolving identifiability issues, and the selection of priors, all of which are integral to robust causal modeling.
A formal graphical framework, grounded in monoidal categories and causal theories, introduces algebraic structures that enhance our understanding of causal relationships and facilitate the flow of information between variables. Monoidal categories represent independent or parallel processes using tensor products, while causal theories formalize how these processes interact. Markov categories, in particular, provide a compositional framework for reasoning about random mappings and noisy processing units \cite{Schauer2023CompositionalityIA, Fritz2022TheDC}. Objects in Markov categories represent spaces of possible states or data, and morphisms act as channels that may introduce noise, enabling a precise interpretation of causal relationships between variables \cite{shiebler2021categorya}.

String diagrams serve as a powerful tool for visualizing and analyzing causal models. These diagrams are compatible with morphisms in Markov categories when they can be decomposed according to the specifications of the causal model. For example, the decomposition of a string diagram might represent a causal system where variables influence each other through noisy channels, reflecting probabilistic dependencies. This compatibility illustrates the deep connection between causal structures and probabilistic relationships \cite{perrone2024markov}. Key concepts within this framework, such as second-order stochastic dominance and the Blackwell-Sherman-Stein Theorem, further refine our understanding of causal inference \cite{fritz2023representable}.
In this categorical framework, causal models abstractly represent causal independence, conditionals, and intervention effects, moving beyond model-specific methods like structural equation models or causal Bayesian networks. Instead, causal models are formalized as probabilistic interpretations of string diagrams, with equivalence established through natural transformations between functors, known as abstractions and equivalences. This abstraction enables a more general and unified understanding of causality, providing a principled foundation for reasoning about causal relationships and intervention effects.

In \cite{Sennesh2023ComputingWC}, the integration of categorical structures such as symmetric monoidal categories and operads with amortized variational inference is explored within frameworks like DisCoPyro. Symmetric monoidal categories enable the modeling of compositional processes, while operads formalize hierarchical and modular structures, both of which align naturally with the iterative and modular nature of variational inference frameworks. This integration demonstrates how Markov categories can bridge the gap between abstract mathematical foundations and practical machine learning applications, enhancing both the efficiency and expressiveness of Bayesian models. By leveraging categorical methods, it becomes possible to construct models for parametric and nonparametric Bayesian reasoning on function spaces, such as Gaussian processes, and to define inference maps analytically within symmetric monoidal weakly closed categories. These developments highlight the potential of Markov categories to provide a unified and robust foundation for supervised learning problems and general stochastic processes.
From an algebraic perspective, \cite{Mahadevan2022CategoroidsUC} introduces the concepts of \textbf{categoroid} and \textbf{functoroid} to characterize universal properties of conditional independence. \textbf{Categoroids extend traditional categories to account for probabilistic structures, while functoroids generalize functors to capture relationships of conditional independence.}
As aforementioned, there also exists research involving the category of quasi-Borel spaces, which is cartesian closed and thus supports higher-order functions and continuous distributions, thereby providing a robust framework for probabilistic reasoning \cite{Braithwaite2022DependentBL}. This approach aligns with the Curry-Howard isomorphism, providing a universal representation for finite discrete distributions. The integration of these advanced probabilistic reasoning techniques allows for the development of sophisticated models and inference algorithms, facilitating the exploration of probabilistic queries and the assignment of probabilities to complex events and beliefs \cite{Schauer2023CompositionalityIA}. 

For applications, the authors of \cite{yang2023monadic} employed the monad structure to perform automatic differentiation in reverse mode for statically typed functions that contain multiple trainable variables. 

\section{Developments in Invariance and Equivalence-based Learning}

In machine learning, invariance and equivalence are critical concepts that arise in a variety of contexts. For example, in classification tasks, ensuring consistent outcomes under transformations such as image splitting, zooming, or rotation is essential for building robust models. These transformations often relate to the geometric structure of data, which is frequently conceptualized as lying on a manifold. Similarly, in problems or network architectures that share parts of their architecture, analyzing the consistent effects of these shared elements on data is crucial for understanding equivalence. Additionally, in inference systems where semantics play a role, considerations such as how semantics evolve during the training process, which grammatical principles are preserved, and how they influence learning outcomes are vital. These concepts, though varied, can all be framed as forms of invariance or equivalence.

In category theory, approaches to studying invariance and equivalence can be broadly divided into two main types. The first approach employs functorial methods, where invariance and equivalence are treated as properties preserved by functors. Functors provide a formal mechanism for mapping between categories while preserving structural relationships, making this framework relatively straightforward and computationally efficient for practical applications. In this section, we will provide an overview of functorial constructions and related methodologies.
The second approach involves higher-order category theory, which examines invariance or equivalence at a more abstract level, using constructs such as topoi, stacks, or infinity categories. These advanced methods are particularly effective for capturing complex relationships, such as hierarchical or multi-scale dependencies, and for preserving information within networks or nodes. However, these approaches are computationally expensive and challenging to implement in practical applications. We will discuss these advanced methods in the next section, along with their implications for machine learning. 
Here, we will briefly introduce common homological methodologies, focusing on:
\begin{itemize}
    \item The work of Shiebler \cite{Dan2020a, Dan2020b}, which leverages functorial constructions;
    \item Other methodologies rooted in functorial constructions; 
    \item Persistent homological approaches, which analyze topological features in data. See for example, \cite{edelsbrunner2008persistent,Chi2018}.
\end{itemize}

The importance of invariance and equivalence in machine learning lies in the fact that many critical questions about a machine learning system can often be expressed in terms of transformation invariance and equivalence. For example (see \cite{Dan2020c}):

\begin{itemize}
    \item How does the model change in response to changes in the data?
    \item How does the model respond to transformations in the training procedure?
    \item How does the system react to changes in the data or the model?
\end{itemize}

Category theory provides a powerful framework for representing and reasoning about structures and patterns in complex systems. A functor can be thought of as a mapping between categories that preserves the essential properties of identity morphisms and the composition of morphisms. This preservation ensures that the structural relationships within a category are maintained under transformations. 
Casting algorithms as functors offers several practical benefits. It allows us to identify commonalities between different algorithms, derive extensions that maintain functorial properties, and recognize modifications that break functoriality. For instance, a functorial algorithm ensures that transformations of input data (such as rotations or translations) result in consistent and predictable changes in the output. This is particularly valuable for understanding which invariances algorithms preserve, uncovering hierarchies and similarities between algorithms, and guiding the design of new algorithms that adhere to functorial constraints.
Shiebler studied three areas of machine learning from category-theoretic perspectives: gradient-based methods, probabilistic methods, and invariant and equivariant learning (see \cite{Dan2020a}, \cite{Dan2020b}, \cite{Dan2020c}).
Especially, in invariant and equivariant learning, functoriality plays a critical role in designing algorithms that respect geometric transformations, such as those found in image processing or natural language understanding.

\subsection{Functorial Constructions and Properties}
The objective of the thesis \cite{Dan2020c} is to study the compositional and functorial structure in machine learning systems, focusing on how assumptions, problem statements, and models compose and respond to changes, which is vital for predicting system behavior. In particular, the thesis focuses on the two questions (see Questions 2 and 3 in \cite{Dan2020c}):

\begin{enumerate}
    \item How does the structure of a model reflect the structure of its training dataset? 
    \item Can we identify common structures that underlie seemingly different machine learning systems? 
\end{enumerate}

Shiebler \cite{Dan2020a} characterizes a class of hierarchical overlapping clustering algorithms as functors that factor through a category of simplicial complexes. He developed a pair of adjoint functors that map between simplicial complexes and the outputs of clustering algorithms. In a subsequent study \cite{Dan2020b}, Shiebler further characterizes manifold learning algorithms as functors that map metric spaces to optimization objectives, factoring through hierarchical clustering functors. This characterization is used to prove refinement bounds and construct a hierarchy of manifold learning algorithms based on their equivariants. Several manifold learning algorithms are described as functors at different levels of the hierarchy. By projecting manifold learning algorithms along a spectrum based on certain criteria, new manifold learning algorithms can be derived.

In this section, we mainly introduce algorithms that extract structure from unlabeled data, i.e., unsupervised learning. Studying these algorithms' properties helps people understand how they separate signals from noise, focusing on the invariant and equivariant properties of the functors.

\begin{enumerate}
    \item \textbf{Clustering}: A clustering algorithm takes a finite metric space and assigns each point in the space to a cluster.
    \item \textbf{Manifold Learning}: Manifold learning algorithms, such as Isomap, Metric Multidimensional Scaling, and UMAP, construct \(\mathbb{R}^d\)-embeddings for the points in \(X\), which are interpreted as coordinates for the support \(\mu_X\). These techniques are based on the assumption that this support can be well-approximated with a manifold.
\end{enumerate}

In the following, we first introduce the fundamental notions and then delve into the core ideas.

\subsubsection{Preliminaries and Notions}
In many learning systems, especially manifold learning, a dataset is represented as a finite set of points. Simultaneously, invariance and equivariance are expressed as ``keeping the distances between points''. \textbf{Uber-metric spaces} allow for infinite distances and non-identical points with zero distance.

\begin{Definition}[Finite Uber-Metric Space and the Category \(\mathsf{UMet}\)~\cite{kelly1982enriched}]
A \textbf{finite uber-metric space} is a pair \((X, d_X)\), where:
\begin{itemize}
    \item \(X\) is a finite set,
    \item \(d_X: X \times X \to \mathbb{R} \cup \{\infty\}\) is a function satisfying the following properties:
    \begin{enumerate}
        \item \textbf{Identity:} \(d_X(x, x) = 0\) for all \(x \in X\),
        \item \textbf{Symmetry:} \(d_X(x_1, x_2) = d_X(x_2, x_1)\) for all \(x_1, x_2 \in X\),
        \item \textbf{Triangle Inequality:} \(d_X(x_1, x_3) \leq d_X(x_1, x_2) + d_X(x_2, x_3)\) for all \(x_1, x_2, x_3 \in X\).
    \end{enumerate}
\end{itemize}

A map \(f: X \to Y\) between two uber-metric spaces \((X, d_X)\) and \((Y, d_Y)\) is called \textbf{non-expansive} if:
\[
d_Y(f(x_1), f(x_2)) \leq d_X(x_1, x_2) \quad \text{for all } x_1, x_2 \in X.
\]

The \textbf{category \(\mathsf{UMet}\)} consists of:
\begin{itemize}
    \item \textbf{Objects:} Finite uber-metric spaces,
    \item \textbf{Morphisms:} Non-expansive maps between uber-metric spaces.
\end{itemize}
\end{Definition}

To structure the locals on a manifold and to relate those locals with common topological structures, such as simplices, the following notions are necessary.

 \begin{Definition}[Non-Nested Flag Cover\cite{Dan2020a}]
Let \(X\) be a set. A \textbf{non-nested flag cover} \(C_X\) of \(X\) is a collection of subsets of \(X\) satisfying the following conditions:
\begin{itemize}
    \item \textbf{Non-Nestedness:} If \(A, B \in C_X\) and \(A \subseteq B\), then \(A = B\).
    \item \textbf{Flag Property:} The simplicial complex associated with \(C_X\), defined by:
    \[
    \mathcal{K}(C_X) = \{ \sigma \subseteq X \mid \sigma \subseteq A \text{ for some } A \in C_X \},
    \]
    is a \textbf{flag complex}. This means that every set of vertices in \(\mathcal{K}(C_X)\) that are pairwise connected by edges spans a simplex in \(\mathcal{K}(C_X)\).
\end{itemize}
\end{Definition}

\begin{Definition}[Refinement preserving and the category \(\mathsf{Cov}\)~\cite{Dan2020a}]
A map \(f: (X, C_X) \to (Y, C_Y)\) is \textbf{refinement preserving} if for any set \(S\) in \(C_X\), there exists some set \(S^\prime\) in \(C_Y\) such that \(f(S) \subseteq S^\prime\).
Non-nested flag covers form the \textbf{category \(\mathsf{Cov}\)} with refinement-preserving functions as the morphisms.
 \end{Definition}

For example, consider the tuples \((\{1, 2, 3\}, \{\{1, 2\}, \{2, 3\}\})\) and \((\{a, b\}, \\\{\{a, b\}\})\) as objects in \(\mathsf{Cov}\). The function \(f(1) = a, f(2) = b, f(3) = b\) is a morphism between them.

An overlapping clustering algorithm can be described as a functor as follows. The details will be provided in the next subsection.

\begin{Definition}[Flat Clustering Functor~\cite{Dan2020a}]
A \textbf{flat clustering functor} is a functor \(\mathbf{F}:\mathsf{UMet} \to \mathsf{Cov}\) that satisfies the following properties:
\begin{itemize}
    \item \textbf{Identity on Underlying Sets:} For each object \((X, d_X) \in \mathsf{UMet}\), the underlying set of \(\mathbf{F}(X, d_X)\) is \(X\).
    \item \textbf{Preservation of Structure:} \(\mathbf{F}\) maps morphisms (non-expansive maps) in \(\mathsf{UMet}\) to morphisms in \(\mathsf{Cov}\), preserving the clustering structure.
\end{itemize}

Compared to non-overlapping clustering, overlapping clustering algorithms allow elements to belong to multiple clusters, preserving more information about the original space.
\end{Definition}

\begin{Definition}[Finite Simplicial Complex~\cite{hatcher2002algebraic}]
A \textbf{finite simplicial complex} \(S_X\) is a collection of finite subsets of a set \(X\) (called the \textbf{vertex set}) that satisfies the following properties:
\begin{itemize}
    \item \textbf{Closure under Subsets:} If \(\sigma \in S_X\) and \(\tau \subseteq \sigma\), then \(\tau \in S_X\).
    \item \textbf{Finiteness:} \(S_X\) is a finite collection of finite sets.
\end{itemize}
The elements of \(S_X\) are called \textbf{simplices}. An \(n\)-element set in \(S_X\) is called an \textbf{\(n\)-simplex}, and the elements of \(X\) are called the \textbf{vertices} of \(S_X\). The set of all \(0\)-simplices (single-element subsets of \(X\)) forms the vertex set, denoted \(X\).
\end{Definition}

For example, given a graph \(G\) with vertex set \(X\), the collection of cliques in \(G\) forms a simplicial complex, called a flag complex. Another example is the Vietoris-Rips complex as follows.

\begin{Definition}[Vietoris-Rips Complex~\cite{edelsbrunner2010computational}]
Given a finite metric space \((X, d_X)\) and a parameter \(\delta > 0\), the \textbf{Vietoris-Rips complex} \(VR(X, \delta)\) is a simplicial complex defined as follows:
\begin{itemize}
    \item The vertices of \(VR(X, \delta)\) are the elements of \(X\).
    \item A subset \(\sigma \subseteq X\) forms a simplex in \(VR(X, \delta)\) if and only if \(d_X(x, y) \leq \delta\) for all \(x, y \in \sigma\).
\end{itemize}

A sequence of parameters \(\delta_1 \leq \delta_2 \leq \dots\) induces a \textbf{filtration} on these complexes, where \(VR(X, \delta_1) \subseteq VR(X, \delta_2) \subseteq \dots\).
\end{Definition}

\begin{Definition}[Simplicial map and the category \(\mathsf{SCpx}\)~\cite{Dan2020a}]
A map \(f: X \to Y\) is a \textbf{simplicial map} if whenever the vertices \(\{x_1, x_2, \dots, x_n\}\) span an \(n\)-simplex in \(S_X\), the vertices \(\{f(x_1), f(x_2), \dots, f(x_n)\}\) span an \(m\)-simplex in \(S_Y\) for \(m \leq n\). The category with finite simplicial complexes as objects and simplicial maps as morphisms is denoted \textbf{\(\mathsf{SCpx}\)}.
\end{Definition}

The subsets of a non-nested flag cover form a simplicial complex, and the maximal simplices of \(S_X\) constitute a non-nested cover. More precisely, they give rise to adjoint functors \( \mathbf{Flag}: \mathsf{SCpx} \to \mathsf{Cov} \) and \( \mathbf{S}_{\text{fl}}: \mathsf{Cov} \to \mathsf{SCpx} \). Hence, these two functors describe how to transform between simplicial complexes and non-nested flag covers. The connected components functor \( \mathbf{\pi}_0: \mathsf{SCpx} \to \mathsf{Cov} \) is used when the connected components of \(S_X\) form a non-nested flag cover.

A functor \(\mathsf{UMet} \to \mathsf{SCpx}\) maps \((X, d_X)\) to a complex whose simplices are sets \(\{x_1, x_2, \dots, x_n\}\) such that \(d(x_i, x_j) \leq \delta\) for some choice of \(\delta\).

The category \((0,1]^{op}\) is denoted by \(\mathsf{I}^{op}\), in which objects are \(a \in (0,1]\) and morphisms are defined by the relation \(\geq\).

The following definition is necessary for the categorization of fuzzy classification, which is the process of grouping elements into fuzzy sets where membership functions are defined by the truth value of a fuzzy propositional function.

\begin{Definition}[Fibered fuzzy simplicial complex and the category \(\mathsf{FSCpx}\)~\cite{Dan2020a}]
A \textbf{fibered fuzzy simplicial complex} is a functor \(\mathbf{F}_X: \mathsf{I}^{op} \to \mathsf{SCpx}\), where the vertex set of the simplicial complex \(\mathbf{F}_X(a)\) is the same for all \(a \in \mathsf{I}\). Fibered fuzzy simplicial complexes form the \textbf{category \(\mathsf{FSCpx}\)}. The morphisms are natural transformations \(\mu\) such that each component \(\mu_{a}\) for \(a \in \mathsf{I}\) is a simplicial map.
\end{Definition}

\begin{Definition}[Functor \(\mathbf{FinSing}\)~\cite{Dan2020a}]
The \textbf{functor} \(\mathbf{FinSing}: \mathsf{UMet} \to \mathsf{FSCpx}\) maps an uber-metric space \((X, d_X)\) to a fuzzy simplicial complex whose simplices are the sets \(\{x_1, x_2, \dots, x_n\}\) such that \(d(x_i, x_j) \leq -\log(a)\).
\end{Definition}

\subsubsection{Functorial manifold learning}
Shiebler's research in this direction primarily focuses on clustering methodologies, with an emphasis on categorizing clustering algorithms through the abstraction of manifold learning techniques. Manifold learning is often employed as a preprocessing step to reduce the dimensionality of high-dimensional data, mitigating the curse of dimensionality and enabling clustering algorithms to operate more effectively. By embedding data into a lower-dimensional space that retains its geometric structure, manifold learning can reveal meaningful patterns that improve clustering outcomes compared to directly clustering the original high-dimensional data.

Consider a common setup in manifold learning: a finite metric space  \((X, d_X)\) is sampled from a larger space \(\mathbf{X}\) according to a probability measure \(\mu_{\mathbf{X}}\). 
The goal of manifold learning is to construct \(\mathbb{R}^d\)-embeddings for the points in \(X\), which are interpreted as coordinates for the support of \(\mu_{\mathbf{X}}\). Given an uber-metric space \((X, d_X)\), where \(X = \{x_1, x_2, \dots, x_n\}\), the goal is to find an \(n \times m\) real-valued matrix \(A\) in \(\mathsf{Mat}_{n,m}\). For example, in \textbf{Metric Multidimensional Scaling}, the main objective is to find a matrix \(A \in \mathsf{Mat}_{n,m}\) that minimizes \(\sum_{i,j \in \{1, \dots, n\}} (d_X(x_i, x_j) - \|A_i - A_j\|)^2\).

Given a loss function \(l: \mathsf{Mat}_{n,m} \to \mathbb{R} \cup \{\infty\}\) that accepts \(n\) embeddings in \(\mathbb{R}^m\) and outputs a real value, and a set of constraints \(c: \mathsf{Mat}_{n,m} \to \mathbb{B}\), a \textbf{manifold learning optimization problem} aims to find the \(n \times m\) matrix \(A \in \mathsf{Mat}_{n,m}\) that minimizes \(l(A)\) subject to \(c(A)\).
To summarize, a manifold learning problem is a function that maps a pseudo-metric space \((X, d_X)\) to a pairwise embedding optimization problem in the form of \((|X|, m, \{l_{ij}\})\). The optimization is to minimalize \(\{l_{ij}\}\). The invariance here lies on:
If a manifold learning problem maps an isometric quasi-metric space to an embedding optimization problem with the same solution set, it is called \textbf{isometrically invariant}. The output of an isometrically invariant manifold learning algorithm does not depend on the ordering of elements from $X$. This specific kind of manifold learning problems can be factorized by hierarchical clustering. Namely, given any isometrically invariant manifold learning problem $\mathcal{M}$, there exists a manifold learning problem \( \mathcal{L} \circ \mathcal{H} \), where $\mathcal{H}$ is a overlapping hierarchical clustering problem, and $\mathcal{L}$ is a function that maps the output of $\mathcal{H}$ to an embedding optimization problem, such that in any quasi-metric space \( (X, d_X) \), the solution spaces of the images of $\mathcal{M}$ and \( \mathcal{L} \circ \mathcal{H} \) are identical.

The categorization of Problem $\mathcal{L}$ is as follows.

\begin{Definition}[Category of manifold learning optimization problems \(\mathsf{L}\) \cite{Dan2020b}]
The \textbf{categorization of a problem \(\mathcal{L}\)}, denoted \(\mathsf{L}\), is specified as follows.
\begin{itemize}
\item Objects: Tuples \((n, \{l_{ij}\})\), where \(n\) is a natural number and \(l_{ij}: \mathbb{R}_{\geq 0} \to \mathbb{R}\) is a real-valued function that satisfies \(l_{i^\prime j^\prime}(x) = 0\) for \(i^\prime > n\) or \(j^\prime > n\).
\item Morphisms: \((n, \{l_{ij}\}) \leq (n^\prime, \{l^\prime_{ij}\})\) when \(l_{ij}(x) \leq l^\prime_{ij}(x)\) for all \(x \in \mathbb{R}_{\geq 0}\) and \(i, j \in \mathbb{N}\).
\end{itemize}
Given a choice of \(m\), the object \((n, \{l_{ij}\})\) in \(\mathsf{L}\) is viewed as a pairwise embedding optimization problem. To complete the categorization, a category \(\mathsf{Loss}\) is defined as a pullback of \(\mathsf{L}^{op}\) and \(\mathsf{L}\) (see the diagram, \( \mathbf{U} \) is a forgetful functor that maps \((n, \{l_{ij}\})\) to \(n\).), where the objects are tuples \((n, \{c_{ij}, e_{ij}\})\), and \((n, \{c_{ij}, e_{ij}\}) \leq (n^\prime, \{c^\prime_{ij}, e^\prime_{ij}\})\) whenever \(c^\prime_{ij}(x) \leq c_{ij}(x)\) and \(e_{ij}(x) \leq e^\prime_{ij}(x)\) for any \(x \in \mathbb{R}_{\geq 0}\) and \(i, j \in \mathbb{N}\). 
\end{Definition}

$$\scalebox{0.15}{\includegraphics{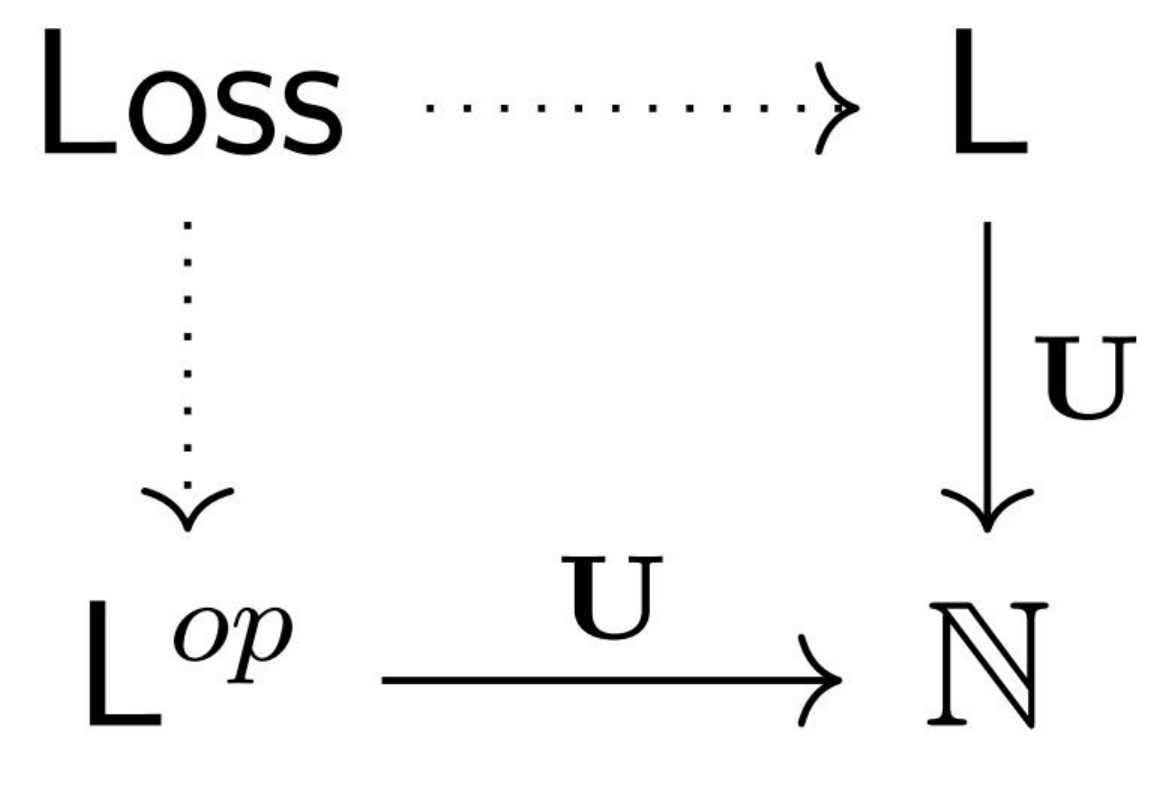}}$$ 

In particular, in the case of Metric Multidimensional Embedding Optimization Problem \((|X|, m, \{l_{ij}\})\), where \(l_{ij}(\delta) = (d_X(x_i, x_j) - \delta)^2\), each object \((n, \{c_{ij}, e_{ij}\})\) corresponds to the pairwise embedding optimization problem \((n, m, \{l_{ij}\})\), where \(l_{ij}(\delta) = c_{ij}(\delta) + e_{ij}(\delta)\).

Then, manifold learning algorithms can be represented as functors \(\mathsf{I}^{op} \to \mathsf{Loss}\). This gives rise to the category \(\mathsf{FLoss}\), where the objects are these functors \(\mathsf{I}^{op} \to \mathsf{Loss}\) that commute with the forgetful functor mapping \((n, \{c_{ij}, e_{ij}\})\) to \(n\), and the morphisms are natural transformations.

\begin{Definition}[\(\mathsf{D}\)-manifold learning functor \cite{Dan2020b}]
\label{manifold learning functor}
Given the subcategories \(\mathsf{D}\) of \(\mathsf{UMet}\) and \(\mathsf{D}^\prime\) of \(\mathsf{FCov}\), the composition \(\mathbf{L} \circ \mathbf{H}: \mathsf{D} \to \mathsf{FLoss}\) is called a \textbf{\(\mathsf{D}\)-manifold learning functor} if:
\begin{itemize}
\item \(\mathbf{H}: \mathsf{D} \to \mathsf{D}^\prime\) is a hierarchical \(\mathsf{D}\)-clustering functor;
\item \(\mathbf{L}: \mathsf{D}^\prime \to \mathsf{Loss}\) maps a fuzzy, non-nested flag cover with vertex set \(X\) to some \(\mathbf{F}_X \in \mathsf{FLoss}\) with cardinality \(|X|\).
\end{itemize}
\end{Definition}

There are two notable hierarchical clustering functors, single linkage \(\mathbf{SL}\) and maximal linkage \(\mathbf{ML}\),  since they represent two extremes, respectively. By leveraging functoriality and Definition \ref{manifold learning functor}, the authors concluded that all non-trivial clustering algorithms exist on a spectrum between maximal and single linkage to gain a similar insight of manifold learning. 

Specifically, if \(\mathbf{L} \circ \mathbf{H}\) is a \(\mathsf{D}\)-manifold learning functor, then the inequality \(\mathbf{L} \circ \mathbf{ML} \leq \mathbf{L} \circ \mathbf{H} \leq \mathbf{L} \circ \mathbf{SL}\) holds, up to a rescaling factor. There are many manifold learning functors that lie between these extremes. For any functor \(\mathbf{L}: \mathsf{FCov} \to \mathsf{Loss}\) and a sequence of clustering functors \(\mathbf{ML} \to \mathbf{H_1} \to \mathbf{H_2} \to \cdots \to \mathbf{H_n} \to \mathbf{SL}\), one obtains \(\mathbf{L} \circ \mathbf{ML} \to \mathbf{L} \circ \mathbf{H_1} \to \cdots \to \mathbf{L} \circ \mathbf{H_n} \to \mathbf{L} \circ \mathbf{SL}\).

\begin{Definition}[Functor \(\mathbf{LE}_{h,\delta}\) \cite{Dan2020b}]
The functor \(\mathbf{LE}_{h,\delta}: \mathsf{FCov}_{inj} \to \mathsf{Loss}\) maps the fuzzy, non-nested flag cover \(\mathbf{F}_X: \mathsf{I}^{op} \to \mathsf{Cov}_{inj} \in \mathsf{FCov}_{inj}\) to \((n, l_c, l_e, c)\), where \(n\) is the cardinality of the vertex set of \(\mathbf{F}_X\) and \(l_c, l_e, c\) are defined as follows:
\begin{align*}
W_{ij} &= h\left(\sup_{\leq 0} \{a \mid a \in (e^{-\delta}, 1], \exists S \in F_X(a), x_i, x_j \in S\}\right) \\
D_{ii} &= \sum_{j \in 1 \dots n} W_{ij} \\
l_e(A) &= \sum_{i,j \in 1 \dots n} \|A_i - A_j\|^2 W_{ij}, \quad l_c = 0, \quad c(A) = (A^T D A) == I
\end{align*}
\end{Definition}

For example, Laplacian Eigenmaps can be expressed as the composition \(\mathbf{LE}_{h,\delta} \circ \mathbf{ML}: \mathsf{UMet}_{inj} \to \mathsf{Loss}\).

\begin{Definition}[Functor  \(\mathbf{MMDS}\) \cite{Dan2020b}]
The functor \(\mathbf{MMDS}: \mathsf{FCov}_{bij} \to \mathsf{Loss}_m\) maps the fuzzy, non-nested flag cover \(\mathbf{F}_X: \mathsf{I}^{op} \to \mathsf{Cov}_{bij} \in \mathsf{FCov}_{bij}\) to \((n, l_c, l_e, c)\), where \(n\) is the cardinality of the vertex set of \(\mathbf{F}_X\), and \(l_c, l_e, c\) are defined as follows:
\begin{align*}
d_{F_{X_{ij}}} &= \inf\{-\log(a) \mid a \in (0,1], \exists S \in F_X(a), x_i, x_j \in S\} \\
l_c(A) &= \sum_{i,j \in 1 \dots n} l_{c_{ij}}(A) = \sum_{i,j \in 1 \dots n} \left(d^2_{F_{X_{ij}}} + \|A_i - A_j\|^2\right) \\
l_e(A) &= \sum_{i,j \in 1 \dots n} l_{e_{ij}}(A) = \sum_{i,j \in 1 \dots n} -2d_{F_{X_{ij}}}\|A_i - A_j\|^2, \quad c(A) = \text{true}
\end{align*}
\end{Definition}

The algorithm is not functorial over \(\mathsf{FCov}_{inj}\) because \(l_c\) may contain positive terms while \(l_e\) may contain negative terms. However, it is functorial over \(\mathsf{FCov}_{bij}\) since \(l_c\) increases and \(l_e\) decreases as the distance increases in \((X, d_X)\). The Metric Multidimensional Scaling (MMDS) algorithm is thus defined as \(\mathbf{MMDS} \circ \mathbf{ML}: \mathsf{UMet}_{bij} \to \mathsf{Loss}\). In this case, the loss function is \(l_c(A) + l_e(A) = \sum_{i,j \in 1 \dots n} (d_X(x_i, x_j) - \|A_i - A_j\|)^2\).

\(\mathbf{UMAP}: \mathsf{UMet}_{isom} \to \mathsf{Loss}\) is the composition of five functors, where the first four constitute a \(\mathsf{UMet}_{isom}\)-hierarchical clustering algorithm. Specifically, these functors are:
\begin{itemize}
\item $\mathbf{LocalMetric}$: Constructs a local metric space around each point in \(X\);
\item \((\mathbf{FinSing} \circ -)\): Converts each local metric space into a fuzzy simplicial complex;
\item $\mathbf{colim}$: Takes a fuzzy union of these fuzzy simplicial complexes;
\item \((\mathbf{Flag} \circ -)\): Converts the resulting fuzzy simplicial complex into a fuzzy non-nested flag cover;
\item $\mathbf{FCE}$: Constructs a loss function based on this cover.
\end{itemize}
Since $\mathbf{LocalMetric}$ does not preserve non-expansiveness, \textbf{UMAP} is not functorial over \(\mathsf{UMet}_{bij}\).

The functorial perspective on manifold learning offers a natural framework for generating new manifold learning algorithms by recombining the components of existing algorithms. For example, swapping \(\mathbf{ML}\) with \(\mathbf{SL}\) allows the definition of a Metric Multidimensional Scaling-like manifold learning functor \(\mathbf{MMDS} \circ \mathbf{SL}\), which encourages chained embeddings.

As research in this direction has advanced(most will be mentioned in Section~\ref{re_3}), several open problems remain to be solved:

\begin{itemize}
    \item Developing more robust theories regarding the resistance of various types of unsupervised and supervised algorithms to noise;
    \item Exploring the possibility of imposing stricter bounds on the stability of our results by shifting from a finite-dimensional space to a distributional perspective, potentially incorporating concepts such as surrogate loss functions and divergence measures.
\end{itemize}

\subsubsection{Functorial Clustering}
Clustering algorithms have been widely studied in data mining and machine learning, with applications including summarization, learning, image segmentation, and market segmentation. There are various methods for clustering, categorized based on the types of clusters obtained. In \cite{jain1988algorithms}, the following classifications are mentioned:
\begin{itemize}
    \item Disjoint clustering: When an element belongs to only one cluster, e.g., clustering by content $(AA, A, B, B15, C, and D)$.
    \item Fuzzy clustering: An element can belong to all clusters, but with a certain degree of membership, e.g., clustering within a color range.
    \item Overlapping clustering: An element can belong to multiple clusters, e.g., people who like Eastern and Western cuisines.
\end{itemize}

Another kind of classification is \textbf{independent and non-independent clustering}. Independent clustering does not allow overlapping, while non-independent clustering allows overlapping. In independent clustering, the \textbf{internal classification} similarity matrix, also known as \textbf{semi-supervised learning}, is used. \textbf{External classification} uses tags. Internal classification structure data can be detailed into \textbf{hierarchical clustering} and \textbf{planning clustering}. Hierarchical clustering can be further divided (these algorithms consider the similarity between new clusters formed by splitting existing clusters), taking into account a parameter that indicates the number of clusters. This parameter indicates the number of clusters.

These algorithms are suitable for clustering documents using implicit subtopics, dense graphs, subgraphs, central-based methods, density, target functions, and tree structures, allowing the most recent algorithms to perform repetitive identification and prediction tasks.

In \cite{Dan2020a}, they introduced a framework for describing a class of hierarchical overlapping clustering algorithms as functors decompose through the category of simplicial complexes. Additionally, they introduced a pair of adjoint functors to map between simplicial complexes and the outputs of clustering algorithms. As mentioned before, the maximal and single linkage clustering algorithms are introduced as compositions of the flagification and connected components functors with the finite singular set functor, respectively. They also proved that all other hierarchical overlapping clustering functors are refined by maximal linkage and can further refine single linkage.

The main problem they investigated can be described as follows. Suppose there is a finite data set \(X\) sampled from a larger space \(\mathbf{X}\) according to a probability measure \(\mu_\mathbf{X}\) defined on \(\mathbf{X}\). Clustering algorithms group the points in \(X\) together. The outputs of clustering algorithms must be consistent with the structure of the input. The authors formalized these algorithms as functors, which preserve injectivity, homogeneity, and composability when mapped between categories. This perspective can be used to identify commonalities among different algorithms, derive algorithm extensions that preserve functoriality, and identify modifications that disrupt functoriality. 
Earlier work includes the functorial description Carlsson et al.\cite{carlsson2013classifying} used to prove that a large class of non-nested clustering algorithms can be combined through single linkage, and developed a simple framework for generating clustering algorithms. In \cite{Dan2020a}, they contributed in the following aspects.
\begin{enumerate}
    \item Defining the concepts of maximal linkage and single linkage using finite topological strange attractors.
    \item Reformulating existing results on hierarchical non-nested clustering algorithms\cite{culbertson2016consistency} using the functorialization of purely simplicial flag complexes.
    \item Applying functorial methods to partition a finite cluster into infinite spaces.
\end{enumerate}

As mentioned in the preliminaries, there exists a pair of adjoint functors between the category of non-nested flag covers (which define topology and metrics based on graph-like structures) and the category of fibered fuzzy simplicial complexes (which capture the information of fuzzy classification): \( (\mathbf{S}_\text{fl} \circ -): \mathsf{FCov} \to \mathsf{FSCpx} \) and \( (\mathbf{Flag} \circ -): \mathsf{FSCpx} \to \mathsf{FCov} \).
The topological refinement process, known as flagification, is thus represented by the functor \( \mathbf{FlagCpx} = (\mathbf{S}_\text{fl} \circ -) \circ (\mathbf{Flag} \circ -) \), which converts any fuzzy simplicial complex into a fuzzy simplicial flag complex. The flagification process is completed by the \( \mathbf{Flag} \) functor. This allows for the integration of any general cover at any local into the maximal cover at that local, which corresponds to the class in the classification.

In earlier research, Spivak and McInnes\cite{spivak_metric_realization, mcinnes2019topological} introduced the adjoint functors \( \mathbf{FinReal} \) and \( \mathbf{FinSing} \), which operate similarly to the realization functor and the singular set functor in algebraic topology, facilitating the conversion between fuzzy simplicial sets and uber-metric spaces. Under the framework introduced in \cite{Dan2020a}, these functors can be further decomposed into more fundamental functors.
First, the following functor $\mathbf{Pair}$ is needed.
\begin{Definition}[Functor $\mathbf{Pair}$ \cite{Dan2020a}]
The functor \( \mathbf{Pair}: \mathsf{UMet} \to \mathsf{FSCpx} \) is defined as follows.
  \begin{itemize}
      \item For objects: A finite uber-metric space \( (X, d_X) \) is sent to a fibered fuzzy simplicial complex \( \mathbf{F}_X: \mathsf{I}^{op} \to \mathsf{SCpx} \), where for \( \alpha \in (0,1] \), \( \mathbf{F}_X(\alpha) \) is a simplicial complex whose 0-simplices are \( X \), and the 1-simplices satisfy \( d_X(x_1, x_2) \leq -\log(\alpha) \) for \( \{x_1, x_2\} \subseteq X \). \( \mathbf{F}_X(\alpha) \) has no \( n \)-simplices for \( n > 1 \).
      \item For morphisms: A map \( f: X \to Y \) is sent to a natural transformation between the two fibered fuzzy simplicial complexes \( \mathbf{Pair}(f) \), where the distribution at \( \alpha \) is given by \( f \).
  \end{itemize}
\end{Definition}

The use of a negative logarithmic function(in \( d_X(x_1, x_2) \leq -\log(\alpha) \) for \( \{x_1, x_2\} \subseteq X \)) is justified because \( -\log(\alpha) \) is a monotonic decreasing function from \([0,1]\) to \([0,\infty]\). If \( d_X(x_1, x_2) = 0 \), then the strength of the simplex \(\{x_1, x_2\}\) in \( \mathbf{Pair}(X, d_X) \) is 1, and as \( d_X(x_1, x_2) \) approaches zero, the strength of the simplex \(\{x_1, x_2\}\) approaches 1. For \( \alpha \leq \alpha^\prime \), if an \( n \)-simplex exists in \( \mathbf{F}_X(\alpha^\prime) \), then it certainly exists in \( \mathbf{F}_X(\alpha) \). Within \((0,1]\), the largest \( \alpha \) for which the fibered fuzzy simplicial complex \( \mathbf{F}_X \) has the strength is 1. Functor \( \mathbf{FinSing}: \mathsf{UMet} \to \mathsf{FSCpx} \) is then decomposed as \( \mathbf{FinSing} = \mathbf{FlagCpx} \circ \mathbf{Pair} \).

For \(\alpha \in (0,1]\), \( \mathbf{FinSing}(X, d_X)(\alpha) \) is a flag complex, where the set \(\{x_1, x_2, \dots, x_n\} \subseteq X\) forms a simplex if the pairwise distances between the points in the set are all less than or equal to \(-\log(\alpha)\). That is, \( (X, d_X) \) corresponds to a \(-\log(\alpha)\) Vietoris-Rips complex.

\begin{Definition}[Functor \( \mathbf{FinReal}\) \cite{Dan2020a}]
Its adjunct, the functor  \( \mathbf{FinReal}:\mathsf{FSCpx} \to \mathsf{UMet} \) is then defined as follows.
\begin{itemize}
    \item For objects: The vertex set \( X \) of \( \mathbf{F}_X: \mathsf{I}^{op} \to \mathsf{SCpx} \) is mapped to \( (X, d_X) \), where \( d_X(x_1, x_2) = \inf\{-\log(\alpha) \mid \alpha \in (0,1], \{x_1, x_2\} \in F_X(\alpha)[1]\} \).
    \item For morphisms: \( \mathbf{FinReal} \) sends a natural transformation \( \mu: \mathbf{F}_X \to \mathbf{F}_Y \) to the function \( f \) defined by \( \mu \) on the vertex sets of \( \mathbf{F}_X \) and \( \mathbf{F}_Y \) (the function must be non-expansive because for all \( \alpha \in (0,1] \), if \( \{x_1, x_2\} \in \mathbf{F}_X(\alpha)[1] \), then \( \{f(x_1), f(x_2)\} \) must be in \( \mathbf{F}_Y(\alpha)[0] \) or \( \mathbf{F}_Y(\alpha)[1] \)).
\end{itemize}
\end{Definition}

Note that \( \mathbf{FinReal F}_X \) is an uber-metric space, where the distance between points \( x_1 \) and \( x_2 \) is determined by the strength of the 1-simplex connecting them.

A nested clustering algorithm accepts a finite uber-metric space \( (X, d_X) \) and returns a non-nested flag cover of \( X \). Therefore, the following clustering functors are needed. Here, we suppose that \( \Lambda_\epsilon \) is the metric space where the distance between any two elements is \(\epsilon \in \mathbb{R} \geq 0\). Other algorithms, such as extrapolative clustering methods, can also be concluded into this framework with a slight modification.

\begin{itemize}
    \item \textbf{Flat Clustering Functor}: A functor \( \mathbf{C}: \mathsf{UMet} \to \mathsf{Cov} \), which is constant on the underlying set.
    \item \textbf{Non-trivial Flat Clustering Functor}: \( \mathbf{C} \) is a flat clustering functor, where there exists a clustering parameter \(\delta_\mathbf{C} \in \mathbb{R} \geq 0\) such that \( \mathbf{C}(\Lambda_{\epsilon}) \) for any \(\epsilon > \delta_\mathbf{C}\) is a single simplex containing two points, and for any \(\epsilon \leq \delta_\mathbf{C}\), \( \mathbf{C}(\Lambda_{\epsilon}) \) is a pair of two simplices. The clustering parameter \(\delta_\mathbf{C}\) is the upper bound on the distance at which the clustering functor identifies two points as belonging to the same simplex.
\end{itemize}

To capture data structure at different scales, one needs to consider the Hierarchical Clustering Functors, defined as follows \cite{Dan2020a}.

\begin{itemize}
    \item  A functor \( \mathbf{H}: \mathsf{UMet} \to \mathsf{FCov} \) is defined such that for any \(\alpha \in (0,1]\), \( \mathbf{H}(-)(\alpha): \mathsf{UMet} \to \mathsf{Cov} \) is a flat clustering functor.
    \item Non-trivial Hierarchical Clustering Functor \( \mathbf{H} \): A hierarchical clustering functor where for all \(\alpha \in (0,1]\), \( \mathbf{H}(-)(\alpha): \mathsf{UMet} \to \mathsf{Cov} \) is a flat clustering functor with clustering parameter \(\delta_{\mathbf{H},\alpha}\).
\end{itemize} 

Under this framework, the hierarchical clustering functor is further specified with the functor $\mathbf{S}_\text{fl}$.

\begin{Definition}[Hierarchical clustering functor \cite{Dan2020a}]
The category \(\mathsf{FCov}\) has objects that are functors \(\mathbf{F}_X: \mathsf{I}^{op} \to \mathsf{Cov}\) such that \(\mathbf{S}_\text{fl} \circ \mathbf{F}_X\) is a fibered fuzzy simplicial complex, and morphisms are natural transformations. A \textbf{hierarchical clustering functor} \(\mathbf{H}: \mathsf{UMet} \to \mathsf{FCov}\) is a flat clustering functor at each \(a \in \mathsf{I}\).
\end{Definition}

Examples of non-trivial hierarchical clustering functors include \textit{single linkage} and \textit{maximum linkage}. See \cite{Dan2020a}.
\begin{itemize}
    \item \textbf{Single Linkage} \(\mathbf{SL} = (\pi_0 \circ -) \circ \mathbf{FinSing}\): The points \(x_1, x_2 \in X\) lie in the same cluster with strength at least \(a\) if there exists a sequence of points \(x_1, x_i, x_{i+1}, \dots, x_{i+n}, x_2\) such that \(d(x_j, x_{j+1}) \leq -\log(a)\).
    \item \textbf{Maximum Linkage} \(\mathbf{ML} = (\mathbf{Flag} \circ -) \circ \mathbf{FinSing}\): The points \(x_1, x_2\) lie in the same cluster with strength at least \(a\) if the largest pairwise distance between them is no larger than \(-\log(a)\).
\end{itemize}

Thus, \( \mathbf{ML} \) and \( \mathbf{SL} \) become non-trivial hierarchical clustering functors. Single linkage clustering always results in a partition of \( X \), whereas maximal linkage clustering does not. The hierarchical clustering obtained from maximal linkage can reconstruct an uber-metric space. It can be proven that there exists an adjunction \( \mathbf{FinReal} \circ (\mathbf{S}_\text{fl} \circ -) \dashv \mathbf{ML} \).

\subsection{Persistent Homology}
The applications of topology in machine learning mainly include data analysis, feature extraction, dimensionality reduction, network structure design, and optimization. \textbf{Topological Data Analysis (TDA)} is at the core of these applications. It uses tools from topology to study the shape and connectivity of data, helping to extract important features from the data. Especially in high-dimensional data, traditional data analysis methods may be limited, while the topological features provided by TDA, such as \textbf{Betti numbers} and \textbf{persistent homology} (PH), are critical techniques for identifying patterns and structures in the data. For example, in the identification of complex data structures, image processing, and bioinformatics, TDA methods have been successfully applied to extract feature information that accurately reflects the intrinsic structure of the data. In machine learning, the distribution and structure of data often directly impact the learning outcomes. By leveraging TDA techniques, people can abstract the shapes within the data and extract meaningful features. PH performs a multi-scale analysis of the shape and structure of a dataset, revealing its intrinsic topological features. This analysis typically uses persistence diagrams to represent these features. These diagrams capture characteristics such as holes, voids, and connected components at multiple scales, providing important information for tasks such as classification or clustering in machine learning.

The construction of PH originates from algebraic topology and provides stable, computable, and topologically meaningful invariants for geometrical datasets such as graphs and point clouds. A classical example of PH encodes the topological features of sublevel sets \( f^{-1}((-\infty,t]) \) of a function \( f: X \to \mathbb{R} \) on a topological space \( X \), such as a graph or a point cloud. 

Theoretically, the research object of PH is sequences of linear maps between vector spaces:

\[ \cdots \to V_i \xrightarrow{\Phi_i} V_{i+1} \xrightarrow{\Phi_{i+1}} V_{i+2} \xrightarrow{\Phi_{i+2}} \cdots \]

PH focuses on identifying which vectors \( v_i \in V_i \) remain non-zero over multiple iterations, termed \textbf{persistence modules}. In ML, usually, these vector spaces are \textbf{homology groups} \( V_i = H_n(X_i) \) of topological spaces \( X_i \), which form a sequence:

\[ \cdots \hookrightarrow X_i \xrightarrow{l_i} X_{i+1} \hookrightarrow X_{i+2} \hookrightarrow \cdots \hookrightarrow X \]

The basic category employed in these researches is \(\mathsf{Pers}\), defined as follows.

\begin{Definition}[Persistence module and its category \cite{edelsbrunner2008persistent}]
    A \textbf{persistence module} \( V \) is a functor from the poset \( (\mathbb{R}, \leq) \) to the category \( \mathsf{Vect}_k \) of vector spaces over \( k \). A morphism \( \eta : V \rightarrow W \) between two persistence modules is a natural transformation between functors. The \textbf{category of persistence modules} is then denoted by \textbf{$\mathsf{Pers}$}, which is an abelian category admitting pointwise kernels, cokernels, images, and direct sums.
\end{Definition}

The goal is to identify the homology cycles that persist across multiple iterations. The main theorem in Persistent Homology (PH), the algebraic stability theorem, is applied to analyze noise in data \cite{chazal2009proximity}. Most of the early mathematical results were detailed in \cite{edelsbrunner2008persistent}, with critical contributions including \cite{ghrist2008barcodes, edelsbrunner2012persistent, huber2021persistent, dlotko2012computing}, and so on. Additionally, some research is associated with quiver representation theory, which can be considered a ``weak'' version of category theory.

A fundamental concept in topological data analysis is the \textbf{persistence barcode}, an algebraic invariant of persistence module. It characterizes the stability of topological features in an expanding family of spaces. A persistence barcode comprises a multiset of intervals on the extended real line, with each interval representing the lifespan of a topological feature within a filtration. This filtration can be based on structures such as point clouds, graphs, functions, simplicial complexes, or chain complexes. Longer intervals denote more robust features, whereas shorter intervals typically signify noise. This barcode serves as a complete invariant, encapsulating all topological information within a filtration \cite{ghrist2008barcodes}.

Additionally, the differentiability of the persistence map is considered a significant property in many studies, as it enables the application of the Newton–Raphson method. For instance, in \cite{gameiro2016continuation}, an algorithmic framework was proposed that scales effectively with the growth in data size and dimension.

In \cite{Leygonie2022}, the author provides a detailed summary of previous results and explores new theoretical issues regarding the PH descriptor for data analysis. 
These issues include establishing the notion of differentiability for PH to facilitate its application in machine learning and analyzing the fiber of PH to determine its discriminative power as a relevant descriptor. The main results of the work of this group can be summarized as follows:
\begin{enumerate}
    \item The category/space of barcodes \(\mathsf{Bar}\) is infinite-dimensional and has singularities. They defined differentiability and derivatives for maps between elements in \(\mathsf{Bar}\). This framework can be applied to PH-based gradient-descent optimization \cite{Leygonie2022a}.
    
    \item They defined Extended Persistent Homology (EPH) and integrated it into a machine learning pipeline, employing the Laplacian operator to classify datasets of graphs \cite{Yim2021}.
    
    \item They examined the properties of the fiber of the persistence map from the space of filter functions on a fixed simplicial complex to the space \(\mathsf{Bar}\). By introducing two natural actions of increasing homeomorphisms of \(\mathbb{R}\) on these spaces, they obtained the equivariance map. They discovered that the fiber has the structure of a polyhedral complex, allowing the fiber map to be viewed as a functor from barcodes to polyhedral complexes \cite{Leygonie2022b}.
    
    \item They developed an algorithm that computes the polyhedral complex forming the fiber for arbitrary simplicial complexes \( K \). This facilitates the computation of some homology groups and statistics about the fibers \cite{Leygonie2022c}.
\end{enumerate}

Jardine \cite{Jardine2019a} and \cite{Jardine2019b} examine Vietoris-Rips complexes, Lesnick complexes, and other related structures from a homotopic perspective, which are associated to an inclusion of data sets. These complexes are represented as homotopy types through their underlying posets of simplices and are regarded as systems or functors defined on a parameter space. The category of systems of spaces allows for a partial homotopy theory based on controlled equivalences. Specifically, they showed that the map of branch points associated to the inclusion of data sets is a controlled homotopy equivalence.
In \cite{Jardine2020c}, an algorithm is provided to compute the path category (fundamental category) invariant for a finite oriented simplicial complex. Additionally, a stability theorem for oriented complexes is established, which includes a specific stability result for path categories.

The survey paper by \cite{Chi2018} provides a systematic review of persistent homology (PH) and PH-based supervised and unsupervised models from a computational perspective. The paper identifies three main challenges in combining PH with machine learning: the topological representation of data, PH-based distance measurements or metrics, and PH-based feature representation. It discusses PH-based machine learning in four key steps: simplicial complex construction, PH analysis, topological feature extraction, and topology-based machine learning.

Another survey paper by \cite{Ballester2023} reviews the applications of topological data analysis (TDA), including persistent homology and Mapper, in studying the structure and behavior of neural networks and their datasets. The paper examines recent peer-reviewed articles—summarized in Appendix A—across four dimensions: the structure of a neural network, input and output spaces, internal representations and activations, and training dynamics and loss functions. These selected articles cover a wide range of applications, helping readers understand the methods used. The applications discussed include regularization, neural network pruning, detection of adversarial, out-of-distribution, and shifted examples, detection of trojaned networks, model selection, accuracy prediction, and quality assessment of generative models. The paper also provides a detailed account of challenges and potential future work in the field. 

\begin{enumerate}
\item Persistent path homology (see, for example, \cite{Chowdhury2018}) can be used to better analyze the structure of neural networks, as traditional path homology often overlooks the weights of edges.

\item Building on the work of Zhou et al. (2021) \cite{Zhou2020}, one could refine the regularization of generative neural networks and enhance neural network design using TDA.

\item There is a lack of theoretical work—most existing studies are based on experimental results—explaining the connection between topological summaries extracted from persistence diagrams and the properties of neural networks, such as their generalization capacities. Despite the challenges posed by the high computational complexity in both time and memory required to compute persistent invariants, exploring the use of multiparameter persistence in machine learning is a promising direction. Additionally, applying TDA to other architectures, such as transformers, could yield valuable insights.
\end{enumerate}

It is not yet fully understood why persistent homology (PH) is effective in certain applications, or what specific types of topological or geometric information it captures that contribute to its strong performance. The paper by \cite{Turkes2022} explores this question, particularly through experiments with synthetic point clouds in \(\mathbb{R}^2\) and \(\mathbb{R}^3\). The study demonstrates that PH can detect the number of holes, curvature, and convexity, even in scenarios with limited training data, noisy and out-of-distribution test data, and constrained computational resources. These findings may offer valuable insights for improving existing learning architectures, such as deep learning models, by incorporating PH features or kernels, as well as PH-based priors (see \cite{Zhao2020}; \cite{Solomon2021}).

Several interesting theoretical and computational problems related to persistent homology remain open:

- How can effective vectorizations be derived from a set of persistence diagrams obtained from subsamples of a fixed point cloud? \cite{Solomon2021}

The thesis by \cite{Zhou2023} explores more refined persistent invariants, extending beyond the usual persistent objects from persistent homology by replacing the homology functor with homotopy group functors, the cohomology ring functor, or the functor of taking chain complexes.

There is also some work on \(A_{\infty}\)-algebra or \(A_{\infty}\)-coalgebra structures in persistent homology or persistent cohomology, as seen in \cite{Gui2014, Herscovich2018}.

\subsection{Other Related Research}
\label{re_3}
The learnability of different neural architectures can be characterized directly by computable measures of data complexity. The paper \cite{William2018} reframes the problem of architecture selection as understanding how data determines the most expressive and generalizable architectures suited to that data beyond inductive bias. Using algebraic topology as a measure for data complexity, the authors show that the power of a network to express the topological complexity of a dataset in its decision region is a strictly limiting factor in its ability to generalize. They provide the first empirical characterization of the topological capacity of neural networks. Their analysis shows that at every level of dataset complexity, neural networks exhibit topological phase transitions. This connection between theory and empirically driven conjectures aids in selecting architectures for fully-connected neural networks. As background, Substantial Analysis has been employed to explore how various properties of neural networks, such as the depth, width, and connectivity, related to their expressivity and generalization capability. However, they were not used to determine an architecture in practice. Neural Architecture Search (NAS) views architecture selection as a compositional hyperparameter search. NAS yields expressive and powerful architectures, but it is difficult to interpret the architectures beyond their empirical optimality. Thus, they propose a third alternative: data-first architecture selection. In practice, experts design architectures with some inductive bias about the data, and the most expressive neural architectures for learning on a particular dataset are solely determined by the nature of the true data distribution. So the architecture selection can be rephrased as: given a learning problem (some dataset), which architectures are suitably regularized and expressive enough to learn and generalize on that problem? A natural approach to this question is to develop some objective measure of data complexity, and then characterize neural architectures by their ability to learn subject to that complexity. Then given some new dataset, the problem of architecture selection is reduced to compute the data complexity and choose the appropriate architecture.

Paper \cite{Giovanni2020} proposes a topological description of neural network expressive power. It adopts the topology of the space of decision boundaries realized by a neural architecture as a measure of its intrinsic expressive power. By sampling a large number of neural architectures with different sizes and design, they show how such measure of expressive power depends on the properties of the architectures, like depth, width and other related quantities. Let $F$ be the family of neural networks of a given architecture $F$ with $n$ parameters. It is common to study the set $F$ by studying its corresponding space of parameters $\mathbb{R}^n$, since for each vector in $\mathbb{R}^n$ there exists a neural network with those parameters, and each neural network describes a decision boundary. The question is: given a network architecture, what are the decision boundaries that this architecture can reproduce? The decision boundaries may be more topologically complex than planes. They consider the persistent homology of the decision boundary as an measure for the topological expressive power of a neural network. 

Paper \cite{Steven} discusses Normalizing Flows which preserve relationships between data and the latent structure of data. It is a type of generative model that transforms an intractable distribution to a more desirable distribution using bijective functions. It has a variety of mathematical applications such as variational inference, density estimation, and others such as speech generation and image generation. 

Paper \cite{Maha2022} unifies structure prediction in causal inference and reinforce learning \textbf{RL}) in terms of a simplicial object $X_n,n\geq0$, contravariant functors from ordinal numbers into a category. Higher order objects in $X_n,n\geq2$ induce higher order symmetries. The idea behind is that causal inference involves determining a structure that encodes direct causal effects between a pair of objects and multiple directed acyclic graphs models are equivalent because of symmetries induced by conditional independences. Structure discovery in causal inference and RL is modeled as the Kan extension problems of ``filling horns'', where the horn $\lambda_k^n$ is a subset of a contravariant functor of an $n$-simplex that results from removing its interior and the face opposite the $k$-vertex. Shiebler also presented similar ideas in \cite{shiebler2022kan}. Employing a similar structure, the authors of \cite{Mahadevan2024} proposed a generative AI architecture named GAIA, which employs a hierarchical model structured as a simplicial complex. 
The parameter updating is guided by lifting diagrams over simplicial sets, corresponding to different learning problems. The applications of their research were further revealed in \cite{mahadevan2024empowering}. Paper \cite{Maha2023} presents a hierarchical framework called UCLA (Universal Causality Layered Architecture), where at the top-most level, causal interventions are modeled as a higher-order category over simplicial sets and objects. At the second layer, causal models are defined as a category, for example defining the schema of a relational causal model or a symmetric monoidal category representation of DAG models. The third layer corresponds to the data layer in causal inference, where each causal object is mapped functorially into a set of instances using the category of sets and functions between sets. The fourth homotopy layer defines ways of abstractly characterizing causal models in terms of homotopy colimits, defined in terms of the nerve of a category, a functor that converts a causal (category) model into a simplicial object. Each functor between layers is characterized by a universal arrow, which define universal elements and representations through the Yoneda Lemma, and induces a Grothendieck category of elements that enables combining formal causal models with data instances, and is related to the notion of ground graphs in relational causal models. Causal inference between layers is defined as a lifting problem, a commutative diagram whose objects are categories, and whose morphisms are functors that are characterized as different types of fibrations. Paper \cite{Maha2021} characterizes homotopical equivalences between causal DAG models, exploiting the close connections between partially ordered set representations of DAGs (posets) and finite Alexandroff topologies. Causal homology is commonly used in quantum research to investigate spacetime \cite{alvarez2015}.

\section{Developments in Topos-based Learning}
Experimental deep learning involves specific datasets, ablation studies, and significance testing. However, most new algorithms primarily enhance methodologies rather than contribute to foundational theories. Since around 2018, the exploration and application of topos theory in machine learning have progressed, but it is still in the early stages.
Despite this, several studies, including reports by Laurent Lafforgue\cite{Lafforgue2022AI}, Olivia Caramello\cite{Caramello2022}, and Belfiore et al.\cite{belfiore2021topos}, have provided abstract frameworks and new directions that offer theoretical insights into general machine learning, including deep learning networks. Recently, topos-based research has emerged for specific networks, such as the transformer network \cite{villani2024topos}.

Some emerging ideas in machine learning can be examined from various categorical perspectives. Employing low-order categorical structures, such as specific base categories with topmost bi-morphisms, or high-order categorical frameworks, such as (co)sheaves and topoi, represents different ways to model the structural relationships inherent in machine learning. For example, in the following discussion, we introduce Lafforgue's report, which proposes using geometric morphisms between topoi to transfer and constrain similar structures in data and information. These geometric morphisms not only connect different topological or algebraic contexts through their associated topoi but also serve as a geometric model of the ``coarse-graining'' process and its reverse, ``refinement''. This is particularly significant from a logical perspective, as geometric morphisms effectively model the evolution of data and relationships across different layers or scales.

As the mainstream approach to applying topos theory in machine learning, the field is closely linked to Grothendieck topoi and classifying topoi\cite{Lafforgue2022AI}, which are then generated into categories of
stacks(which is a higher-categorical analogue of Grothendieck topoi)\cite{belfiore2021topos}.
The fundamental characteristics of topos theory lie in the generalization of the concept of ``space''. In simple terms, a topos can be considered a category consisting of spaces that possess specific properties, where each space locally resembles a given reference space. The categorical structure makes it analogous to the category of sets, which means many operations allowed in \(\mathsf{Set}\)—such as taking products, limits, and exponential objects—are also permissible in a topos. 

In general, neural networks can be considered as systems for transmitting and composing data through edges, with additional processing applied as needed. Existing data can be readily duplicated and combined with other data in a form similar to tensors.

Moreover, the topos itself can be considered a space with similar properties to its base category. When the topological properties of the base category are inadequate for specific operations or analyses, these can be transferred to the corresponding topos. The topos structure is particularly effective in representing geometric properties through logic since the internal logic of topoi is an intuitionistic higher-order logic. Additionally, algebraic structures can be incorporated as invariances maintained by geometric morphisms.

In this section, we aim to introduce recent advancements in the application of topos theory to machine learning.

In addition to the categorization of theoretical and experimental results, the works in this research direction can be broadly divided into two types based on whether ``semantics'' are introduced to correspond to intuitionistic logic. When exploring the deep concepts underlying the structure and 'emergent' properties of neural networks, considering semantics is rational because the flow of information transferred and transformed within computers inherently resembles a language, which aligns with type theory. For instance, outside the machine learning context, earlier works include \cite{asher2011lexical, marta2018}, an so on. These works developed a system of types, investigated the formal implications, and integrated lexical meaning theory into a compositional semantic framework. Studies such as \cite{KawaharA_1992categorical, kawaharA_1999, hyland1989theory}, and the recent \cite{Katsumata2023} have categorically interpreted the (axiomatic) semantics and structure-preserving functors that express interpretations of programming languages and inference processes using topoi. In particular, the semantics of constructive higher-order logic can be translated into function types within topoi. 

Although \cite{babonnaud2019topos} also explores this direction, the ontological hierarchy it discusses is particularly inspiring for machine learning due to two key reasons. One is the natural compatibility between the hierarchical structure in language composition and the network framework. They have demonstrated that an actual hierarchy can be obtained by implementing and applying his method to corpora through experiments. The other reason, which is more significant and currently a weak point in research, involves ``ontological concepts''. As argued in \cite{saba2018logical}, ontological concepts (which correspond to types in a strongly-typed ontology) and logical concepts (which relate to the properties and relationships between objects of various ontological types) are fundamentally different but are not yet clearly distinguished in topos-theoretic machine learning frameworks.

In addition, \cite{tasic2014knowability} discusses the similarities between ontological perspectives and the mathematical theory of topoi, hinting at the potential for us to differentiate and integrate these 'ontological' and 'logical' concepts within a topos-based theory. \cite{tasic2014knowability} also draws a parallel between Turing machines and topoi, suggesting the potential of applying topos theory to Turing machine learning networks (containing many considerable open directions).

However, from an application perspective, sometimes network performance or learning methodology can be enhanced by focusing solely on the space-like properties of topoi, which are derived from (co)sheaves.
Relate works, such as \cite{awodey2006topological,kishida2011generalized, awodey2008topology, steve2011, steve2014}, introduce the topological interpretation of logic using the open-set language. However, these studies focus on interpreting logic through topological language, which is the opposite of our goal in developing topos-theoretic machine learning: to translate topological properties into logic.

Additionally, a recent work \cite{Inigo2023ToposofNoise} explores the topos of noise, offering promising insights for statistical machine learning.

In the following sections, we will systematically introduce the main research in this direction.

\subsection{The reports of Laurent Lafforgue}
The initial framework for this approach was outlined in the report of Laurent Lafforgue on the potential roles of Grothendieck topos theory in AI. The most critical aspect of this research is the investigation of artificial neural networks using (classifying) topos theory. Topos theory corresponds to geometric logic, enabling the translation of abstract geometric properties—such as invariance and equivariance preserved by neural networks—into logical expressions that can enhance the understanding and learning mechanisms of these systems.

The core ideas for applying topos theory to explain machine learning are outlined as follows:
\begin{enumerate}
    \item \textbf{Explaining Emergence and Preserved Properties: }Phenomena such as ``emergence'' in machine learning and the consistent preservation of properties like invariance and equivariance during data processing can be modeled and explained using the language of higher-order categories and geometric logic.
    \item \textbf{Geometric Representation of Data Spaces: }Instead of relying on linear spaces for data representation, one can consider spaces analogous to geometric objects. These geometric spaces provide richer, more meaningful contexts, especially when considering the evolution of data over time.
    \item \textbf{Local-to-Global Analysis: }The properties of a space can be captured through the geometric theory of its points. This means that understanding the local geometric theory at each point is sufficient to grasp the global behavior of the space.
    \item \textbf{Logical Construction of Mathematical Theories: }Mathematical theories are constructed using a logical framework composed of language, axioms, and deduction rules. Every mathematical statement can be systematically derived within this framework, which serves as the foundation for describing machine learning processes.
    \item \textbf{Categories and Generic Models: }A mathematical theory in logic can be effectively expressed using a category with suitable properties and a ``generic model''. This model essentially acts as a functor that preserves these properties.
    \item \textbf{Mapping Between Theories: }To define a continuous mapping \(f\) between spaces with theories \(\mathbb{T}\) and \(\mathbb{T}^\prime\), one only needs to take a model of \(\mathbb{T}\) and construct a model of \(\mathbb{T}^\prime\).
    \item \textbf{Role of Classifying Topos: }A classifying topos serves as the representing object for the functor that maps a topos to the category of structures of a specific type. This allows the encoding and manipulation of neural network properties within a unified mathematical framework.
\end{enumerate}

\subsubsection{Fundamental Notions}
In this paragraph, we outline some fundamental concepts that are essential for understanding the foundational aspects of research on machine learning through topos theory. Even though these concepts can be defined in various ways, we believe the definitions provided here best capture the geometrical intuition behind them. This approach aims to offer readers without an extensive mathematical background a general understanding of the topic.

\begin{Definition}[Sieve \cite{grothendieck1962seminar}]
    A \textbf{sieve} on an object \( X \) in a category \( \mathsf{C} \) is a collection \( S \) of morphisms in \( \mathsf{C} \) with codomain \( X \) that is closed under precomposition. Specifically, if \( f: Y \to X \) is a morphism in \( S \), and \( g: Z \to Y \) is any morphism in \( \mathsf{C} \), then the composition \( f \circ g: Z \to X \) must also be in \( S \). 
\end{Definition}

Sieves are central in defining covering families within a Grothendieck topology, where they represent the set of morphisms through which a presheaf can factor.

\begin{Definition}[Grothendieck Topology \cite{grothendieck1962seminar}]
    A \textbf{Grothendieck topology} on a category \( \mathsf{C} \) defines a notion of ``covering'' for each object in \( \mathsf{C} \), facilitating the “gluing” of objects. 

    Formally, a Grothendieck topology \( J \) on \( \mathsf{C} \) is a function that assigns to each object \( c \in \mathsf{C} \) a set of sieves \( J(c) \) on \( c \), satisfying the following conditions:
    \begin{enumerate}
        \item \textbf{Maximality Axiom}: The maximal sieve \( M_c = \{ f \mid \text{cod}(f) = c \} \) is in \( J(c) \) for every \( c \in \mathsf{C} \).
        \item \textbf{Stability under Base Change}: For any morphism \( f: d \to c \) and any sieve \( S \in J(c) \), the pullback sieve \( f^*(S) = \{ g: e \to d \mid f \circ g \in S \} \) belongs to \( J(d) \).
        \item \textbf{Local Character Condition}: If \( S \) is a sieve on \( c \), and for every morphism \( f: d \to c \) in a sieve \( T \in J(c) \), the pullback \( f^*(S) \in J(d) \), then \( S \in J(c) \).
    \end{enumerate}
\end{Definition}

Essentially, a Grothendieck topology provides a way to discuss sheaves on a category $\mathsf{C}$, extending the notion of topological space to more abstract settings.

\begin{Definition}[Site \cite{grothendieck1962seminar}]
    A \textbf{site} is a category \( \mathsf{C} \) equipped with a Grothendieck topology \( J \), denoted as \( (\mathsf{C}, J) \). 

    A \textbf{morphism of sites} \( F: (\mathsf{C}, J) \to (\mathsf{D}, K) \) consists of a functor \( F: \mathsf{C} \to \mathsf{D} \) that satisfies the following condition: for every \( J \)-covering sieve \( S \) on an object \( c \in \mathsf{C} \), the sieve \( F(S) \) generated in \( \mathsf{D} \) is a \( K \)-covering sieve. 
\end{Definition}

Sites provide a foundational framework for the study of sheaves and Grothendieck topoi. The concept of ``site'' serves as the foundational structure on which sheaves can be defined. A site generalizes the notion of a topological space in that it allows for the definition of 'open sets' and 'coverings' in a categorical and very abstract manner.

\begin{Definition}[Grothendieck Topos \cite{maclane1994topos}]
    A \textbf{Grothendieck topos} is a category that generalizes the notion of the category of sheaves of sets on a topological space. 

    Formally, a Grothendieck topos is a category equivalent to the category of sheaves of sets on some site \( (\mathsf{C}, J) \), where:
    \begin{itemize}
        \item A \textbf{presheaf} on \( \mathsf{C} \) is a contravariant functor \( \mathbf{P}: \mathsf{C}^{op} \to \mathsf{Set} \).
        \item A \textbf{sheaf} on \( (\mathsf{C}, J) \) is a presheaf \( \mathbf{P} \) that satisfies the \textbf{gluing condition}: For every \( J \)-covering sieve \( S \) of \( c \in \mathsf{C} \), if \( \{x_f \in \mathbf{P}(\text{dom}(f)) \mid f \in S\} \) is compatible (i.e., \( \mathbf{P}(g)(x_f) = x_{f \circ g} \) for \( f \in S \)), there exists a unique \( x \in \mathbf{P}(c) \) such that \( \mathbf{P}(f)(x) = x_f \) for all \( f \in S \).
        \item The \textbf{category of sheaves} \( \text{Sh}(\mathsf{C}, J) \) consists of all such sheaves with morphisms given by natural transformations.
    \end{itemize}
    Thus, a Grothendieck topos is any category equivalent to \( \text{Sh}(\mathsf{C}, J) \) for some site \( (\mathsf{C}, J) \).
    \label{gro_top}
\end{Definition}

Grothendieck topoi are characterized by having certain categorical properties, such as all small limits and colimits, exponentials, and a subobject classifier. They provide a rich framework for generalizing geometric and logical concepts.

\begin{Definition}[Classifying Topos \cite{maclane1994topos}]
    A \textbf{classifying topos} for a geometric theory \( \mathbb{T} \) is a Grothendieck topos \( S[\mathbb{T}] \) that serves as a "space" representing the models of \( \mathbb{T} \). It is equipped with a universal model \( U \) of \( \mathbb{T} \), such that:

    \begin{enumerate}
        \item For any Grothendieck topos \( E \) and any model \( X \) of \( \mathbb{T} \) in \( E \), there exists a unique (up to isomorphism) geometric morphism \( f: E \to S[\mathbb{T}] \) such that the inverse image functor \( f^* \) maps the universal model \( U \) to \( X \).
        \item This implies that for every Grothendieck topos \( E \), the category of \( \mathbb{T} \)-models in \( E \) is equivalent to the category of geometric morphisms \( E \to S[\mathbb{T}] \).
    \end{enumerate}
\end{Definition}

The classifying topos \(S[\mathbb{T}]\) serves as a sort of `template' topos that encodes how structures of type \(\mathbb{T}\) should behave in any other topos. The geometric morphisms from \(E\) to \(S[\mathbb{T}]\) essentially translate the universal model in \(S[\mathbb{T}]\) to specific models in \(E\), providing a foundational link between different interpretations of the theory \(\mathbb{T}\) across various contexts.

Toposes are associated with geometric morphisms.

\begin{Definition}[Geometric Morphism \cite{maclane1994topos}]
    A \textbf{geometric morphism} between two toposes \( \mathcal{X} \) and \( \mathcal{Y} \) is a pair of adjoint functors \( (u^*, u_*) \), where:
    \begin{itemize}
        \item \( u^*: \mathcal{Y} \to \mathcal{X} \) is the left adjoint functor that preserves finite limits.
        \item \( u_*: \mathcal{X} \to \mathcal{Y} \) is the right adjoint functor, ensuring the adjunction \( u^* \dashv u_* \).
    \end{itemize}
    The preservation of finite limits by \( u^* \) and the adjunction with \( u_* \) characterizes the geometric morphism. Furthermore, \( u^* \) automatically preserves all small colimits due to the existence of the right adjoint \( u_* \).

    In essence, a geometric morphism \( f: \mathcal{X} \to \mathcal{Y} \) consists of the functor \( u^* \) that faithfully preserves the categorical structure required for logical and geometric reasoning.
\end{Definition}

\begin{Definition}[Subobject Classifier \cite{maclane1992sheaves}]
    In a category \( \mathcal{C} \) with a terminal object \( 1 \), a \textbf{subobject classifier} is an object \( \Omega \in \mathcal{C} \) together with a morphism \( \text{true}: 1 \to \Omega \) satisfying the following universal property:
    \begin{itemize}
        \item For every monomorphism \( \iota: A \to X \) in \( \mathcal{C} \), there exists a unique morphism \( \chi_\iota: X \to \Omega \) (called the \textbf{characteristic morphism}) such that the following pullback diagram commutes:
        \[
        \begin{tikzcd}
            A \arrow[r] \arrow[d, "\iota"'] & 1 \arrow[d, "\text{true}"] \\
            X \arrow[r, "\chi_\iota"'] & \Omega
        \end{tikzcd}
        \]
    \end{itemize}

    Intuitively, the subobject classifier \( \Omega \) represents the truth values for membership in subobjects, with \( \text{true} \) indicating inclusion.
\end{Definition}

\[
\begin{tikzcd}[row sep=large, column sep=large]
A \arrow[r] \arrow[d] & P \arrow[d, "\text{subobject classifier}", description] \\
X \arrow[r, "\chi_A"] & \Omega
\end{tikzcd}
\]

In conclusion, Grothendieck topos theory offers a unifying framework that bridges various domains of geometry and logic. Laurent Lafforgue et al. focus on applying this framework to machine learning to enhance interpretability, innovate new learning methodologies, and guide the overall development of the field of machine learning.

\subsubsection{The idea in the reports of Laurent Lafforgue}
The basic idea in the reports of Laurent Lafforgue \cite{Laurent2024} and \cite{Lafforgue2022AI} can be roughly concluded as follows.
These reports adopt the perspective that elements within the same category—such as various images labeled under the same class in a machine learning task—should be described using a unified language characterized by consistent symbols and axioms.
This language provides a structured and systematic way to capture shared properties among elements.
To distinguish between these elements, additional grammatical rules are necessary. For example, while all images containing cats may belong to the same category, specific features such as the relative positions of cats within each image can be captured using additional coordinate-based descriptions. To formalize this, a symbolic language characterized by explicit axioms and deduction rules is introduced to derive first-order ``geometric'' theories. The realistic elements that conform to the corresponding constraints are subsequently abstracted into models of these theories.

For a given geometric theory $\mathbb{T}$, the models of $\mathbb{T}$ in a geometric category $\mathsf{C}$ form a locally small category $\mathbb{T}\text{-mod}(\mathsf{C})$, which encompasses all elements adhering to the constraints of $\mathbb{T}$. Moreover, the geometric theory $\mathbb{T}$ can be interpreted as providing a semantic description within a Grothendieck topos $\mathcal{E}_X \simeq \text{Sh}(\mathsf{C}, J)$. This topos is defined based on a topological space $X$, which encodes the context in which the analysis of data occurs. The geometry of $\mathsf{C}$ is governed by a Grothendieck topology $J$, characterized by sieves of the form $(X_i \to X)_{i \in I}$(as defined in Definition~\ref{gro_top}). These sieves establish a local-global relationship by defining how local constraints (e.g., features or properties at specific points in the data) extend to global structures.

Consider the category of $\mathsf{Set}$-valued presheaves on $\mathsf{C}$, defined as functors $\mathsf{C}^\text{op} \to \mathsf{Set}$. This framework facilitates the study of the properties of $\mathsf{C}$ itself while leveraging constructive features inherited from $\mathsf{Set}$. Let $\hat{\mathsf{C}} := [\mathsf{C}^\text{op}, \mathsf{Set}]$ denote the collection of all presheaves on $\mathsf{C}$. Consequently, the category of $\mathbb{T}$-models on $\hat{\mathsf{C}}$ can be expressed as:
\[
\mathbb{T}\text{-mod}(\hat{\mathsf{C}}) = [\mathsf{C}^\text{op}, \mathbb{T}\text{-mod}(\mathsf{Set})].
\]
This equivalence means that the category of geometric models of the theory $\mathbb{T}$ aligns with the category of set-valued $\mathsf{C}$-indexed diagrams.

Now, consider a base topological space $X$ as a parameter characterizing the context for analysis. The continuous family of $\mathbb{T}$-models over $X$ is represented by $\mathbb{T}\text{-mod}(\mathcal{E}_X)$. The model at a specific point $x \in X$ can then be viewed as a $\mathbb{T}\text{-mod}(\mathsf{Set})$-valued sheaf, reflecting local properties of the model. The associated functor $x^*$ maps global models $\mathbb{T}\text{-mod}(\mathcal{E}_X)$ to local models $\mathbb{T}\text{-mod}(\mathsf{Set})$ and is derived from the fiber functor $x^*: \mathcal{E}_X \to \mathsf{Set}$.

Morphisms between parameters, such as a map $f: X' \to X$, induce transformations between sheaves via the pullback $f^*: \mathcal{E}_X \to \mathcal{E}_{X'}$. This pullback, in turn, induces transformations between $\mathbb{T}$-models:
\[
f^*: \mathbb{T}\text{-mod}(\mathcal{E}_X) \to \mathbb{T}\text{-mod}(\mathcal{E}_{X'}).
\]
These transformations encapsulate the semantics of $\mathbb{T}$-models under changes in the base context.

The classifying topos $\mathcal{E}_{\mathbb{T}}$ provides a universal semantic framework for $\mathbb{T}$-models, equipped with a universal model $U_{\mathbb{T}}$. The relationship between geometric morphisms and $\mathbb{T}$-models is captured by the following equivalence of categories:
\[
\{\text{Geom}(\mathcal{E}, \mathcal{E}_{\mathbb{T}})\} \longrightarrow \{\mathbb{T}\text{-mod}(\mathcal{E})\}, \quad (\mathcal{E} \xrightarrow{f} \mathcal{E}_{\mathbb{T}}) \longmapsto f^* U_{\mathbb{T}}.
\]
This equivalence formalizes how the universal model $U_{\mathbb{T}}$ is pulled back to represent $\mathbb{T}$-models in any topos $\mathcal{E}$.

Subtopoi represent refinements of Grothendieck topologies and are directly associated with quotients of theories. In this context, the logical components of a theory—such as sorts, function symbols, relation symbols, formulas, implications, and provability—are expressed categorically within the topos $\mathcal{E}_{\mathbb{T}}$. When the pullback functor $f^*$ of a topos submersion $f: \mathcal{E}' \to \mathcal{E}$ is faithful, it uniquely factors any topos morphism $f$ into:
\[
\mathcal{E}' \xrightarrow{\text{submersion}} \text{Im}(f) \hookrightarrow \mathcal{E}.
\]
This decomposition captures the precise interaction between subtopoi, morphisms, and their underlying semantics.

Logically, the quotient theory \(\mathbb{T}'\) of \(\mathbb{T}\), corresponding to the morphism \(\text{Im}(f) \rightarrow \mathcal{E}\), acts as the theory of a model \(M\). Specifically, an implication is provable in \(\mathbb{T}'\) if and only if it is verified by \(M\) in the topos \(\mathcal{E}'\). Subtopoi \(\mathcal{E}_i\) of a given topos \(\mathcal{E}\) can be organized hierarchically, and operations such as the union \(\bigcup_{i \in I} \mathcal{E}_i \rightarrow \mathcal{E}\), the intersection \(\bigcap_{i \in I} \mathcal{E}_i \rightarrow \mathcal{E}\), and the quotient \(\mathcal{E}_2 \backslash \mathcal{E}_1 \rightarrow \mathcal{E}\) are well-defined. These operations satisfy relationships such as \(\mathcal{E}_2 \backslash \mathcal{E}_1 \subseteq \mathcal{E}' \iff \mathcal{E}_2 \subseteq \mathcal{E}_1 \cup \mathcal{E}'\), reflecting the logical interplay between subtopoi and theories.

In this framework, basic realistic elements are treated as distinct entities with largely unknown semantic content, each structured as a topos. The known attributes of these elements are represented within categories \(\mathsf{C}_i\), which are connected to their corresponding topoi \(\mathcal{E}_i\) via functors \(\mathsf{C}_i \to \mathcal{E}_i\). The semantic contributions of all \(\mathcal{E}_i\)'s collectively form a vocabulary \(\Sigma\), representing an unrestricted theory devoid of axioms. The corresponding syntactic category is denoted by \(\mathsf{C}_\Sigma\). Each \(\mathsf{C}_i\), constrained by existing knowledge, is associated with two functors: the \textbf{naming functor} \(\mathbf{N}_i\) and the \textbf{partial knowledge functor} \(\mathbf{k}_i\).

The naming functor \(\mathbf{N}_i\) induces a topos morphism \(\hat{\mathsf{C}}_i \rightarrow \hat{\mathsf{C}}_\Sigma\), and any quotient theory \(\mathbb{T}\) of \(\Sigma\) specifies a subtopos \(\mathcal{E}_{\mathbb{T}} \hookrightarrow \hat{\mathsf{C}}_\Sigma\). The pullbacks of these subtopoi, denoted as \((\hat{\mathsf{C}}_i)^{\mathcal{J}_{\mathbb{T}}} \rightarrow \hat{\mathsf{C}}_i\), are defined by a topology \(\mathcal{J}_{\mathbb{T}}\), which serves as an “extrapolation principle” or “interpretation” imposed on \(\mathsf{C}_i\) by \(\mathbb{T}\).

In contexts such as \textbf{syntactic learning} and \textbf{formalized inductive reasoning}, the partial knowledge functor \(\mathbf{k}_i : \mathsf{C}_i \to \mathcal{E}_i\) induces a topos morphism \((\hat{\mathsf{C}}_i)^{\mathcal{J}_{\mathbb{T}}} \to \mathcal{E}_i\), provided that the quotient theory \(\mathbb{T}\) adheres to the extrapolation principle. This relationship can be expressed formally as:
\[
\left( {(\hat{\mathsf{C}}_i)}_{\mathcal{J}_i^\mathbb{T}} \to \hat{\mathsf{C}}_i \right) = \mathbf{N}_i^{-1} \left( \mathcal{E}_\mathbb{T} \to \hat{\mathcal{E}} \right).
\]

The process of \textbf{information extraction} is effectively described using topos-theoretic language. For any index \(i\), the image of the naming functor, \(\text{Im}(\mathbf{N}_i) \to \mathcal{E}_{\mathbb{T}}\), corresponds to a quotient theory \(\mathbb{T}_i\), which provides a description of \(\mathsf{C}_i\) within \(\mathbb{T}\). Information extraction from the family \(\text{Im}(\mathbf{N}_i) = \mathcal{E}_{\mathbb{T}_i} \hookrightarrow \mathcal{E}_{\mathbb{T}}\) into a framework expressed by \(\mathbb{T}'\) is then conceptualized as a topos morphism \(f : \mathcal{E}_{\mathbb{T}} \to \mathcal{E}_{\mathbb{T}'}\). This morphism maps the subtopoi \(\text{Im}(\mathbf{N}_i)\) to their corresponding images \(f_* \text{Im}(\mathbf{N}_i) = f_* \mathcal{E}_{\mathbb{T}_i} = \mathcal{E}_{\mathbb{T}_i'} \hookrightarrow \mathcal{E}_{\mathbb{T}'}\).

The morphism \( f \) can be decomposed into simpler morphisms \( \mathcal{E}_{\mathbb{T}_0} \xrightarrow{f_1} \mathcal{E}_{\mathbb{T}_1} \xrightarrow{f_2} \cdots \xrightarrow{f_r} \mathcal{E}_{\mathbb{T}_r} \), starting from \(\mathbb{T}_0 = \mathbb{T}\) and ending at \(\mathbb{T}_r = \mathbb{T}'\), with intermediate theories \(\mathbb{T}_{\alpha}\) for \(1 \leq \alpha < r\). Each \(f_{\alpha} : \mathcal{E}_{\mathbb{T}_{\alpha-1}} \rightarrow \mathcal{E}_{\mathbb{T}_{\alpha}}\) is facilitated by the syntactic functor \( f^*_{\alpha} : \mathsf{C}_{\mathbb{T}_{\alpha}} \rightarrow \mathsf{C}_{\mathbb{T}_{\alpha-1}} \), which integrates new concepts in accordance with the language used in the previous step.

In summary, this study demonstrates the powerful application of topos theory in structuring and analyzing complex information across different domains. By modeling realistic elements as topoi and utilizing the hierarchical structure of quotient theories and subtopoi, semantic content is integrated into a cohesive categorical framework. In the context of machine learning, each realistic element, represented as a topos, encapsulates both known and unknown properties, capturing their structural and semantic complexities through categorical notions. The process of information extraction, articulated through topos morphisms, elucidates the transformation and refinement of information across different theoretical contexts. This study also offers a systematic method for decomposing localized information and integrating it into broader frameworks, enabling a logical and structured understanding of complex systems.

\subsection{Artificial Neural Networks: Corresponding Topos and Stack}
Another contribution in 2021 from the team of Laurent Lafforgue, developed by Jean-Claude Belfiore and Daniel Bennequin \cite{belfiore2021topos, bennequin2021}, introduced stack structures into the discussion of artificial neural networks. These works employed the concepts of topoi and stacks to deepen the understanding of deep neural networks (DNNs) and suggested that this framework could be extended to other types of artificial neural networks. Their focus is on 'clarifying the inside of the black box' of neural networks to explain emergent phenomena, such as reasoning capabilities, within a logical and systematic framework. To this end, \textbf{categorical logic}, built upon the topos-stacks structure, serves as a bridge integrating logic, topology, and algebra. While Martin-Löf Type Theory (MLTT) provides a theoretical backdrop for structured reasoning and constructive mathematics, the primary emphasis of these works lies in categorical logic as a means to unify logical and statistical reasoning in machine learning. MLTT introduces key constructs such as types (generalizations of sets), dependent and inductive types, and logical equality, providing tools for reasoning about data and structures in a rigorous framework. However, the works extend beyond MLTT by leveraging the hierarchical and semantic capabilities of categorical logic within the topos framework.
The motivation behind these mathematical and logical tools arises from the gap between the efficiency of DNNs in statistical learning and Bayesian machines in probabilistic inference. While Bayesian machines provide inferences through probabilities, they lack the logical reasoning capabilities required for broader applications. By introducing categorical logic into machine learning frameworks, these works aim to integrate the strengths of both approaches. As logic serves as a bridge linking topology and algebra, the topos-stacks structure provides a systematic method for modeling, decomposing, and integrating information within efficient machine learning structures.

As introduced in the work of Laurent Lafforgue \cite{Lafforgue2022AI}, topoi constructed on generalized topological spaces, particularly Grothendieck sites, exhibit well-structured and versatile properties, making them closely associated with geometric theories.
These frameworks provide a foundation for exploring the relationships between geometry and logic. Further discussions in these two papers delve into the logical foundations of intuitionistic and contextual inference, emphasizing their applications in scientific computation, including connections to Martin-Löf Type Theory (MLTT). MLTT serves as a constructive mathematical framework, integrating type theory and formal logic, which are essential for reasoning and computation in a categorical context. Additionally, the works explore the use of topoi in fibered sites, commonly referred to as \textbf{stacks}, which extend the versatility of these structures by incorporating fibered categories.

In the following sections, we will introduce the main ideas and results of these two papers. Before that, we first introduce some basic terminology that frequently appears in their work.

\subsubsection{Preliminaries and Terminology}
The essential definitions for understanding this work are provided below.

\begin{Definition}[Groupoid \cite{lurie2009higher}]
A \textbf{small groupoid} \(\mathcal{G}\) is a space or homotopy type in which, for any points \(a\) and \(b\) in \(\mathcal{G}\), all paths $\mathsf{C}$ and \(d\) within the path space (type) \(a = b\), and all homotopies (2-paths) \(e\) and \(f\) within the path space type \(c = d\), the path space type \(e = f\) is contractible. Additionally, there is a \textbf{category} $\mathsf{Grpd}$ whose \textbf{objects} are these small groupoids and whose \textbf{morphisms} are the groupoid homomorphisms (functors).
\end{Definition}

\begin{Definition}[Stack \cite{lurie2009higher}]
Suppose \(\mathsf{C}\) is a site. A \textbf{stack} on \(\mathsf{C}\) is a 2-functor \(\mathbf{X}: \mathsf{C}^{\text{op}} \rightarrow \mathsf{Grpd}\) or \(\mathbf{X}: \mathsf{C}^{\text{op}} \rightarrow \mathsf{Cat}\), which meets the following condition:
(\textbf{Descent Condition}) \(\mathbf{X}\) maps open covers in \(\mathsf{C}\) to 2-limits in \(\mathsf{Grpd}\) or \(\mathsf{Cat}\). Specifically, if \(U = \{U_i \rightarrow U\}_{i \in I}\) is an open cover in \(\mathsf{C}\), then \(\mathsf{C}(U)\) represents its Čech complex. \(\mathbf{X}(U)\) is the 2-limit of the 2-functor \(\mathbf{X}(-): {\mathsf{C}}(U)^{\text{op}} \rightarrow \mathsf{Grpd}\) or \(\mathbf{X}(-): \mathsf{C}(U)^{\text{op}} \rightarrow \mathsf{Cat}\).
\end{Definition}

The equivalence stated in the following definition facilitates discussions on stacks, characterizing them as either pseudofunctors or groupoid fibrations. This characterization adheres to a descent condition within a Grothendieck topology on $\mathsf{C}$.

\begin{Definition}[Fibration in groupoids \cite{maclane1992sheaves}]
A \textbf{fibration fibered in groupoids} is defined by a functor \( \mathbf{p}: \mathsf{E} \rightarrow \mathsf{B} \). This functor, which functions as a (strict) functor from \( \mathsf{B}^{\text{op}} \rightarrow \mathsf{Cat} \), classifies \( \mathbf{p} \) and factors through the inclusion \( \mathsf{Grpd} \hookrightarrow \mathsf{Cat} \).  
When the context is restricted to pseudofunctors with values in \( \mathsf{Grpd} \subseteq \mathsf{Cat} \), this framework categorizes \textbf{Grothendieck fibrations in groupoids}. These are described by the functor \( \mathbf{f} : \mathbf{Func}(\mathsf{C}^{\text{op}}, \mathsf{Grpd}) \rightarrow \mathbf{FibGrpd}(\mathsf{C})\).
\end{Definition}

\begin{Definition}[Grothendieck Construction \cite{maclane1992sheaves}]
    The \textbf{Grothendieck construction} is a process that transforms a pseudofunctor \( \mathbf{F}: \mathsf{C} \to \mathsf{Cat} \) into a Grothendieck fibration. Specifically:
    \begin{itemize}
        \item The Grothendieck construction produces a category \( \int \mathbf{F} \), often called the \textbf{category of elements} of \( \mathbf{F} \), equipped with a canonical projection functor \( \pi: \int \mathbf{F} \to \mathsf{C} \).
    \end{itemize}

    \textbf{Structure of \( \int \mathbf{F} \):}
    \begin{itemize}
        \item \textbf{Objects:} The objects of \( \int \mathbf{F} \) are pairs \( (c, a) \), where \( c \in \mathsf{C} \) and \( a \in \mathbf{F}(c) \) (i.e., \( a \) is an object in the category \( \mathbf{F}(c) \)).
        \item \textbf{Morphisms:} A morphism in \( \int \mathbf{F} \) from \( (c, a) \) to \( (c', a') \) consists of a pair \( (f, \phi) \), where:
        \begin{itemize}
            \item \( f: c \to c' \) is a morphism in \( \mathsf{C} \),
            \item \( \phi: \mathbf{F}(f)(a) \to a' \) is a morphism in \( \mathbf{F}(c') \).
        \end{itemize}
    \end{itemize}

    The functor \( \pi: \int \mathbf{F} \to \mathsf{C} \) is defined by projecting \( (c, a) \) to \( c \) and \( (f, \phi) \) to \( f \). This projection satisfies the axioms of a Grothendieck fibration.

    \textbf{Equivalence of 2-Categories:} The Grothendieck construction establishes an equivalence of 2-categories:
    \[
    \mathsf{Func}(\mathsf{C}^{\text{op}}, \mathsf{Cat}) \simeq \mathsf{Fib}(\mathsf{C}),
    \]
    where \( \mathsf{Fib}(\mathsf{C}) \) is the 2-category of Grothendieck fibrations over \( \mathsf{C} \).

    When restricted to pseudofunctors with values in \( \mathsf{Grpd} \subseteq \mathsf{Cat} \), this equivalence specializes to:
    \[
    \mathsf{Func}(\mathsf{C}^{\text{op}}, \mathsf{Grpd}) \simeq \mathbf{FibGrpd}(\mathsf{C}),
    \]
    where \( \mathbf{FibGrpd}(\mathsf{C}) \) is the category of fibrations fibered in groupoids over \( \mathsf{C} \).
\end{Definition}

The subsequent critical concept is \textbf{type theory}, a branch of mathematical logic that provides a formal framework for constructing languages. The development of a type theory typically proceeds in two stages: (1) \textbf{raw syntax}, which defines readable but initially meaningless expressions, and (2) \textbf{derivable judgments}, which assign meaning to these expressions by defining contexts, types, and terms inductively. Variables within judgments are often managed using indices to effectively handle scope and binding.

The semantics of a type theory (potentially incorporating logic) within a categorical framework can be described through an adjunction:

\[
\mathsf{type\ theories} \underset{\mathbf{Lan}}{\overset{\mathbf{Con}}{\rightleftarrows}} \mathsf{Categories}
\]

where:
\begin{itemize}
    \item \(\mathbf{Lan}\) (semantics) assigns to a category its internal type theory, with types, terms, and propositions (if present) corresponding to objects, morphisms, and subobjects of the category.
    \item \(\mathbf{Con}\) (syntax) constructs the syntactic category of a type theory, where objects, morphisms, and subobjects correspond to the types (or contexts), terms, and propositions of the type theory.
\end{itemize}

A common approach to constructing the semantics of type theory involves the use of model categories under specific conditions. These categories provide a robust framework for higher categorical semantics, including homotopy type theory (HoTT). For example, higher inductive types and path-based equality can be effectively modeled in this context, as discussed in Sections 2.4 and 3.5 of \cite{belfiore2021topos}.

The deep connection between type theory and category theory is encapsulated in the following correspondence: intensional Martin-Löf dependent type theory is equivalent to locally cartesian closed \((\infty, 1)\)-categories. This relationship highlights the role of categorical frameworks in representing the semantics of dependent type theories, making them a foundational tool in modern logic and computation.

\begin{Definition}[Locally Cartesian Closed \cite{lurie2009higher}]
    Let \( \mathsf{C} \) be an \((\infty, 1)\)-category. 
    \begin{enumerate}
        \item \(\mathsf{C}\) is said to have \textbf{finite \((\infty, 1)\)-limits} if:
        \begin{itemize}
            \item It contains a terminal object, and
            \item For every object \( x \in \mathsf{C} \), the over-category \( \mathsf{C}/x \) has finite products.
        \end{itemize}
        \item \(\mathsf{C}\) is said to be \textbf{locally Cartesian closed} if:
        \begin{itemize}
            \item For every object \( x \in \mathsf{C} \), the over-category \( \mathsf{C}/x \) is a cartesian closed \((\infty, 1)\)-category. This means that \( \mathsf{C}/x \) has all finite limits, and for any two objects \( u, v \in \mathsf{C}/x \), the exponential object \( v^u \) exists.
            \item Equivalently, for any morphism \( f: y \to x \) in \( \mathsf{C} \), the pullback functor \( f^*: \mathsf{C}/y \to \mathsf{C}/x \) has a right adjoint, denoted \( \Pi_f \), which satisfies the adjunction:
            \[
            \mathsf{Hom}_{\mathsf{C}/x}(f^*v, u) \simeq \mathsf{Hom}_{\mathsf{C}/y}(v, \Pi_f(u)).
            \]
        \end{itemize}
    \end{enumerate}
\end{Definition}

\begin{Definition}[Lifting, Weak Equivalence, and Model Category \cite{lurie2009higher}]
    Let \( \mathsf{C} \) be a category.

    \begin{enumerate}
        \item \textbf{Lifting Property:}  
        Given a commutative diagram:
        \[
        \begin{tikzcd}
            A \arrow[r, "f"] \arrow[d, "i"'] & X \arrow[d, "p"] \\
            B \arrow[r, "g"'] \arrow[ru, dashed, "h"] & Y
        \end{tikzcd}
        \]
        If there exists a morphism \( h: B \to X \) such that \( h \circ i = f \) and \( p \circ h = g \), we say:
        \begin{itemize}
            \item \( i \) has the \textbf{left lifting property} with respect to \( p \),
            \item \( p \) has the \textbf{right lifting property} with respect to \( i \).
        \end{itemize}

        \item \textbf{Weak Factorization System (WFS):}  
        A \textbf{weak factorization system} on \( \mathsf{C} \) is a pair \( (\mathcal{L}, \mathcal{R}) \), where:
        \begin{itemize}
            \item Every morphism \( f: X \to Y \) in \( \mathsf{C} \) can be factored as \( f = \eta \circ \epsilon \), where \( \epsilon \in \mathcal{L} \) and \( \eta \in \mathcal{R} \).
            \item \( \mathcal{L} \) is precisely the class of morphisms having the left lifting property against every morphism in \( \mathcal{R} \), and \( \mathcal{R} \) is the class of morphisms having the right lifting property against every morphism in \( \mathcal{L} \).
        \end{itemize}

        \item \textbf{Model Category:}  
        A \textbf{model category} is a category \( \mathsf{C} \) that is complete and cocomplete, equipped with three distinguished classes of morphisms:
        \begin{itemize}
            \item \textbf{Cofibrations} (\( \mathcal{C} \)),
            \item \textbf{Fibrations} (\( \mathcal{F} \)),
            \item \textbf{Weak Equivalences} (\( \mathcal{W} \)),
        \end{itemize}
        satisfying the following axioms:
        \begin{enumerate}
            \item \textbf{Two-out-of-three property:} The weak equivalences \( \mathcal{W} \) satisfy the two-out-of-three property: If \( f: X \to Y \) and \( g: Y \to Z \) are morphisms such that two of \( f \), \( g \), and \( g \circ f \) are in \( \mathcal{W} \), then so is the third.
            \item \textbf{Factorization:} Every morphism \( f: X \to Y \) in \( \mathsf{C} \) can be factored in two ways:
            \[
            f = p \circ i, \quad \text{where } i \in \mathcal{C} \text{ and } p \in \mathcal{F} \cap \mathcal{W},
            \]
            \[
            f = q \circ j, \quad \text{where } j \in \mathcal{C} \cap \mathcal{W} \text{ and } q \in \mathcal{F}.
            \]
            \item \textbf{WFS Compatibility:} The pairs:
            \[
            (\mathcal{C}, \mathcal{F} \cap \mathcal{W}) \quad \text{and} \quad (\mathcal{C} \cap \mathcal{W}, \mathcal{F})
            \]
            form weak factorization systems.
        \end{enumerate}
    \end{enumerate}
\end{Definition}

The subsequent concept explores the theme of invariance.

\begin{Definition}[Action of a Sheaf of Categories\cite{maclane1992sheaves}]
    Let \( \mathbf{F}: \mathsf{C} \to \mathsf{Cat} \) and \( \mathbf{M}: \mathsf{C} \to \mathsf{Cat} \) be sheaves of categories over a site \( \mathsf{C} \). An \textbf{action of \( \mathbf{F} \) on \( \mathbf{M} \)} consists of a family of contravariant functors \( f_U: \mathbf{F}(U) \to \mathbf{M}(U) \) for each object \( U \in \mathsf{C} \), satisfying the following conditions:
    
    \begin{enumerate}
        \item For every morphism \( \alpha: U \to U' \) in \( \mathsf{C} \), the functors \( f_U \) are \textbf{equivariant} under the restriction maps \( \mathbf{F}_\alpha: \mathbf{F}(U') \to \mathbf{F}(U) \) and \( \mathbf{M}_\alpha: \mathbf{M}(U') \to \mathbf{M}(U) \). This means that the following \textbf{equivariance formula} holds:
        \[
        f_U \circ \mathbf{F}_\alpha = \mathbf{M}_\alpha \circ f_{U'}.
        \]
        
        \item This equivariance condition ensures compatibility of the action across morphisms in \( \mathsf{C} \), generalizing the concept of group actions to sheaves of categories.
    \end{enumerate}

    The action enables the definition of \textbf{orbits of sections} within \( \mathbf{M} \) under the influence of \( \mathbf{F} \), allowing the study of invariance and equivariance properties analogous to morphisms of stacks.
\end{Definition}

\begin{Definition}[Cosheaf on a Site \cite{maclane1992sheaves}]
    Let \( \mathsf{C} \) be a site with a Grothendieck topology. A \textbf{cosheaf} on \( \mathsf{C} \) is a copresheaf:
    \[
    \mathbf{F}: \mathsf{C} \to \mathsf{Set}
    \]
    that satisfies the following colimit condition for every covering family \( \{U_i \to U\}_{i \in I} \) in \( \mathsf{C} \):
    \[
    \mathbf{F}(U) \cong \varinjlim \left( \bigsqcup_{i,j} \mathbf{F}(U_i \times_U U_j) \to \bigsqcup_{i} \mathbf{F}(U_i) \right),
    \]
    where the diagram reflects the consistency of the copresheaf with respect to overlapping regions of the cover.

    The category of cosheaves on \( \mathsf{C} \), denoted \( \mathbf{CoSh}(\mathsf{C}) \), is the full subcategory of \( \mathbf{CoPSh}(\mathsf{C}) \), the category of copresheaves on \( \mathsf{C} \), consisting of those copresheaves that satisfy this colimit condition.
\end{Definition}

\subsubsection{Topoi and Stacks of Deep Neural Networks}
\label{dnn}
The main approach starts with constructing a base category that corresponds to the network graph structure, using a freely generated category. Then, the learning process is segmented into several irreducible components, each of which can be implemented either independently or in combination. These components are represented as functors that map the base category to well-known categories such as the category of sets (\(\mathsf{Set}\)), groupoids (\(\mathsf{Grpd}\)), or small categories (\(\mathsf{Cat}\)). When dealing with a target category as straightforward as \(\mathsf{Set}\), applying transformations like inverting the category direction turns these learning functors into contravariant functors, effectively transforming them into presheaves on the base category.

The main result of their work can be summarized as follows.
\begin{enumerate}
\item The authors have demonstrated that every artificial deep neural network (DNN) corresponds to an object within a standard Grothendieck topos; the learning dynamics of these networks are depicted as a flow of morphisms in this topos. In fact, they provided detailed explanations of the idea introduced in the reports of Laurent \cite{Lafforgue2022AI,Laurent2024}.

\item Invariant structures within network layers, such as those in CNNs or LSTMs, are linked to Giraud’s stack structures. These invariances are thought to relate to universal properties, allowing inferential extensions of learning data under specific constraints.

\item Artificial languages are defined on the fibers of the stack, which represent presemantic categories. These fibers incorporate internal logics, including intuitionistic, classical, or linear logics, providing a formal framework for reasoning.

\item Network functions are characterized as semantic functions, capable of articulating theories in artificial languages. These functions allow networks to generate meaningful responses to queries based on input data.

\item Entropy is employed to quantify the amount and distribution of semantic information within the network, providing a measure of its semantic capacity.

\item Geometric fibrant objects in Quillen’s closed model categories are used to categorize semantics. These categories are particularly useful for generating homotopy invariants, which capture the fundamental invariance of the system.

\item Intensional Martin-Löf Type Theory (MLTT) is applied to structure geometric fibrous objects and their fibrations. MLTT offers a constructive framework for reasoning about these objects.

\item Grothendieck derivators are utilized to analyze the content and exchange of information within the categorical framework. 
\end{enumerate}

\noindent \textbf{Correspondence between Deep Neural Networks and Grothendieck Topos}
The authors have demonstrated that every artificial deep neural network (DNN) corresponds to an object within a standard Grothendieck topos; the learning dynamics of these networks are depicted as a flow of morphisms in this topos. This builds on foundational ideas introduced in the reports of Laurent \cite{Lafforgue2022AI,Laurent2024}.

The first step in this framework is to conceptualize the network as a graph structure, referred to as the underlying graph. In artificial neural networks, the direct inter-layer connections naturally form a finite-oriented graph \( \Gamma \), which can be represented as a classical directed graph embedded in space-time dimensions. Feedforward and feedback mechanisms within the network correspond to a partial order \(\leq\) and its opposite. The partial order \(\leq\) is defined by the existence of oriented paths formed through the concatenation of directed edges.

For the simplest network architecture, consider a chain as the underlying graph \( \Gamma \). The dynamic objects of this chain in the context of feedforward learning is described using two covariant functors: the feedforward functor \(\mathbf{X}: C^\circ(\Gamma) \to \mathsf{Set}\), and the weight functor \(\mathbf{W} = \Pi: C^\circ(\Gamma) \to \mathsf{Set}\). Here, \( C^\circ(\Gamma) \) denotes the freely generated category, constructed from \( \Gamma \) by including all arrows generated through the composition of edges, without imposing additional relations.
During learning, the feedforward function is implemented as a covariant functor \(\mathbf{X}: C^\circ(\Gamma) \to \mathsf{Set}\). Each layer \( L_k \) of the network corresponds to a set of neuron activities \( X_k \), while each edge \( L_k \to L_{k+1} \) is represented by a mapping \( X_{k+1,k}^w: X_k \to X_{k+1} \), where \( w_{k+1,k} \) are the learned weights.

Similarly, the weights are represented by a covariant functor \( \mathbf{W} = \Pi: C^\circ(\Gamma) \to \mathsf{Set} \), where \( L_k \) corresponds to the product set "all sets \( w_{l+1,l} \) (\( l \geq k \)) of weights after this layer" \( \Pi_k \). The edge \( L_k \to L_{k+1} \) is mapped to the forgetting projection \( \Pi_{k+1,k}: \Pi_{k+1} \to \Pi_k \), which ``forgets'' weights associated with subsequent layers.

The Cartesian product \( X_k \times \Pi_k \) represents all possible configurations of neuron activities and weights, forming a covariant functor \( \mathbf{X} \) that captures the combined feedforward process. The output of the network depends solely on the current input and the network’s parameters, namely the weights. In supervised learning, the backpropagation algorithm is abstracted as a natural transformation of the functor \(\mathbf{W}\) to itself, which holds not only for chain graphs but also for more complex architectures.

To establish a topos for neural networks, the opposite category \(\mathsf{C} = C(\Gamma)\) of \(C^\circ(\Gamma)\) is introduced. The contravariant functors \(\mathbf{X}^\mathbf{W}\) (the dual of \(\mathbf{W}\)), \(\mathbf{W} = \Pi\), and \(\mathbf{X}\) then become presheaves(contravariant functors) on \(\mathsf{C}\), forming objects of the topos \(\mathsf{C}^\wedge\). This topos, associated with neural networks, exhibits a chain-like structure analogous to a multi-layer perceptron.

The object \(\mathbf{X}^w\)(\( X_{k+1,k}^w: X_k \to X_{k+1} \), representing the learned weights \(w_{k+1,k}\), is modeled as a subobject of \(\mathbf{X}\). Each fiber of this subobject, corresponding to a point \(w\) in \(\mathbf{X}\), is defined through the projection map \(\text{pr}_2: \mathbf{X} \to \mathbf{W}\) in the topos \(\mathsf{C}^\wedge\) from the terminal object. The terminal object \(1\), treated as a constant functor per layer (\(\star\)), maps to \(w\), representing a system of weights for the edges of the graph \(\Gamma\).

For chain-like structures, the subobject classifier \(\Omega\) provides logic in the topos. For each \(k \in \mathsf{C}\), \(\Omega(k)\) localizes \(1|k\), forming arrows in \(\mathsf{C}\) pointing to \(k\). These subobjects form an increasing sequence \((\emptyset, \cdots, \emptyset, \star, \cdots, \star)\), reflecting propositions in the topos becoming increasingly certain as the final layer is approached. This progression mirrors the gradual refinement of the network’s inference process.

For more complex topologies of \(\Gamma\), such as in networks with memory units (e.g., RNNs, LSTMs, GRUs), functions and weights may not be definable solely by functors on \(C(\Gamma)\). In such cases, canonical modifications of \(\Gamma\) are necessary, such as ensuring no one-to-many directed edges or duplicating nodes to handle intersecting inputs. Regardless of the network structure, the authors demonstrated that the presheaf framework can always be established. In supervised learning, the Backpropagation algorithm is represented as a flow of natural transformations of the functor \(\mathbf{W}\) to itself, enabling adjustments to the underlying graph to align with specific network architectures.

$$
\scalebox{0.7}{\includegraphics{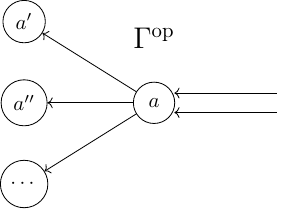}}
\quad
\scalebox{0.7}{\includegraphics{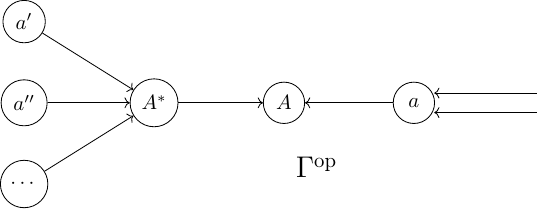}}
$$

Supervised learning involves the choice of an energy function \( F \), where the gradient of \( F \) is calculated during training using the formula \( dF(\delta w) = F^\star d\xi_n(\delta w) \). Here, the pullback \( F^\star \) transforms gradient information, and \(\mathbf{W}_0 = \prod_{a \in \Gamma} W_{aA}\) represents the product of weights over all edges in the graph \(\Gamma\). The expression \( d\xi_n(\delta w_a) = \sum_{\gamma_a \in \Omega_a} d\Phi_{\gamma_a} \delta w_a \) describes the gradient of \( F \), where \(\Phi\) is the mean \( E(F) \) of the energy over a measure on the inputs \(\xi_0\). The gradient of \(\Phi\) is the mean of individual gradients, and \(\delta w_a\), representing variations in weights over the edge \( Aa \), belongs to the tangent space \( T_{w_0}W_{aA} \). All other vectors are assumed to be zero except those denoting the weights over specific edges.

Directed paths \(\gamma_a\) in \(\Gamma\), which span from \(a\) to the output layer \(x_n\), create a zigzag pattern in the graph, such as \( \ldots \leftarrow B' \rightarrow b' \leftarrow B \rightarrow b \leftarrow \ldots \). Each variation \( b' \leftarrow B \) corresponds to the injection \(\rho_{B'}\), defined by the factors \( X_{B'}, X_{B''}, \ldots \) of \( X_B \). The overall formulation of the gradient is then expressed as:
\[
d\xi_n(\delta w_a) = \sum_{\gamma_a \in \Omega_a} \prod_{b_k \in \gamma_a} DX_{b_k B_k}^{w_0} \circ D\rho_{B_k b_{k-1}} \circ \partial_w X_{aA}^w \delta w_a,
\]
where \( DX_{b_k B_k}^{w_0} \) and \( D\rho_{B_k b_{k-1}} \) represent the derivatives of transformations along the directed path, and \( \partial_w X_{aA}^w \) captures the dependency of the output on the weights. 

After generalizing the framework of the network, the definition of the presheaf \(\mathbf{X}^w\) (over \(\mathsf{C}\)) is formalized as follows: At each original vertex, the set \(\mathbf{X}_a^w\) signifies the activity of neurons in the corresponding network layer. The mapping from \(\mathbf{X}_A\) to \(\mathbf{X}_a\) represents dynamic transmission within the network, linking information from all input layers, such as \(a'\), \(a''\), and others. The initial input information and neuronal activity uniquely determine the initial state of the topos. Given a collection of activities \(\varepsilon_{in}^0\) in all initial layers of the network, this collection generates a unique section of the presheaf \(\mathbf{X}^w\), either a singleton or an element of \(\lim_C \mathbf{X}^w\), which in turn induces \(\varepsilon_{in}^0\). The natural alignment of projections, such as weights, with layer conditions is maintained. Defined on \((\mathsf{C}, J)\), the layer \(\mathbf{W}\) represents the potential set of weights for deep neural networks (DNNs) or recurrent neural networks (RNNs).

Regarding topos construction, the research team provided several insights into the potential applications of machine learning based on classifying topoi, with an emphasis on semantics. For example, as discussed in \cite{caramello2022ontologies}, semantic inference is derived from ontologies represented by topoi. A fundamental structure, in the form of an oriented graph, facilitates specific types of information flow through its arrows. Notably, they mentioned that a network of topoi and morphisms relating them can model and define information processing procedures similar to those discussed in this paper. \cite{Caramello2022} provides a summary of the foundational details of their theory.

They observed that a fragment of a topos, consisting of a collection of objects and arrows, can be depicted using sites with underlying oriented graphs that encapsulate this fragment. Moreover, the topos can be described by theories sufficiently equipped with an appropriate vocabulary and axioms.

Topos structures and topos-based semantics have historically been employed in analyzing wireless and biological networks to model and understand the flow of information. A comprehensive review of this research is available in \cite{hmamouche2021new}. These studies highlight several fundamental concepts:
\begin{enumerate}
    \item When local structures within an information processing network exhibit coherence, appearing seamlessly integrated into a unified whole, topos structures become especially relevant. This approach underscores the ``local vs. global'' principle—a central theme in topos theory—which formalizes how local interactions can collectively manifest as global coherence within the network.
    
    \item In information processing networks, each node performs dual roles: receiving and transmitting information. Consequently, it is essential to recognize the internal structure of each node in relation to its information processing capabilities. This perspective aligns with the Grothendieck construction, which associates fibers with the base category, providing a structured representation of nodes and their interactions. This foundational framework enables the development of topos structures, further extended into stacks (2-sheaves) as described in \cite{belfiore2021topos}. Stacks facilitate a richer representation of hierarchical relationships and interdependencies, capturing the complex dynamics of nodes within a network.
\end{enumerate}

\noindent \textbf{The Stack of DNNs}
The second significant result from \cite{belfiore2021topos} explores the stack structure within Deep Neural Networks (DNNs). Utilizing the initial topos framework, the study investigates the network’s functional structure \( \mathbf{X}^w \) and the internal learning constraints of the model \( \mathbf{W} \) (such as weight propagation in feedforward and feedback mechanisms), which are shaped by geometric or semantic invariants. This approach echoes discussions in \cite{Caramello2022}.

Within the context of DNNs, invariants are extracted from the input data and preserved through to the output, presenting a significant challenge in network design. For instance, in convolutional neural networks (CNNs) used for image analysis, each layer of the network must support the action of a group \( G \), and the connections between layers must be compatible with this action. \cite{belfiore2021topos} highlights that the study of invariants should occur within the \textbf{stack} (or 2-sheaf) structure of DNNs.

One model for exploring these invariants employs a contravariant functor defined as a category from the network \( \mathsf{C}^{op} \to G^{\wedge} \), where \( G^{\wedge} \) denotes the topos of \( G \)-sets. In this framework, \( G \) is treated as a category with a single object, and a \( G \)-set (with a left action of \( G \)) is conceptualized as a sheaf valued in the category of sets over \( G \), i.e., \( G^{op} \to \mathsf{Set} \). The analysis extends to the algebraic structure on the space of contravariant functors, where equivariant natural transformations define morphisms between the functors, forming a category \( \mathsf{C}_G^{\sim} \), which also constitutes a topos.

According to established results, over a 1-site, the 2-functors (functors between stacks) on the site are equivalent to fibered categories over the site. Thus, in this framework, stack theory can be comprehensively expressed in terms of fibered categories, as outlined in \cite{vistoli2004notes}.

Within this framework, there exists a category \( \mathsf{F} \), which forms a fibration \( \pi: \mathsf{F} \to \mathsf{C} \) with groups isomorphic to \( G \) and adheres to the stack axioms. This category is equipped with a canonical topology \( J \). As a result, the topos of sheaves valued in \( \mathsf{Set} \), denoted as \( \mathcal{E} = \mathsf{F}^{\sim} \), is naturally equivalent to \( \mathsf{C}_G^{\sim} \) over the site \( (\mathsf{F}, J) \). Therefore, it constitutes a classifying topos.

In more general cases, multiple models for neural networks are possible. The primary focus in their study lies in the internal structure and logic of DNNs. Specifically, the internal logic of classifying topoi operates fiber-to-fiber, reflecting the hierarchical organization of DNNs.

Groupoid structures provide an appropriate framework for this analysis. Presheaves on a groupoid constitute a Boolean topos, where the logic of the topology of a groupoid comprises simple Boolean algebras. Each atom \( Z_i; i \in K \), formed by irreducible \( G_a \)-sets, facilitates semantic analysis by capturing the fundamental invariants within the network.

In convolutional neural networks (CNNs), the irreducible linear representations of the group are expressed in dynamic objects concerning invariant subspaces. These representations preserve geometric and semantic invariants, ensuring consistency across layers. Additionally, intuitionistic Heyting algebras can be incorporated externally to model the future and uncertain past, integrating these discussions into topos theory within the groupoid fibers. The critical insight is that groupoid stacks are deeply connected to internal groupoids. According to the Grothendieck construction, every stack is equivalent to a strict 2-functor from \( \mathsf{C}^{op} \) into the 2-category \( \mathsf{Grpd} \) of groupoids. This strict 2-functor corresponds to a functor into the 1-category of groupoids, which can be represented as a groupoid object in presheaves. Sheafification transforms this into a groupoid, enabling the application of the 2-functor 
\[
\mathbf{F}: \mathsf{Grpd}(Sh(\mathsf{C}, J)) \to Sh_J(\mathsf{C}, \mathsf{Grpd}) \to St(\mathsf{C}, J),
\]
as described in \cite{bunge1979stack}.

Beyond the groupoid structure, various forms of intuitionistic logic under different frameworks can be incorporated. For example, invariance can be studied under group actions, which are captured categorically as (contravariant) functors from \( \mathsf{G} \) to \( \mathsf{V} \). Within a category \( \mathsf{V} \), a morphism \( \phi: u \to v \) can be interpreted as an element of the object \( v \) in \( \mathsf{V} \). Group action orbits can be modeled using fibers and cofibers, with slice categories representing these relationships. This structure extends naturally to stacks, where the fibered category \( \mathcal{F} \) acts upon the fibered category \( \mathcal{M} \).

Consider a sheaf \( \mathcal{F}: \mathsf{C} \to \mathsf{Cat} \), representing the general structure of invariance, and another sheaf \( \mathcal{M}: \mathsf{C} \to \mathsf{Cat} \). Both sheaves are objects within a stack. The action of \( \mathcal{F} \) on \( \mathcal{M} \) involves a family of functors \( \mathbf{f}_U: \mathcal{F}_U \to \mathcal{M}_U \). For any morphism \( \alpha: U \to U' \) within \( \mathsf{C} \), corresponding morphisms in the stack ensure that the following diagram commutes.

\[
\begin{tikzcd}
\mathcal{F}(U) \arrow[r, "\mathbf{f}_U"] \arrow[d, "\mathcal{F}_\alpha"'] & \mathcal{M}(U) \arrow[d, "\mathcal{M}_\alpha"] \\
\mathcal{F}(U') \arrow[r, "\mathbf{f}_{U'}"'] & \mathcal{M}(U')
\end{tikzcd}
\]

This represents the equivariant formula that generalizes group equivariance and corresponds to morphisms within the stack. Specifically, it is possible to define the orbit of sections \( \mathbf{u}_U \to \mathbf{f}_U (\xi_U) \) in the sheaf \( \mathbf{u} | \mathcal{M} \) under the action of the stack \( \mathcal{F} | \xi \).

The next objective is to identify the subobject classifier. Locally, for each object \( U \) in the category \( \mathsf{C} \), the sheaf \( \mathcal{F}(U) \), where \( \mathcal{F}: \mathsf{C}^{op} \to \mathsf{Cat} \), forms a small category. The functors \( \mathcal{F}_{\alpha}: \mathcal{F}(U') \to \mathcal{F}(U) \) are identifiable, making \( \mathcal{F}(U) \) a local classifying topos with the classifying object \( \Omega_U \), denoted by \( \mathcal{E}_U \). Within each \( \mathcal{E}_U \), natural transformations \( \operatorname{Hom}_U(\mathbf{X}_U, \Omega_U) \) correspond directly to subobjects of \( \mathbf{X}_U \).

A section \( (S_U, S_{\alpha}) \) defines a presheaf \( A \), given by \( A_U(\xi) = \operatorname{Hom}_{\mathcal{F}_U}(\xi, S_U) \). Whether the pullback functor \( \mathcal{F}^*_{\alpha}: \mathcal{E}_U \to \mathcal{E}_{U'} \), which maps \( A \circ \mathcal{F}_{\alpha} \) to \( A \), qualifies as geometric determines whether \( \mathcal{F}_{\alpha} \) satisfies the stack property. Even when the strict geometric property is not satisfied, the extended condition of being ``weakly geometric and open'' can be considered.

Moreover, significant attention is paid to two fundamental functors:
\begin{itemize}
    \item \( \lambda_{\alpha} = \Omega_{\alpha} : \Omega_{U'} \to \mathcal{F}^*_{\alpha} \Omega_U \), representing feedforward propagation within the DNN.
    \item \( \lambda'_{\alpha}: \Omega_U \to \mathcal{F}^*_{\alpha} \Omega_{U'} \), representing feedback propagation.
\end{itemize}

These functors stem from the adjunction \( \Omega_{\alpha} \dashv \lambda'_{\alpha} \), which establishes a duality between feedforward and feedback propagation. This leads to the following theorem:

\begin{Theorem}[Theorem~ 2.1 in \cite{belfiore2021topos}]
For each morphism \( \alpha: U \to U' \) within the category \( \mathsf{C} \), when the functor \( \mathcal{F}_{\alpha} \) serves as both a fibration and a morphism of groupoids, it propagates logical formulas and their truth values across the topos from \( U \) to \( U' \) via \( \lambda'_{\alpha} \). Concurrently, the logic within the topos propagates in the reverse direction—from \( U' \) to \( U \)—via \( \lambda_{\alpha} \). Furthermore, the map \( \lambda_{\alpha} \) is the left adjoint of the transpose \( \tau'_{\alpha} \) of \( \lambda'_{\alpha} \). Thus, for any morphism \( \alpha: U \to U' \) in \( \mathsf{C} \), the following condition holds:
\[
\lambda_{\alpha} \circ \tau'_{\alpha} = 1_{\Omega_{U'}}.
\]
\end{Theorem}

To conclude, the intrinsic logic of categorical topology, characterized by transitions between fibers of a stack, is equated to describing the potential (or optimal) flow of information within a neural network. Within this theoretical framework:

\begin{enumerate}
    \item \textbf{Type Theory and Functional Operations}: 
    A type-theoretic language is introduced, where types are derived from presheaves defined on the fibers of the stack. Within this language, terms are used to represent the theories and propositions relevant to the functional operation of deep neural networks (DNNs). Invariance is captured through a pre-semantic category, which acts on the language and possible sets of theories. Network functionality is then expressed as the generation, interpretation, and application of these theories.
    
    \item \textbf{Semantic Interpretation Across Layers}: 
    Semantics in neural networks are established within the stack structure, where different layers of the network interpret logical propositions and statements of the output layer. As one moves closer to the output layer, the fidelity of interpretation increases, ensuring progressively more accurate representations of the intended semantics. This hierarchical structure enables the network to refine its semantic understanding as it processes information through its layers.
\end{enumerate}

\noindent \textbf{Theories, Interpretation, Inference, and Deduction}  
The logical theory specified by and specifying a given category \( \mathsf{C} \), called its \textbf{internal logic}, satisfies the following:
\begin{enumerate}
    \item \textbf{Types} are the objects \( A \) of \( \mathsf{C} \).
    \item \textbf{Contexts} are the slice categories \( \mathsf{C} / A \).
    \item \textbf{Propositions in context} are the \((-1)\)-truncated objects \( \Phi \) of \( \mathsf{C} / A \).
    \item \textbf{Proofs} \( A \vdash \text{PhiIsTrue}: \Phi \) are the generalized elements of \( \Phi \).
\end{enumerate}

The general structure of a formal language can be expressed as:  
\[
(P_1 \vdash Q_1, P_2 \vdash Q_2, \ldots, P_n \vdash Q_n) / P \vdash Q.
\]
In this framework, the conditional validity of a proposition \( R \) is written as \( \vdash_S R \). A valid proof of \( R \) is represented as a directed acyclic graph (DAG) with vertices labeled by valid reasoning steps. Directed edges connect the final vertex's upper endpoint to the initial vertex's lower endpoint. The DAG has exactly one final vertex, whose lower item is labeled \( \vdash_S R \). The initial vertex contains either an empty set or an element of \( S \). Within a formal language \( L \), a theory \( \mathbb{T} \) is a set of propositions assumed to be true, meaning they are derivable via valid proofs from axioms.

A correspondence exists between topos subobjects and logical theories. In interpreting a language \( L \) within a topos \( \mathcal{E} \):
\begin{itemize}
    \item Objects in \( \mathcal{E} \) are associated with types.
    \item Arrows \( A \to B \) correspond to variables of \( B \) in the context of \( A \).
    \item The subobject classifier \( \Omega_{\mathcal{E}} \) corresponds to the logical type \( \Omega_L \).
\end{itemize}

Logical constructs such as products, subsets, power sets, singletons, context transformations, and logical rules are all compatible within the topos framework. The projections of a topos (existence and universality) ensure this compatibility. A theory \( \mathbb{T} \) is represented in \( \mathcal{E} \) if all its axioms are true in \( \mathcal{E} \) and all derivations are valid.

\noindent In many applications of Deep Neural Networks (DNNs), the network must perform semantic analysis on data. Two key questions arise: 
\begin{enumerate}
    \item What does the analysis result specifically mean?
    \item How can an external observer access the network's internal processes for this analysis?
\end{enumerate}
The network is modeled as a dynamic object \( X \) in a topos \( \mathcal{E} \), with learning weight objects \( W \). Here, \( \mathcal{E} \) is a classifying topos with the fibration \( \pi: F \to C \). A morphism \( \alpha: U \to U' \) indicates that the logic in \( U' \) is richer than in \( U \).

Assume a family of typed languages \( \{L_U; U \in C\} \), interpreted in the respective topoi \( \{\mathcal{E}_U; U \in C\} \) of different layers. Feedback mechanisms manifest as natural transformations \( L_\alpha: L_{U'} \to \mathcal{F}^*_\alpha L_U \), extending \( \Omega_\alpha = \lambda_\alpha \). Similarly, feedforward mechanisms are described by natural transformations \( L'_\alpha: L_U \to \mathcal{F}^*_\alpha L_{U'} \), extending \( \lambda'_\alpha \). These mechanisms form adjunctions, ensuring bidirectional propagation of semantics and theories across layers.

In the stack of DNNs, two functors \( \mathcal{F}_\alpha: \mathcal{F}_{U'} \to \mathcal{F}_U \) are defined: one for the ordinary input-output mapping and another from a standard projection of the fiber at a point \( A \). Language \( L \) is obtained as a presheaf over \( \mathsf{C} \), with logical types \( \Omega_L \) and \( \Omega_{L_U} \), and theory sets \( \Theta_U = \mathbf{P}(\Omega_{L_U}) \), where \( \mathbf{P} \) is the power set functor.

The semantic functionality of the dynamic object \( \mathbf{X}^\mathbf{W} \) regarding \( \mathbb{T}_{\text{out}} \) is a family of quotients \( \mathbf{X}^\mathbf{W}_U \) of \( D_U \), equipped with mappings \( S_U: D_U \to \Theta_U \). For each input \( \xi_{\text{in}} \), \( S_U(\xi_U) \) generates a theory consistent with \( \mathbb{T}_{\text{out}}(\xi_{\text{in}}) \), valid bi-directionally across paths. For each neuron \( a \) in \( L_U \), quantized activity \( \epsilon_a \) implies the validity of a proposition \( P_a(\epsilon_a) \) in \( \Omega_{L_U} \). This ensures that for any \( \xi_{\text{in}} \), the generated set of propositions \( P_a(\epsilon_a) \) presupposes the validity of propositions in \( \Omega_{L_{\text{out}}} \), consistent with \( \mathbb{T}_{\text{out}}(\xi_{\text{in}}) \).

Applications and experimental results supporting this framework are provided in \cite{belfiore2021logical}.

From the experiments in \cite{belfiore2021logical}, it is observed that the internal layers interpret the language \( L_{\text{out}} \), indicating that the functor \( \mathbf{f} = \mathbf{g}^* = \mathcal{F}_\alpha^* \) propagates the language backward. This backward propagation provides a mechanism to analyze the logical information associated with any subset \( E \) of neurons in a given layer. Specifically, within the activity of \( E \), the set of propositions predicted to be true in \( \mathbb{T}_{\text{out}}(\xi_{\text{in}}) \) can be identified.

If all involved sets are finite, the amount of information provided by \( E \) can be quantified as:
\[
\frac{\text{Number of predicted propositions}}{\text{Number of required decisions}}.
\]
The average value of this ratio can serve as an assessment metric for the input \( \xi_{\text{in}} \).

Their experiments show that the evolution of \( \mathbf{W} \)-subsets from \( X_0 \) to \( X_t \) is robust, indicating that the dynamics within \( \mathbf{W} \) related to these subsets are potentially ideal points of stability. This robustness suggests that, in practical applications, these subsets can be specifically targeted for stabilization, which could significantly enhance the efficiency and robustness of the learning process.

\noindent \textbf{Model Category and M-L Type Theory of DNNs}  
In the model category, homotopy between mappings can be concretely described. After localization by weak equivalences, weak equivalences become invertible morphisms (homotopy equivalences), although it may not be possible to identify which morphisms are homotopic. However, within the model category, homotopies between mappings, as well as higher-order homotopies between homotopies, can be explicitly observed, akin to the category of topological spaces. This structure optimally preserves the original information.  

In particular, within the stack structure, the internal logic of the classifying topos operates fiber-to-fiber, which suggests that fibrations and cofibrations should be treated persistently, as in Quillen’s models, where the entire space is both fibrantized and cofibrantized.  

This work focuses on a fixed layer of a DNN. The stack set on its architecture exhibits the following properties: the fibers of these stacks are groupoids within a given category, possessing a natural Quillen closed model category structure. This enables the association of homotopy theory and intensional type theory, with stacks corresponding to fibrant objects and permissible contexts.  

\begin{enumerate}
    \item The category of stacks (stacks in groupoids) is equivalent to the category derived from \( \mathsf{C}_X \). Using the groupoid fibration (where stacks are fibrant objects, and weak equivalences are homotopy equivalences within the fibers), properties of the groupoid category \( \mathsf{Grpd}_\mathsf{C} \) can correspond to a Martin-Löf type theory.
    
    \item This conclusion extends to general stack categories using Quillen model theory, not limited to groupoids. A natural association exists between M-L intensional theory and Quillen models. Furthermore, Quillen model theory can connect to Voevodsky’s theory.
    
    \item The categories considered are primarily models in \( \mathsf{Grpd} \) and \( \mathsf{Cat} \). In \( \mathsf{Cat} \), fibrations are isomorphisms lifted by functors, cofibrations are injective functors, and weak equivalences are categorical equivalences.
    
    \item Locally, an association exists between the classifying object and this structural system. Given the partial order structure inherent in DNNs, starting from a poset \( \mathsf{C} \) of a DNN, a closed model category \( \mathsf{M} \) induces a fibration in \( \mathsf{M}_\mathsf{C} \) at each object of \( \mathsf{C} \). This structure provides the context for M-L theory.
    
    \item In this context, a type is defined as a fibration. Logical operators such as conjunction (\( \mathcal{A} \wedge \mathcal{B} \)), disjunction (\( \mathcal{A} \vee \mathcal{B} \)), implication (\( \mathcal{A} \Rightarrow \mathcal{B} \)), negation (\( \perp \)), existential (\( \exists x, \mathcal{B}(x) \)), and universal (\( \forall x, \mathcal{B}(x) \)) apply to types.
    
    \item To summarize, if \( \mathsf{C} \) is a poset of a DNN, there exists a canonical M-L structure where contexts and types correspond to geometric fibrations in the 2-category of contravariant functors \( \mathsf{Cat}_\mathsf{C} \), and base change substitutions correspond to 1-morphisms.

    \item Similar considerations apply to \( \mathsf{Grpd} \). Thus, M-L theory defines languages and semantics over DNNs with internal structures modeled in \( \mathsf{M} \).
\end{enumerate}
Related work primarily focuses on type theory and categorical semantics. The concepts of fibers and pre-semantic categories are derived from these fields, integrating principles from both type theory and categorical semantics. Categorical semantics is distinguished by its structural construction. In this framework, mathematical logic is translated into a formal language describing the collection of monomorphisms into a specific object of a given category, known as the poset of subobjects of that object. 

Additionally, interpretative type theory, particularly dependent type theory, can be translated into a formal language describing slice categories, which include all morphisms into a specified object. Conversely, if a category exists such that a given theory of logic or type theory serves as its internal logic, the theory is considered to possess categorical semantics \cite{shulman2019comparing,shulman2021large}. 

Notably, prior to the work in \cite{belfiore2021topos}, no research had explored the application of (pre-)stacks-based and (pre-)sheaves-based semantics in machine learning. This approach introduces a novel perspective on semantics that is particularly well-suited for programming applications. Integrating this framework with deep neural networks (DNNs) aligns the learning processes with programming techniques, providing a robust foundation for advanced semantic analysis in machine learning.

\noindent \textbf{Dynamics and Homology}  
In the context of supervised and reinforcement learning, the success or failure of network outputs serves as a key measure of performance. The dynamic object \( \mathbf{X}^w \) describes the activities in DNNs, which correspond to decisions made by the network's output (e.g., the ``cat's manifold''). The propositions about the input depend on the final output, and the primary objective is to understand how the entire DNN contributes to decision-making regarding the output. To achieve this, the output propositions are transformed into truth values and the network structure of \( \mathsf{C} \) is expanded accordingly.  

Decision-making for propositions in this expanded structure involves category-theoretic approximations. Specifically, the subset of activities of \( \mathbf{X} \) that confirm the proposition \( P_{\text{out}} \) is described by its right Kan extension. The resulting global invariant, \( M(P_{\text{out}})(\mathbf{X}) \), corresponds to the cohomology \( H^0(\mathsf{C}_+; \mathbf{X}_+) \). The network structure of \( \mathsf{C} \) is expanded with the goal of ensuring equivalence:  
\[
H^0(\mathsf{C}_+; \mathbf{X}_+) \cong H^0(\mathsf{C}; \mathbf{X}).
\]

The dynamics of forward propagation correspond to the limits of \( H^0(\mathbf{X}) \). Starting from \( \mathbf{X}^w \), global dynamic objects with spontaneous activities are defined. Global dynamics are obtained using the implicit function theorem during the learning process. Through backpropagation, new dynamic inputs are introduced for each branch, refining the network's semantic interpretation.

In this work, the authors define stacks and costacks and, through semantic action, construct models in the classifying topos such that their homotopy and homology invariants translate into semantic information quantities and spaces associated with the topos. The internal dimensions of stacks over \( \mathsf{C} \) reflect different levels of information.  
\begin{itemize}
    \item The first level involves pertinent types and objects, representing direct semantic content.
    \item The second level pertains to higher-order information, such as the theory \( \mathbb{T}_{U^\prime} \), which depends on the given section of \( \mathbf{X}^w \).
\end{itemize}

Moreover, effective communication between layers and networks must be considered to achieve precise semantic information. This interplay facilitates a more comprehensive understanding of the global dynamics and decision-making processes within the network.

Existing results introduce the concept of information quantity, analogous to entropy in Shannon's information theory, when integrating information and accounting for redundancy. It has been shown that entropy in the topos structure of Bayesian networks corresponds to first-order cohomology, specifically \( Ext^1(K, \mathcal{F}_P) \), related to the presheaf of numerical functions on the probability layer \( \mathcal{F}_P \). Higher-order mutual information corresponds to cocycles in higher-dimensional homotopy algebra. This work extends these properties to DNNs. By analyzing morphisms from fiber to fiber, the correspondence of subobjects between different fibers is interpreted as transformations of context in topos semantics.

In probabilistic models, the marginalization of probabilities is represented as a covariant functor, corresponding to the transformation of theories. Both forward and backward directions of propagation are considered. In topological network models, these transformations correspond to mappings between different fibers in stacked DNNs. By considering local contexts within topos semantics, transformations of subobjects in different local contexts are obtained. Extending this framework to typed languages, the aforementioned transformations manifest as propositions under different contexts (reality and expectation).

To formalize this further, consider the statement presheaf \( \mathcal{A} \) associated with the stack \( \mathcal{F} \). Expressing its language and theories in a categorical framework requires defining sheaves and cosheaves. By applying semantic modulation in categorical topology, monoids on moduli are defined, endowing their homology or homotopy invariants with semantic information within the topos.

Additionally, a Heyting algebra for a proposition \( P \) provides a partial order for the set of theories within a given context. This endows these sets of theories with a presheaf structure, under which the concept of ``conditioning'' is interpreted as a monoid action on the presheaf. Similarly, copresheaves and cosheaves are considered, where the cosheaf structure incorporates measurable functions concerning given contexts and propositions.

Analogous to existing research in the Bayesian framework, two possibilities arise in this semantic framework for interpreting network layers:
\begin{enumerate}
    \item Each random variable of the network layer corresponds to a proposition \( P \), allowing logical values to be measured.
    \item Random variables represent layers \( U \), gerbe objects, or other types, which are more suitable for understanding the dynamic semantic processes of feedforward and feedback mechanisms.
\end{enumerate}

The semantic information and functional interpretation of these layers can be quantified, enabling the definition of the fuzzy nature of propositions within a given context and dynamic process. Using 0-cochains in homology theory, this information fuzziness can be interpreted in conjunction with the precision of the theory. Another method for comparing information between layers involves establishing the canonical cohomology of categorical neurons, providing a measure for the evolution of fuzziness in the network.

Considering the action on a layer, conditioning can be interpreted as a monoid action on that layer. The monoid's action on propositions and the naturality of operations can be formally defined. A key conclusion is the formula:
\[
I(P_1; P_2)(T) = \psi(T|P_1 \land P_2) - \psi(T|P_1) - \psi(T|P_2) + \psi(T),
\]
which illustrates the association between propositions and theories and quantifies the information contained within propositions under specific conditions.

Furthermore, the study explores the language structure of information based on the results of Carnap and Bar-Hillel. It examines the action of the Galois group on the language and its orbits, the information space of types, the fundamental groups of the information space, internal invariance supported by groupoids of braids, and the stability of information transport between layers, which remains independent of most input details. Additionally, higher structures, such as the 3-category of DNNs, are introduced to formalize these relationships further.

In conclusion, \cite{belfiore2021topos} provides a groundbreaking framework that integrates topos theory, type theory, and categorical semantics to analyze and interpret the dynamics and semantic structures of Deep Neural Networks (DNNs). It introduces novel applications of homology, cohomology, and presheaf-based semantics to model information flow, invariance, and logical operations within DNNs. By extending these ideas to higher categorical structures and incorporating concepts like Galois actions and internal invariants, the work offers a robust mathematical foundation for understanding and enhancing the learning processes of neural networks.

\subsection{Other Related Works}
The compositional categorical structure discussed in Section~\ref{sec_2} closely aligns with topos theory. As shown in \cite{spivak2022learn}, the map \( A \to B \) in \( \mathbf{Para}(\mathsf{SLens}) \) can be naturally interpreted in terms of dynamical systems, specifically as generalized Moore machines. Consequently, the derived category \( \mathsf{p\text{-}Coalg} \), equipped with a coalgebraic structure, forms a topos for dynamical systems on any \( p \in \mathsf{Poly} \). This structure provides a powerful framework where logical propositions can be expressed and manipulated in its internal language.

In \cite{villani2024topos}, transformer networks are examined as a unique case compared to other architectures like convolutional and recurrent neural networks. While these architectures are embedded within a pre-topos of piecewise-linear functions, transformer networks reside in their topos completion, enabling higher-order reasoning. This advancement builds on the concept of ``corresponding to a geometric model'' introduced in \cite{Lafforgue2022AI} and leverages the foundational categorical framework detailed in \cite{cruttwell2022categorical}.

Further contributions to semantic investigation include the extended fibered algebraic semantics for first-order logic presented in \cite{bloomfield2024fibered}, and the relative topos theory framework developed in \cite{caramello2022fibred}, which generalizes the construction of sheaf topoi on locales. These results provide critical insights into the relationships between Grothendieck topoi and elementary topoi, forming a robust mathematical foundation for exploring semantics in categorical and logical contexts.

Several application-oriented results have been reported in the literature, highlighting the integration of semantic frameworks with advanced mathematical and communication theories:

\begin{itemize}
    \item \cite{choi2022unified} leveraged the logical programming language ProbLog to unify semantic information and communication by integrating technical communication (TC) and semantic communication (SC) through the use of internal logics. This approach demonstrates how logical programming can bridge semantic and technical paradigms to enhance communication systems.
    
    \item \cite{chaccour2024less} examined semantic communication in AI applications, focusing on causal representation learning and its implications for reasoning-driven semantic communication networks. The authors proposed a comprehensive set of key performance indicators (KPIs) and metrics to evaluate semantic communication systems and demonstrated their scalability to large-scale networks, thereby establishing a framework for designing efficient, learning-oriented semantic communication networks.
    
    \item \cite{sergeant2024compositional} explored the mathematical underpinnings of statistical systems by representing them as partially ordered sets (posets) and expressing their phases as invariants of these representations. By employing homological algebra, the authors developed a methodology to compute these phases, offering a robust framework for analyzing the structural and statistical properties of such systems.
\end{itemize}

\section{Conclusion}
In this survey, we reviewed recent advancements in category-theoretical and topos-theoretical frameworks within machine learning, categorizing the studies into four main directions: gradient-based learning, probability-based learning, invariance- and equivariance-based learning, and topos-based learning. The primary research in gradient- and probability-based learning shares compositional frameworks, while probability-based and invariance- and equivariance-based learning often draw upon overlapping foundational categories, typically involving similar metric or algebraic structures. In contrast, topos-based learning introduces a distinct perspective, emphasizing global versus local structures rather than focusing solely on the functoriality of basic compositional components. Nonetheless, it integrates homotopical and homological viewpoints, establishing partial alignment with invariance- and equivariance-based learning.

This survey aims to outline potential future directions in ``categorical machine learning.'' A key consideration for future research is to further investigate how structural properties preserved across system components and semantic integration within frameworks can be effectively leveraged to advance machine learning methodologies.

\authorcontributions{
Conceptualization, Yiyang Jia; Supervision, Yiyang Jia; Surveying, Yiyang Jia, Guohong Peng, Zheng Yang, Tianhao Chen; writing---review and editing, Yiyang Jia (Sections 1, 5, 6), Guohong Peng (Section 3), Zheng Yang (Section 4), Tianhao Chen (Section 2); funding acquisition, Yiyang Jia, Guohong Peng. All authors have read and agreed to the published version of the manuscript.
}

\funding{This work is supported in part by the JSPS KAKENHI Grant (No. JP22K13951) and the Doctoral Science Foundation of XiChang University (No. YBZ202206). }

\institutionalreview{Not applicable.}

\informedconsent{Not applicable.}

\dataavailability{Not applicable. } 




\acknowledgments{The authors appreciate Prof. Wei Ke at Macao Polytechnic University for his critical review of this manuscript. }

\conflictsofinterest{The authors declare no conflicts of interest.} 




\begin{adjustwidth}{-\extralength}{0cm}

\reftitle{References}


\bibliography{sn-bibliography}


\PublishersNote{}
\end{adjustwidth}
\end{document}